\documentclass[10pt,letterpaper, twocolumn]{article}
\newcounter{todos}
\AtEndDocument{\ifnum\value{todos}>0 \PackageWarning{TODOS}{There are \arabic{todos} todos left in this paper! Fix them before submitting the paper!} \fi}

\newcommand{\argmin}{\mathop{\mathrm{argmin}}\limits}

\newcommand{\norm}[1]{\lVert#1\rVert}

\newcommand{\V}[1]{\ensuremath{\mathbf{#1}}}

\newcommand{\abs}[1]{\left\lvert#1\right\rvert}

\newcommand*\Bell{\ensuremath{\boldsymbol\ell}}
\newcommand{\R}{\ensuremath{\mathbb{R}}\xspace}

\newcommand{\curve}{\ensuremath{\mathcal{S}}\xspace}

\newcommand{\surface}{\ensuremath{\mathcal{M}}\xspace}

\newcommand{\point}{\ensuremath{\V{x}}\xspace}
\newcommand{\normal}{\ensuremath{\V{n}}\xspace}

\newcommand{\loss}{\mathcal{L}\xspace}

\newcommand{\pointOne}{\ensuremath{\point_{1}}\xspace}
\newcommand{\pointTwo}{\ensuremath{\point_{2}}\xspace}
\newcommand{\pointsetOne}{\ensuremath{\chi_{1}}\xspace}
\newcommand{\pointsetTwo}{\ensuremath{\chi_{2}}\xspace}

\newcommand{\cameracenter}{\ensuremath{\V{o}}\xspace}
\newcommand{\viewDir}{\ensuremath{\V{v}}\xspace}

\newcommand{\lightDir}{\ensuremath{\V{\Bell}}\xspace}
\newcommand{\lightIntensity}{\ensuremath{e}\xspace}
\newcommand{\halfvec}{\ensuremath{\V{h}}\xspace}

\newcommand{\opacity}{\ensuremath{\alpha}\xspace}
\newcommand{\sigmoid}{\ensuremath{\sigma}\xspace}
\newcommand{\sharpness}{\ensuremath{a}\xspace}

\newcommand{\sdfFunc}{\ensuremath{\mathcal{G}}\xspace}
\newcommand{\brdfFunc}{\ensuremath{\mathcal{F}}\xspace}
\newcommand{\shadowFunc}{\ensuremath{\mathcal{S}}\xspace}
\newcommand{\hashenc}{\ensuremath{\mathcal{H}}\xspace}
\newcommand{\angularenc}{\ensuremath{\mathcal{A}}\xspace}

\newcommand{\pixel}{\ensuremath{\V{p}}\xspace}

\newcommand{\colorbar}[3]{
\begin{tabular}[t]{@{}l@{}l@{}}
	\includegraphics[height=#1\linewidth,width=0.5em]{colorbar.pdf} & 
	\begin{tabular}[b]{@{}c}
		#2 \\ [#3pt]
		$0$
	\end{tabular}
\end{tabular}
}

\newcommand{\colorbartwo}[2]{
\tiny
	\begin{tabular}[b]{@{ }l}
        #2 \\
        \vspace{-0.3em}
		\includegraphics[height=#1\linewidth, width=0.5em]{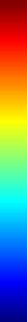} \\ 
        $0$
	\end{tabular}
}

\newcommand{\sdmunips}{\mbox{SDM-UniPS~\cite{ikehata2023sdmunips}}\xspace}
\newcommand{\psnerf}{\mbox{PS-NeRF}~\cite{yang2022psnerf}\xspace}
\newcommand{\supernormal}{\mbox{SuperNormal}~\cite{Cao_2024_CVPR}\xspace}

\newcommand{\diligentmv}{\mbox{DiLiGenT-MV}~\cite{li2020multi}\xspace}

\newcommand{\svnl}{SVNL~\cite{Brahimi_2024_CVPR}\xspace}

\newcommand{\dpir}{DPIR~\cite{Chung_2024_CVPR}\xspace}

\newcommand{\rawnerf}{\mbox{RawNeRF~\cite{mildenhall2022rawnerf}}\xspace}

\usepackage{config/iccv}              
\usepackage{graphicx}
\usepackage{amsmath, mathtools, empheq, esint}
\usepackage{amssymb}
\usepackage{booktabs}
\usepackage{wrapfig}
\usepackage{lipsum}
\usepackage{xspace}
\usepackage{amsthm}
\usepackage{arydshln}
\usepackage{siunitx}
\usepackage{xcolor,colortbl}
\usepackage{multirow}
\usepackage{tabularx}
\graphicspath{{images/}}

\usepackage{bm}
\usepackage{etoc}
\usepackage{makecell}
\usepackage{array}

\usepackage{pifont}
\newcommand{\cmark}{\ding{51}}%
\newcommand{\xmark}{\ding{55}}%

\definecolor{cvprblue}{rgb}{0.21,0.49,0.74}
\usepackage[pagebackref,breaklinks,colorlinks,allcolors=cvprblue]{hyperref}

\usepackage[capitalize]{cleveref}
\crefname{section}{Sec.}{Secs.}
\Crefname{section}{Section}{Sections}
\Crefname{table}{Table}{Tables}
\crefname{table}{Tab.}{Tabs.}

\begin{document}

\title{Neural Multi-View Self-Calibrated Photometric Stereo \\ without Photometric Stereo Cues}
\author{
Xu Cao 
\quad
Takafumi Taketomi
\\
CyberAgent, Japan
\\
{\tt \small \{xu\_cao, taketomi\_takafumi\}@cyberagent.co.jp} \\
\tt \small Code:\url{https://github.com/CyberAgentAILab/MVSCPS}
}

\twocolumn[{
    \renewcommand\twocolumn[1][]{#1}
    \maketitle
    \centering
    \vspace{-1.5em}
    \begin{minipage}{\textwidth}
        \centering
        \includegraphics[trim=000mm 000mm 000mm 000mm, clip=False, width=\linewidth]{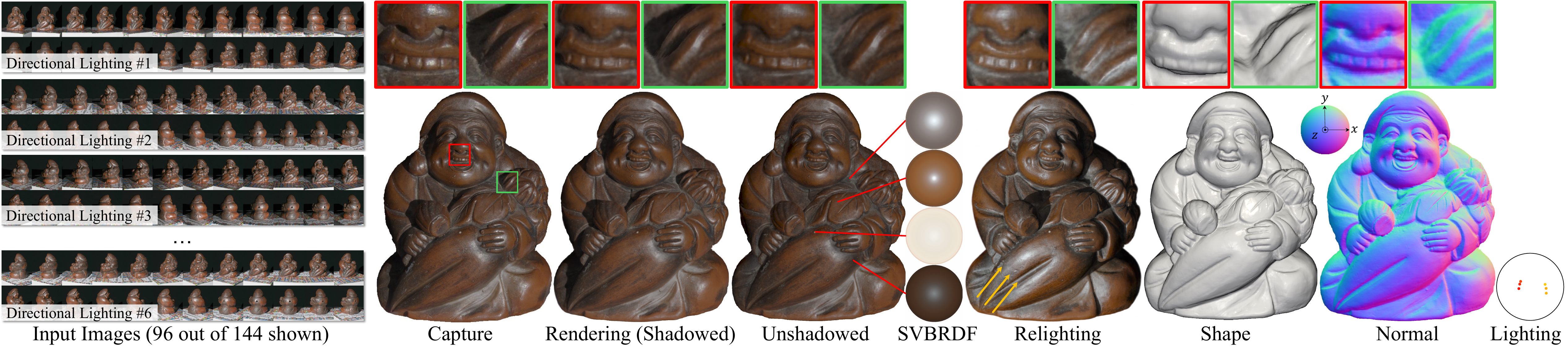}
    \end{minipage}
    \vspace{-1.1em}
    \captionof{figure}{
  Using multi-view posed images captured under varying directional lighting, our method jointly recovers geometry, spatially varying reflectance, and per-light direction and relative intensity, enabling unshadowed rendering and relighting.
  All scene components are estimated in a single-stage optimization from raw pixel measurements, without relying on intermediate cues such as normal maps.
    }
    \label{fig.teaser}
    \vspace{1.5em}
}]
\begin{abstract}
We propose a neural inverse rendering approach that jointly reconstructs geometry, spatially varying reflectance, and lighting conditions from multi-view images captured under varying directional lighting.
Unlike prior multi-view photometric stereo methods that require light calibration or intermediate cues such as per-view normal maps, our method jointly optimizes all scene parameters from raw images in a single stage.
We represent both geometry and reflectance as neural implicit fields and apply shadow-aware volume rendering.
A spatial network first predicts the signed distance and a reflectance latent code for each scene point.
A reflectance network then estimates reflectance values conditioned on the latent code and angularly encoded surface normal, view, and light directions.
The proposed method outperforms state-of-the-art normal-guided approaches in shape and lighting estimation accuracy, generalizes to view-unaligned multi-light images, and handles objects with challenging geometry and reflectance.
\end{abstract}

\section{Introduction}
\label{sec.intro}
Recovering intrinsic scene properties---geometry, reflectance, and lighting---from image observations is a long-standing inverse problem in computer vision and graphics.
The challenge arises from the ambiguity that different combinations of scene properties can produce the same visual appearance.
To mitigate the ambiguity, photometric stereo (PS)~\cite{woodham1980photometric} captures multiple images of the target object from a fixed viewpoint by activating a single light source from a distinct direction per exposure.
Using these view-aligned one-light-at-a-time (OLAT) images, PS estimates per-pixel surface normals and reflectance by analyzing intensity variations across lighting.
To enable full 3D reconstruction, multi-view photometric \mbox{stereo (MVPS)} extends PS by repeatedly capturing OLAT image stacks from multiple viewpoints~\cite{li2020multi,Logothetis_2019_ICCV}.

Although multi-view OLAT stacks provide dense photometric observations across views and directional lighting conditions, existing MVPS methods fail to fully exploit this information due to their stage-by-stage processing.
A typical pipeline begins with calibrating the lighting using specialized objects such as a chrome ball~\cite{chung2024differentiable} or a diffuse white board~\cite{li2020multi}. 
Per-view PS cues, especially the normal maps, are then estimated from each OLAT stack and fused into a coherent 3D shape~\cite{Cao_2024_CVPR,brument2024rnb}.
Reflectance properties are finally recovered based on the reconstructed geometry~\cite{yang2022psnerf}.
This stage-by-stage pipeline accumulates errors across stages.
In particular, estimating PS cues independently for each view neglects cross-view observations and often leads to inconsistent multi-view predictions, ultimately degrading geometry reconstruction and reflectance recovery.

In contrast to stage-by-stage pipelines, we propose an end-to-end neural inverse rendering framework that jointly recovers geometry, reflectance, and lighting via single-stage optimization (\cref{fig.teaser}).
Our rendering pipeline comprises three key components:
(1) a spatial multilayer \mbox{perceptron (MLP)} that predicts both the signed distance and a reflectance latent code for each scene point,
(2) a bidirectional reflectance distribution function (BRDF) MLP that predicts reflectance values given the latent code and normal-view-light directions, and
(3) a shadow MLP that refines shadow factors derived from signed distance function (SDF)-based transmittance.
To improve reconstruction quality, we apply angular encoding to normal-view-light directions and use a weighted L1 color loss.

Despite not relying on intermediate PS cues, our method outperforms stage-by-stage MVPS methods guided by normal maps~\cite{Cao_2024_CVPR,yang2022psnerf} regarding geometry reconstruction quality and lighting estimation accuracy.
Thanks to end-to-end optimization that exploits raw pixel measurements, our method remains robust under sparse or even no lighting variations, where stage-by-stage MVPS methods deteriorate due to inaccurate PS estimation.
Further, with a neural BRDF representation, our method is effective for various challenging reflectance types, such as ceramic (\cref{fig.teaser}) and bronze metallic (\cref{fig.real_world_results}) objects.

A practical benefit of decoupling MVPS from per-view PS cues is improved capture flexibility.
Traditional MVPS methods often use an electronically controlled LED array to automate light switching under each view~\cite{li2020multi, Logothetis_2019_ICCV}. 
In addition to such a setup, our method supports view-unaligned OLAT capture with minimal light switching.
For each light, we keep it active and fixed to the camera while capturing a full set of multi-view images, and then switch to the next light for another set of multi-view images (\cref{fig.teaser}, left). 
Viewpoints need not be aligned across lighting\footnote{In fact, the seminal MVPS work~\cite{esteban2008multiview} uses such view-unaligned OLAT capture, yet it has rarely been followed by subsequent studies.}.

In summary, our key contributions are as follows:
\begin{itemize}
    \item  We propose a robust neural inverse rendering framework for MVPS that jointly estimates geometry, reflectance, and lighting in a single-stage optimization without relying on intermediate PS cues or light calibration;
    \item We show that a neural latent-driven BRDF can be effectively trained from scratch using multi-view OLAT images, and a combination of angular encoding and a weighted L1 loss improves overall reconstruction quality;
    \item We validate our method qualitatively on real-world view-unaligned OLAT images, demonstrating its applicability to objects with challenging geometry and reflectance.
\end{itemize}

\section{Related Work}
\label{sec.related_work}
Our work is related to MVPS that inputs multi-view OLAT images and uses neural BRDFs in the rendering pipeline.
\subsection{Multi-View Photometric Stereo}
We categorize existing MVPS methods into three types based on their reliance on intermediate PS cues.

\vspace{0.3em}
\noindent
\textbf{Stage-by-stage} methods process multi-view and multi-light images in separate stages.
\underline{\emph{Multi-view-first}} methods initialize a coarse shape using multi-view images and then refine it using per-view OLAT images.
The initial shape can be a visual hull obtained from multi-view foreground masks~\cite{Logothetis_2019_ICCV,esteban2008multiview}, a surface reconstructed via multi-view stereo~\cite{park2016robust,park2013multiview,wu2010fusing}, or sparse 3D points obtained by structure from motion~\cite{zhou2013multi,li2020multi}.
\underline{\emph{Multi-light-first}} methods estimate per-view PS cues from each OLAT stack and then fuse them into a coherent 3D shape.
The PS cues include normal maps~\cite{Cao_2024_CVPR,chang2007multiview}, normal plus depth maps~\cite{kaya2022uncertainty}, normal plus reflectance maps~\cite{brument2024rnb,bruneau2025multi}, azimuth maps~\cite{mvas2023cao}, or depth maps~\cite{zhao2023mvpsnet}.

Stage-by-stage pipelines are suboptimal due to information isolation and error accumulation across stages. 
Single-view PS estimation neglects available multi-view observations, requires dense lighting variations, yet still fails to ensure cross-view consistency. 
Subsequent multi-view reconstruction ignores the original images and cannot faithfully correct errors propagated from inconsistent PS cues. 
Consequently, recent methods such as SuperNormal~\cite{Cao_2024_CVPR} exhibit geometric artifacts caused by normal inconsistency across views.
In contrast, our method mitigates geometric artifacts and remains robust under sparse or even no lighting variations, thanks to the end-to-end pipeline.

\vspace{0.3em}
\noindent
\textbf{Semi-end-to-end} methods use both multi-view OLAT images and intermediate PS cues to reconstruct geometry and optionally reflectance parameters, and are limited to view-aligned OLAT images.
Multi-view normal maps are used to guide the training of neural radiance field (NeRF)~\cite{kaya2022neural} or neural SDF~\cite{logothetis2024nplmv,kaya2023multi,Santo_2024_CVPR} from images.
\psnerf initializes a radiance field using multi-view normal maps and light-averaged images, and then estimates BRDF parameters from OLAT images with fixed geometry.
\svnl jointly estimates neural SDF, Disney BRDF, and point light parameters, but relies on per-view shadow cues to handle cast shadows.
Unlike \svnl, our rendering pipeline is shadow-aware and generalizes to view-unaligned OLAT images without compromising performance.

\vspace{0.3em}
\noindent
\textbf{End-to-end} methods recover scene properties directly from raw images without relying on intermediate PS cues, enabling them to handle both view-aligned and unaligned OLAT images.
To the best of our knowledge, only \dpir and our method use this end-to-end design.
\dpir uses point-based volume rendering to reconstruct a hybrid point and neural SDF representation, along with basis BRDF parameters, but requires known light directions and intensities.
In contrast, our method uses neural SDF-based volume rendering that achieves better geometric fidelity and does not require light calibration.

\subsection{Neural BRDFs}
\label{sec.neiral_brdf}
The BRDF defines how light is reflected from a surface point as a function of the incident and outgoing light directions~\cite{nicodemus1965directional}. 
A BRDF further depends on spatial position when it is spatially varying, and normal directions when defined in local coordinates~\cite{rusinkiewicz1998new}.
To characterize a BRDF, seminal works propose analytic models based on empirical observations~\cite{Phong1975IlluminationFC,blinn1977models} or the micro-facet assumption~\cite{cook1982}, often including parameters like diffuse albedo and specular roughness.

Recently, neural fields have emerged as a common BRDF parameterization option in inverse rendering~\cite{hofherr2025neural}.
\emph{\underline{Neural analytic BRDFs}} predict analytic BRDF parameters for each scene point, then use a predefined analytic model to evaluate the final BRDF values, such as Disney BRDFs~\cite{cheng2023wildlight,zhang2022iron}, simplified Disney BRDFs~\cite{Brahimi_2024_CVPR,srinivasan2021nerv,yao2022neilf,boss2021nerd}, and polarimetric BRDFs~\cite{dave2022pandora,li2025neisf++,li2024neisf}.
\emph{\underline{Neural basis BRDFs}} models the specular BRDF by basis functions, such as spherical Gussians~\cite{yang2022psnerf,li2022selfps} or learned basis~\cite{Chung_2024_CVPR}.
Spatial MLPs are then optimized to predict albedos and specular coefficients.
\emph{\underline{Neural latent-driven BRDFs}} predict BRDF values directly via an MLP conditioned on a spatially varying latent code, without relying on analytic models or basis functions.
NeRFactor~\cite{zhang2021nerfactor} pre-trains a latent-driven BRDF for the specular term using the measured BRDF dataset~\cite{Matusik2003ADR}. 
Then, the pre-trained BRDF is frozen in per-scene inverse rendering, and spatial MLPs are trained from scratch to predict albedos and specular latent codes.
In contrast, our method does not decompose the BRDF into diffuse and specular terms, but instead uses a single latent-driven MLP to directly predict the final BRDF values.
We show that neural latent-driven BRDFs can be optimized from scratch for inverse rendering tasks with multi-view OLAT inputs.

\section{Approach}
\label{sec.method}
We aim to reconstruct 3D geometry, SVBRDF, and lighting parameters given multi-view OLAT images of a target static object, along with the corresponding foreground masks and camera parameters.
In the following, we describe the image formation model in \cref{sec.capture_setup}, the forward rendering pipeline in \cref{sec.forward_rendering}, and the optimization strategy in \cref{sec.optimization}.

\subsection{Image Formation}
\label{sec.capture_setup}
We assume that each image of the target object is captured under a single directional light.
For each light \mbox{source $j$} among $M$ light sources, we capture images from $N_j$ viewpoints, resulting in a total of $\sum_{j=1}^{M} N_j$ OLAT images.
The number of viewpoints $N_j$ need not be identical under different light sources.
For clarity, we will omit lighting and viewpoint indices unless otherwise specified.

Consider a scene point $\point \in \R^3$ with normal $\normal \in \mathcal{S}^2$.
Neglecting inter-reflections, its radiance $r(\point) \in \R_+$ measured along the viewing direction $\viewDir \in \mathcal{S}^2$ under a light source with direction $\lightDir \in \mathcal{S}^2$ and intensity $\lightIntensity \in \R_+$ is given by
\begin{equation}
    r(\point) = \lightIntensity f(\point, \normal, \viewDir, \lightDir) (\normal^\top \lightDir)_+.
    \label{eq.render_equation_parallel_light}
\end{equation}
Here, \mbox{$f(\cdot) \in \R_+$} represents the BRDF, and $(\normal^\top \lightDir)_+ = \max(\normal^\top \lightDir, 0)$ accounts for cosine foreshortening. 
For color images, \cref{eq.render_equation_parallel_light} is applied independently to each channel, with channel-specific light intensities and BRDF values.

Given the radiance at each scene point, we perform shadow-aware volume rendering to compute the radiance observed at a pixel $\pixel \in \R^2$.
For a ray cast from the camera center $\cameracenter$ along direction \viewDir, we sample scene points as $\{\point_k = \V{o}+t_k\viewDir, \, t_{k+1}>t_k\}_{k=0}^{N}$.
The radiance measured at \pixel is
\begin{align}
    r(\pixel) &=  \sum_k T_k\alpha_k s(\point_k, \lightDir) r(\point_k),
\label{eq.volume_rendering_color}
\end{align}
where $s(\point_k, \lightDir) \in \{0, 1\}$ is a shadow factor indicating whether the scene point $\point_k$ is self-occluded from the light direction $\lightDir$, $\alpha_k$ denotes the opacity at the sample point $\point_k$ and $T_k= \prod_{m=0}^{k-1} (1-\opacity_m)$ represents the accumulated transmittance along the ray up to point $\point_k$.

Computing shadow factors at every sample point along the ray leads to quadratic complexity with respect to the number of samples.
To accelerate rendering, we assume that all sample points share the same shadow factor as the surface point $\point'$ (\ie, the first ray-surface intersection).
\Cref{eq.volume_rendering_color} is then approximated as
\begin{align}
    r(\pixel)
     & \approx s(\point', \lightDir)  \sum_k T_k\alpha_k  r(\point_k)\\
     &= \lightIntensity s(\point', \lightDir)  \sum_k T_k\alpha_k  f(\point_k, \normal_k, \viewDir, \lightDir)(\normal_k^\top \lightDir)_+.
\label{eq.volume_rendering_color_approx}
\end{align}
As such, evaluating \cref{eq.volume_rendering_color_approx} requires geometry, reflectance, and lighting parameters, which are detailed in \cref{sec.forward_rendering}. 

\begin{figure*}
    \centering
    \includegraphics[width=\linewidth]{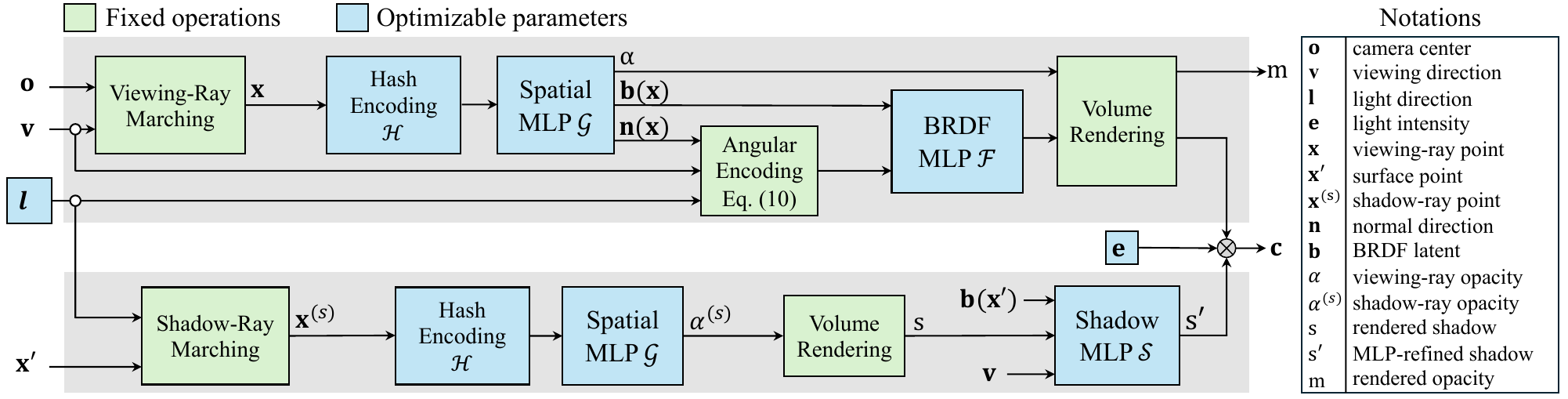}
    \vspace{-1.5em}
    \caption{\textbf{Overview of forward rendering.} The top branch computes local shading, and the bottom branch models shadow effects.}
    \label{fig.pipeline}
    \vspace{-1em}
\end{figure*}

\subsection{Scene Parameterization}
\label{sec.forward_rendering}
\Cref{fig.pipeline} illustrates our forward rendering pipeline.
We represent geometry as a neural SDF, reflectance as a neural BRDF, lighting as per-light directions and intensities, and use a shadow MLP to refine the rendered shadow factors.

\vspace{0.5em}
\noindent
\textbf{Geometry.}
We use a spatial MLP to predict the signed distance $g(\point) \in \R$ and a BRDF latent $\V{b}(\point)$ for each point \point:
\begin{equation}
    \left(g(\point), \V{b}(\point)\right) = \sdfFunc\left(\hashenc(\point;\boldsymbol{\phi});\boldsymbol{\theta}\right) .
\end{equation}
Here, \sdfFunc is the spatial MLP with parameters $\boldsymbol{\theta}$  and  $\hashenc$ is the multi-resolution hash encoding~\cite{mueller2022instant} with parameters $\boldsymbol{\phi}$.
The surface \surface is the set of zero-level points of the SDF:
\begin{equation}
    \surface = \{\point \mid g(\point)=0\}.
\end{equation}
A scene point's normal is the normalized SDF gradient
\begin{equation}
    \normal(\point; \boldsymbol{\phi}, \boldsymbol{\theta}) = \overline{\nabla g}.
\end{equation}
In this paper, the bar over a vector indicates it is normalized, \ie,$\overline{\nabla g}=\frac{\nabla g}{\norm{\nabla g}_2}$.
This normal remains an analytic function of the spatial MLP parameters~\cite{igr2020icml}.
Recent studies~\cite{Cao_2024_CVPR,brument2024rnb} have shown that supervising SDF gradients by multi-view normal maps alone can recover surfaces and that multi-resolution hash-encoding is beneficial for high-frequency detail recovery and training efficiency.

A scene point's opacity can be converted from signed distances~\cite{wang2021neus}. 
Given the set of sample points $\{\point_k \}$ on a ray, the opacity $\opacity_k$ at each point is defined as
\begin{align}
\label{eq.sdf_opacity}
    \opacity_k &= \max\left(\frac{\sigmoid(g(\point_k))-\sigmoid(g(\point_{k+1}))}{\sigmoid(g(\point_k))}, 0\right),
\end{align}
where $\sigmoid(x)=\frac{1}{1+\exp(-\sharpness x)} $ is the sigmoid function with an optimizable sharpness \sharpness. 

\vspace{0.5em}
\noindent
\textbf{BRDF.}
We use a single MLP to parameterize the BRDF:
\begin{equation}
    f(\point, \normal, \viewDir, \lightDir) = \brdfFunc\left(\V{b}(\point), \angularenc(\normal, \viewDir, \lightDir);\boldsymbol{\psi}\right).
\end{equation}
Here, $\brdfFunc$ is BRDF MLP with parameters $\boldsymbol{\psi}$, $\V{b}(\point)$ and $\normal$ are from the spatial MLP, 
$\angularenc(\normal, \viewDir, \lightDir)$ is angular \mbox{encoding (AE)} of normal-view-light directions.
We design the AE as
\begin{equation}
    \angularenc(\normal, \viewDir, \lightDir) = [\normal^\top \halfvec, \lightDir^\top\halfvec, \normal^\top \lightDir, \normal^\top \viewDir, (\normal^\top \halfvec)^{10}]^\top,
\label{eq.angularEnc}
\end{equation}
where $\halfvec=\overline{\lightDir+\viewDir}$ is the halfway vector.
To ensure non-negative BRDF values, ReLU activations are used in all MLP layers.
The output dimension of \brdfFunc is identical to the number of image channels, \eg, $3$ for RGB images.

The study~\cite{rusinkiewicz1998new} shows that proper parameterization of BRDF variables improves functional approximation from measured reflectance.
Similarly, angular encoding transforms normal-view-light directions into a rotation-invariant representation that facilitates learning and improves generalization.
With angular encoding, scene points with the same BRDF are more likely to be mapped to the same latent code, even if they differ in world-space normals, and the MLPs need not implicitly learn the rotation invariance.

\begin{figure}
\scriptsize
    \centering
    \begin{tabular}{@{}c@{}c@{}c@{}c@{ }c}
         \includegraphics[width=0.24\linewidth]{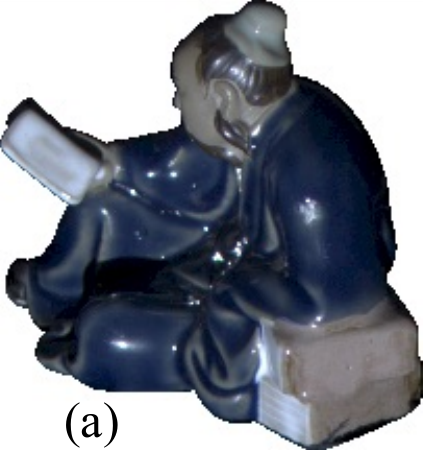} &       \includegraphics[width=0.24\linewidth]{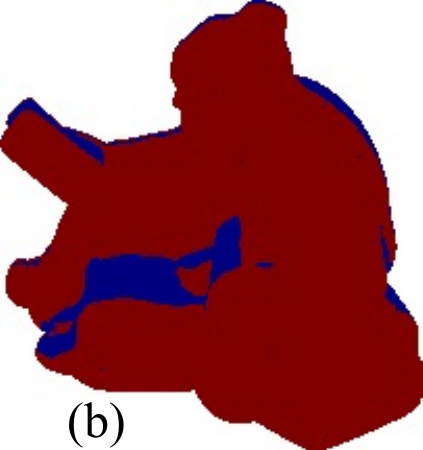} &\includegraphics[width=0.24\linewidth]{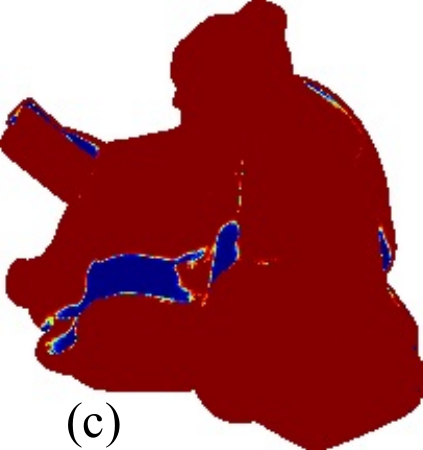} &\includegraphics[width=0.25\linewidth]{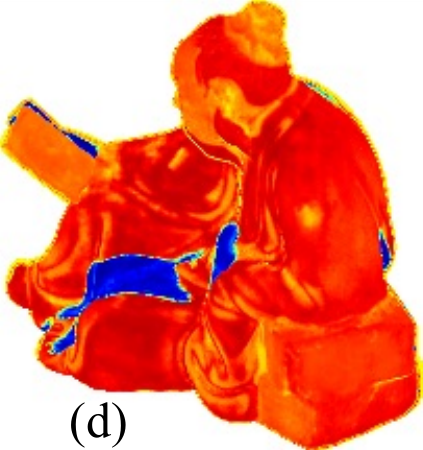} 
         &
         \colorbar{0.1}{1}{11}
    \end{tabular}
    \vspace{-1.5em}
    \caption{Our shadow rendering strategy. \textbf{(a)} Captured image. \textbf{(b)} Binary shadow map rendered from the GT mesh with GT light direction. \textbf{(c)} Volume-rendered shadow map. \textbf{(d)} Shadow MLP refined shadow map.}
\label{fig.shadow_render}
    \vspace{-1em}
\end{figure}

\vspace{0.5em}
\noindent
\textbf{Lighting.}
We assume that each light source is static to the camera and parameterize light $j$ by its camera-space direction $\lightDir_j \in \curve^2$ and RGB intensities $\V{\lightIntensity}_j \in \R_+^3$.
During forward rendering, the light direction is transformed to world coordinates as $\lightDir=\V{R}_i\lightDir_j$, where $\V{R}_i \in SO(3)$ is the camera-to-world rotation matrix for the $i$-th viewpoint.

\vspace{0.5em}
\noindent
\textbf{Self-shadow.}
We first compute the shadow factor $s$ via transmittance accumulation and refine it to $s'$ using a shadow MLP.
First, we approximate the ray-surface intersection $\point'$ as:
\begin{equation}
    \point' \approx \cameracenter + d\viewDir, \quad \text{with} \quad d=  \sum_k T_k\alpha_k t_k.
\end{equation}
We cast a shadow ray from the surface point in the light direction, and sample points on it as $\{\point_k^{(s)} = \point'+t_k^{(s)} \lightDir, \, t_k^{(s)} \in [t_{near}, t_{far}]\}$.
We then compute the opacities $\{\alpha_k^{(s)}\}$ according to \cref{eq.sdf_opacity}, and the shadow factor $s$ is
\begin{equation}
    s = 1 - \sum_k T_k^{(s)} \alpha_k^{(s)}.
    \label{eq.shadow_render}
\end{equation}
\Cref{eq.shadow_render} states that when the light source is occluded, the shadow ray will intersect the surface, thus the accumulated transmittance reaches $1$, making the shadow factor $0$.
The volume-rendered shadow factors are nearly binary, but the measured shadow regions are not purely dark due to inter-reflections (\cref{fig.shadow_render} (a)).
To alleviate the influence, we use a shadow MLP \shadowFunc to refine the rendered shadow factors:
\begin{equation}
    s'= \shadowFunc(\V{b}(\point'), s, \viewDir;\boldsymbol{\varphi}).
\end{equation}
The viewing direction is included to account for view-dependent effects.

\subsection{Joint Optimization}
\label{sec.optimization}
Given the scene parameterization described in \cref{sec.forward_rendering}, we render the pixel value\footnote{We assume a linear radiometric response, \ie, pixel intensities are proportional to incoming radiance.} according to \cref{eq.volume_rendering_color} as 
\begin{equation}
    \V{c}(\pixel) = s' \V{\lightIntensity} \otimes \sum_k T_k\alpha_k  \brdfFunc\left(\point_k, \normal_k, \viewDir, \lightDir\right)\sigma_+(\normal_k^\top \lightDir).
\label{eq.optimization_render}
\end{equation}
Here, $\V{c}(\pixel) \in \R_+^3$ is the RGB color, and $\otimes$ denotes element-wise multiplication.
We approximate $\max(\cdot, 0)$ using the softplus \mbox{function $\sigma_+(\cdot)$} to improve numerical stability.
With the rendered colors $\V{c}$, we optimize spatial MLP, BRDF MLP, shadow MLP, and lighting parameters 
to reproduce the measured \mbox{colors $\hat{\V{c}}$} within the region of interest.

\vspace{0.5em}
\noindent
\textbf{Initialization.}
We initialize the spatial MLP parameters such that the initial SDF zero-level set approximates a sphere of radius $r$~\cite{atzmon2020sal}.
The lighting is initially set to frontal directional lighting with unit intensity, \ie, a direction of $[0, 0, -1]^\top$ in camera coordinates and RGB intensity of $[1, 1, 1]^\top$.
To ensure the light directions remain unit length during optimization, we always use the normalized ones $\overline{\lightDir}$ in forward rendering.

\vspace{0.5em}
\noindent
\textbf{Loss.} The loss function is the weighted sum of color loss, mask loss, and Eikonal loss:
\begin{equation}
    \loss = \loss_{color} + \lambda_1 \loss_{mask} + \lambda_2 \loss_{Eikonal},
\end{equation}
where $\lambda_1$ and $\lambda_2$ are hyperparameters balancing each term's influence.
Inspired by \rawnerf, we use a weighted L1 color loss to account for the high-dynamic-range images captured under a directional light source:
\begin{equation}
    \loss_{color} = \sum_{\pixel} \abs{\frac{{\V{c}(\pixel)} - \hat{\V{c}}(\pixel)}{\text{sg}(\V{c}(\pixel))+\epsilon}},
    \label{eq.color_loss}
\end{equation}
where $\text{sg}(\cdot)$ denotes the stop-gradient operator, and $\epsilon$ is a tiny constant avoiding zero division.
Unlike \rawnerf, which uses a weighted L2 loss, we find that a weighted L1 loss performs better for OLAT image inputs, as discussed in \cref{sec.ablation}.
The mask loss and Eikonal loss follow standard practice~\cite{wang2021neus,Cao_2024_CVPR} and are detailed in the supplementary.
The mask loss encourages the surface silhouette to align with the input masks, while the Eikonal loss enforces the neural SDF to take unit-norm spatial gradients in the vicinity of its zero-level set~\cite{igr2020icml}.

After optimization, the surface mesh can be extracted from the SDF by marching cubes~\cite {lorensen1987marching}, relighting can be realized by changing the input light directions, and unshadowed rendering can be realized by omitting the multiplication with $s'$ in \cref{eq.optimization_render}.

\section{Experiments}
\label{sec.experiment}
\Cref{sec.benchmark} evaluates MVPS methods in terms of geometry reconstruction, reflectance, and lighting estimation quality and includes ablation studies of the proposed method.
\Cref{sec.application} presents qualitative real-world inverse rendering results using view-unaligned OLAT images.
Implementation details are provided in the supplementary.

\subsection{Evaluation using View-aligned OLAT Images}
\label{sec.benchmark}
\noindent
\textbf{Dataset.}
\diligentmv is a widely used benchmark dataset for MVPS.
It contains 5 objects, each captured from 20 viewpoints under 96 distant point light sources (abbreviated as 96L20V hereafter).
The 96 lighting conditions are identical in camera coordinates across all views.
Ground-truth meshes are produced with a 3D scanner, manually cleaned, and aligned with the captured views.
The ground-truth normal maps are rendered from the scanned meshes, and light directions and intensities are calibrated using a diffuse white board.

\begin{figure*}
    \centering
    \begin{tabular}{@{}c@{}c@{}c@{}c@{}c@{}}
        \includegraphics[width=0.2\linewidth]{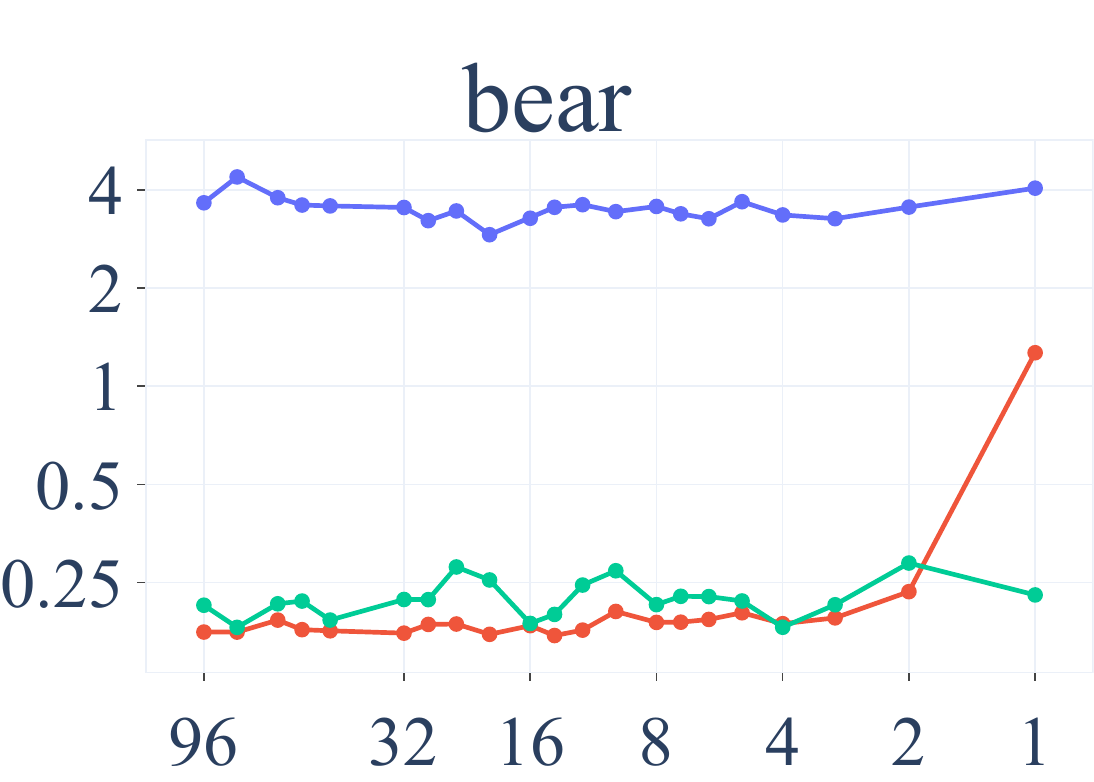}&
        \includegraphics[width=0.2\linewidth]{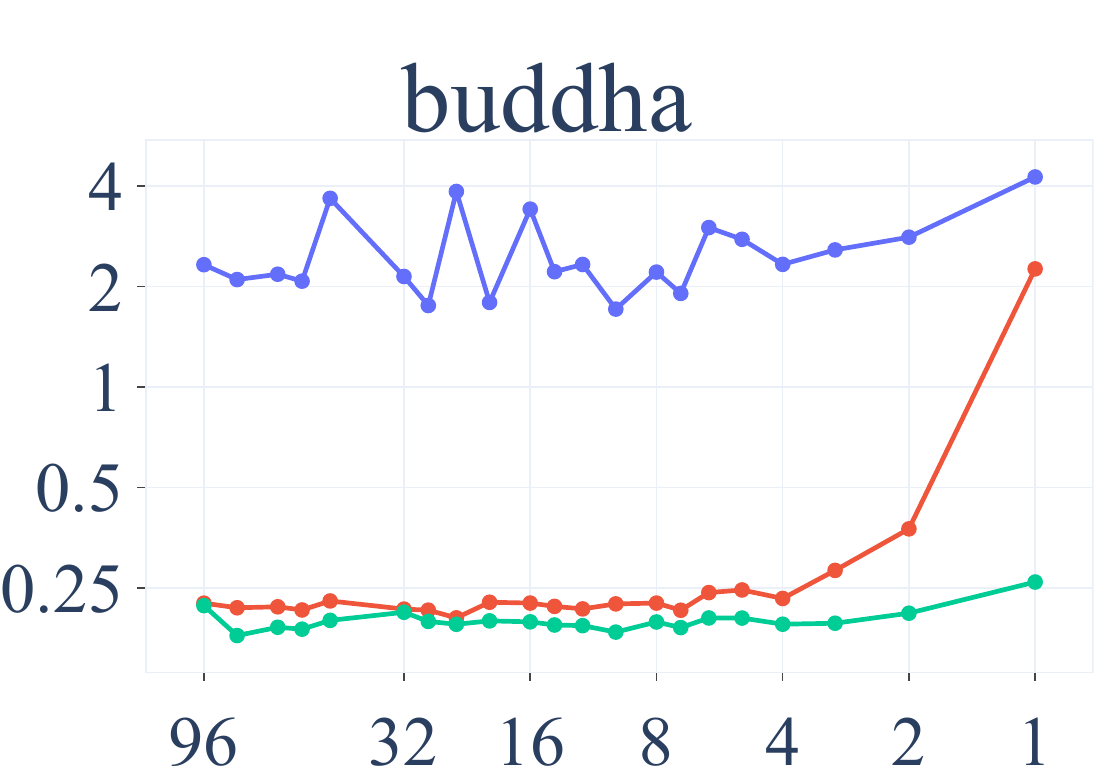}&
        \includegraphics[width=0.2\linewidth]{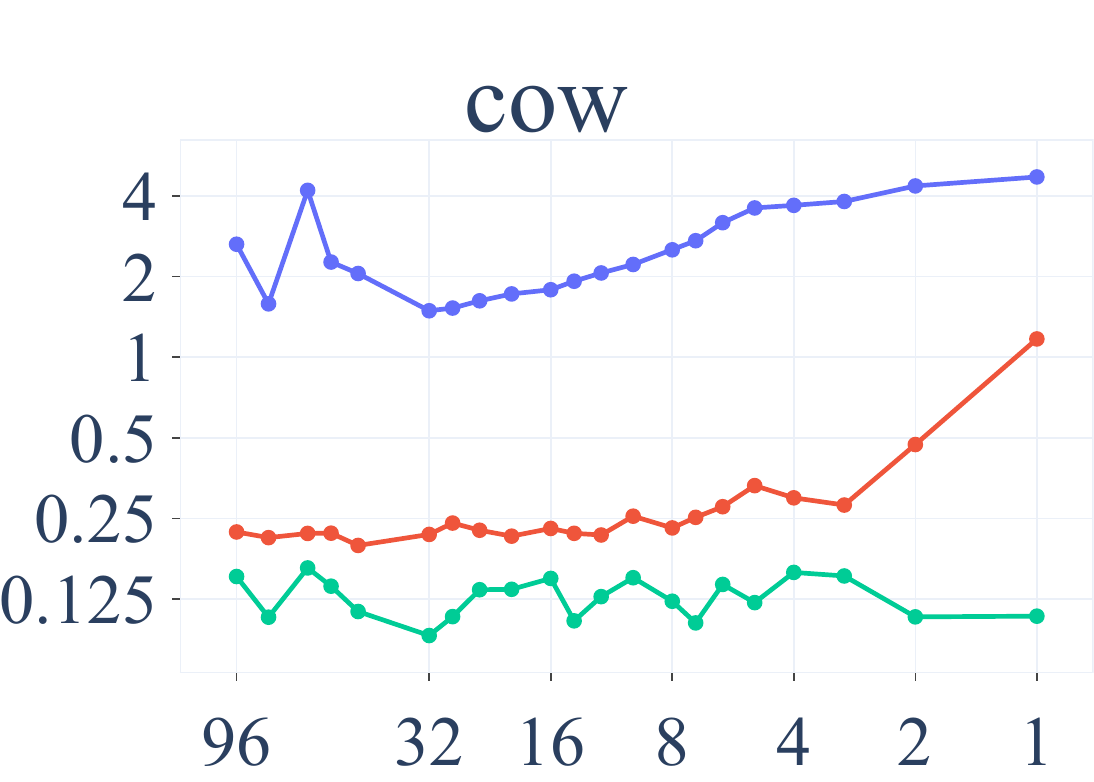}&
        \includegraphics[width=0.2\linewidth]{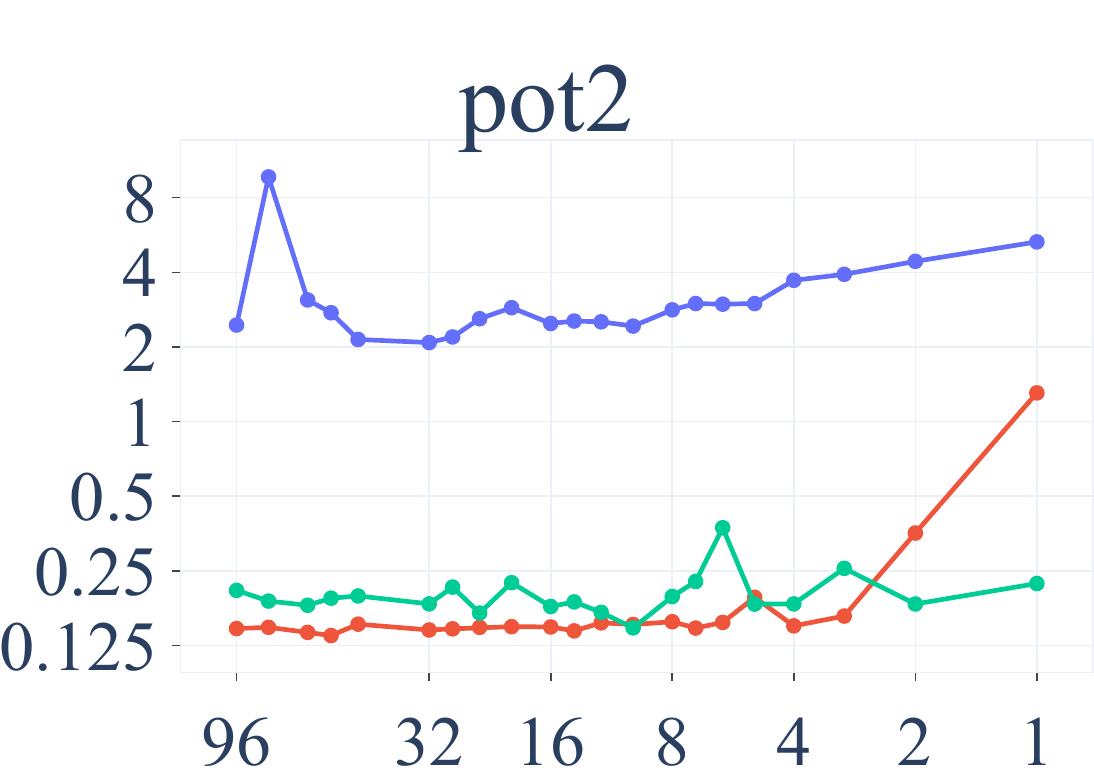}&
        \includegraphics[width=0.2\linewidth]{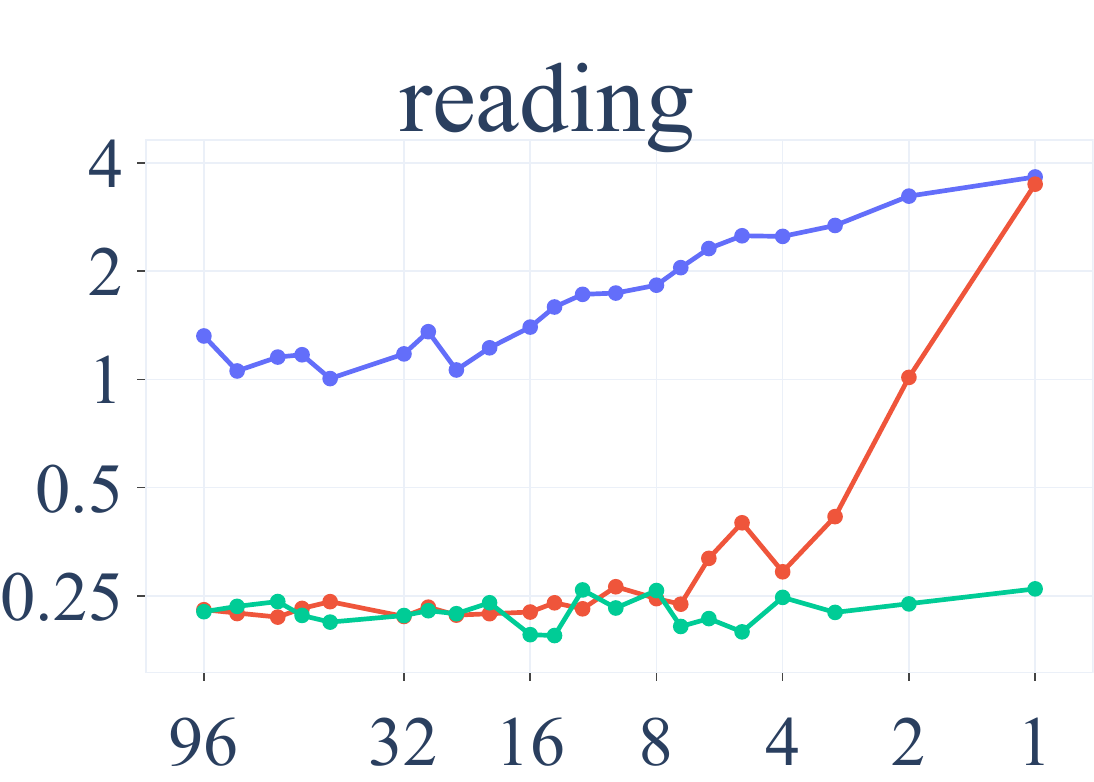} \\
        \includegraphics[width=0.2\linewidth]{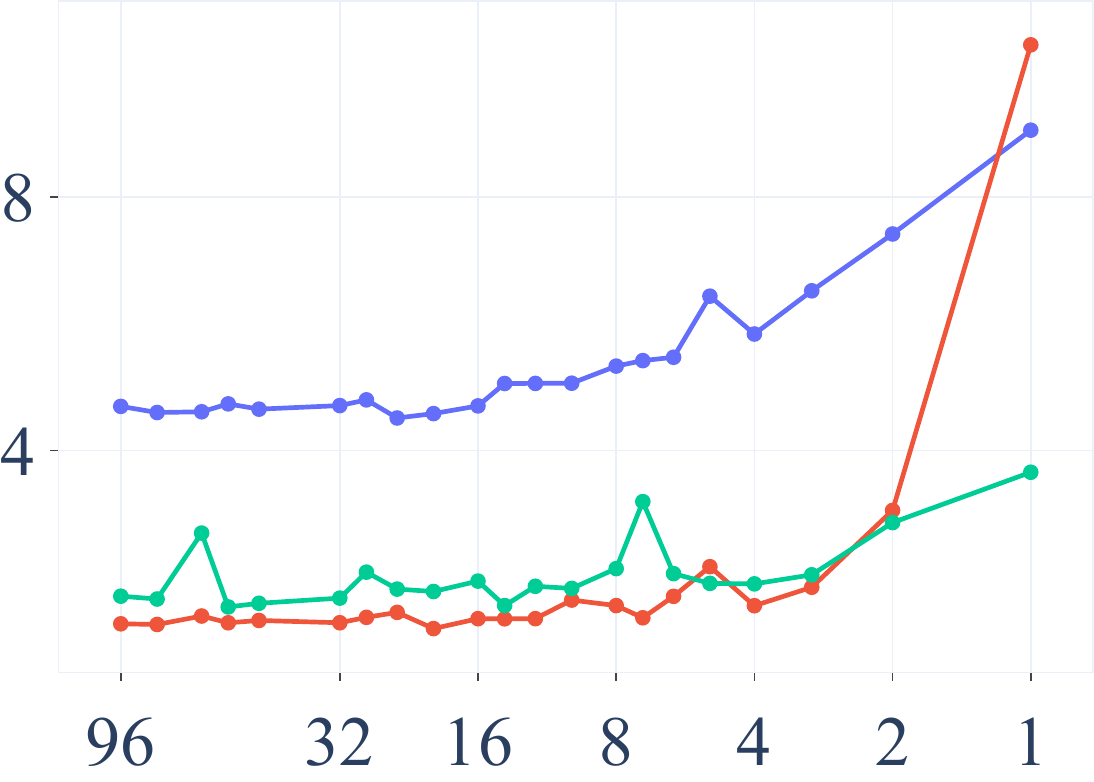}&
        \includegraphics[width=0.2\linewidth]{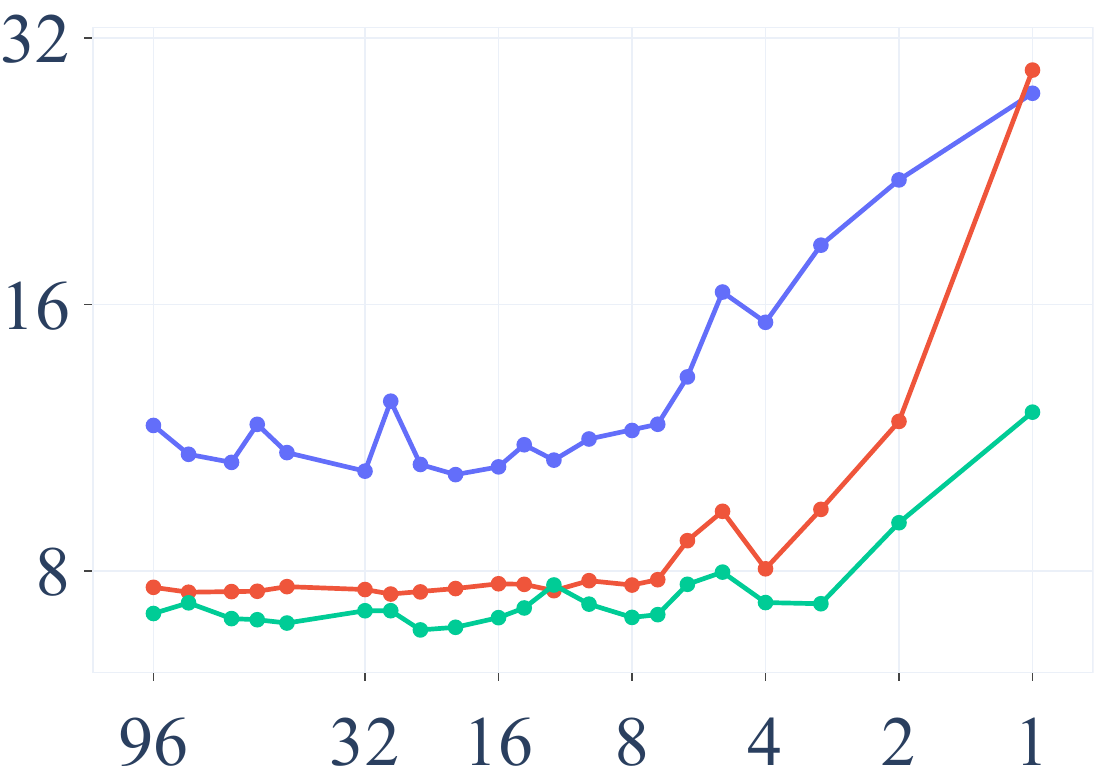}&
        \includegraphics[width=0.2\linewidth]{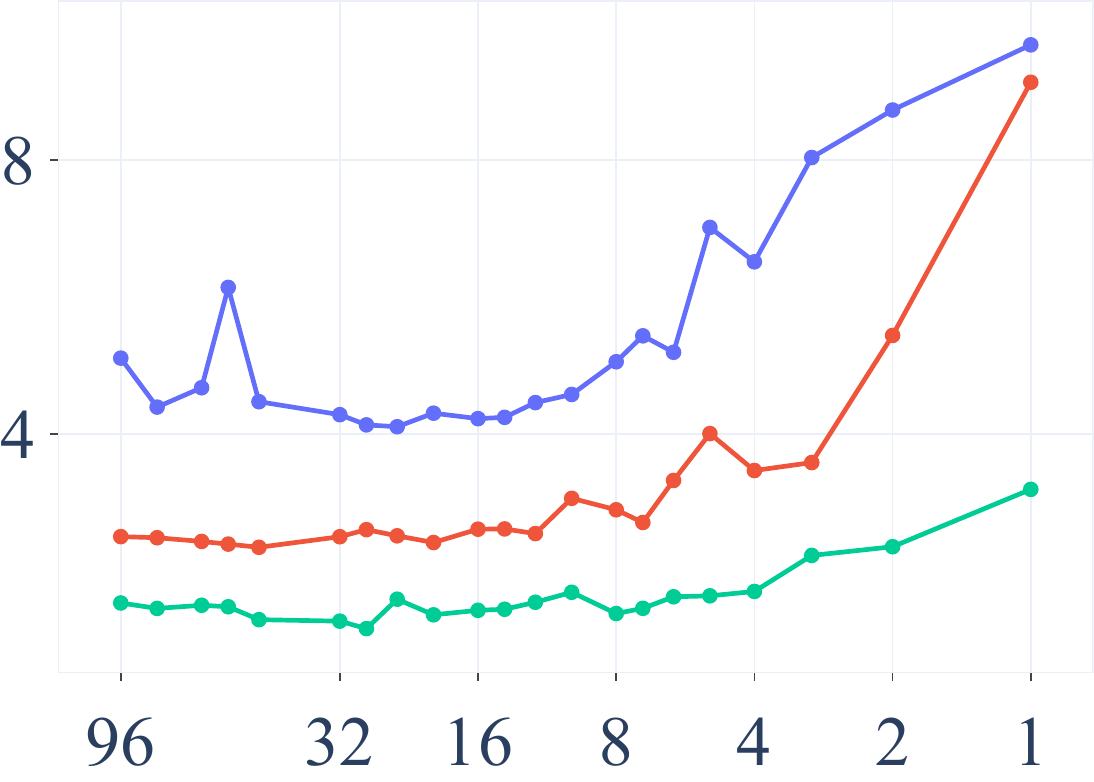}&
        \includegraphics[width=0.2\linewidth]{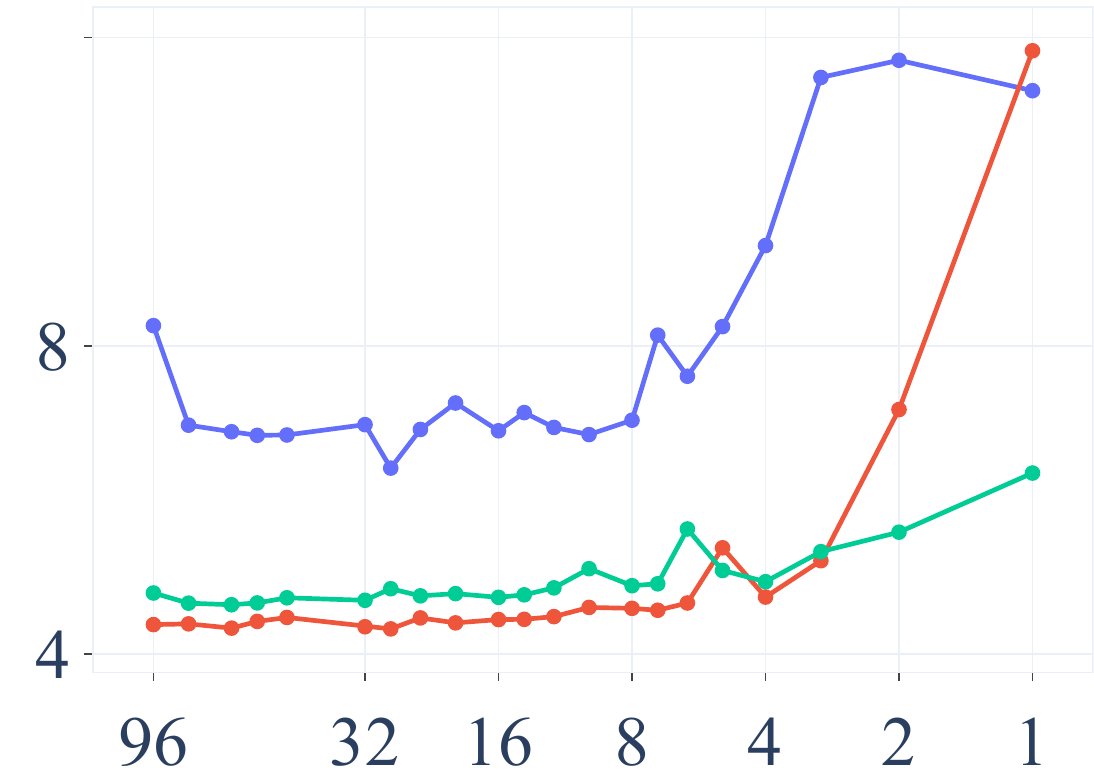}&
        \includegraphics[width=0.2\linewidth]{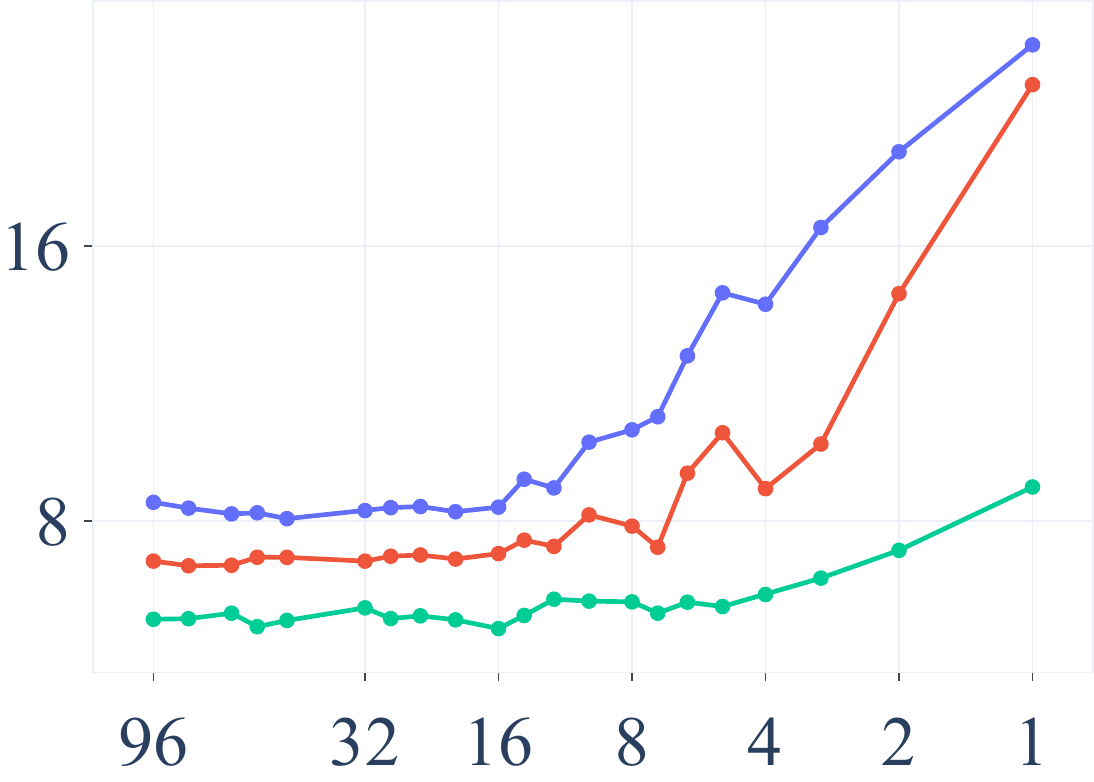} \\
        \multicolumn{5}{c}{\includegraphics[width=0.3\linewidth]{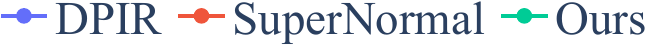}}
    \end{tabular}
    \vspace{-1em}
    \caption{
    \textbf{Quantitative evaluation of shape and normal maps. (Top)} Chamfer distance (in mm, lower is better). \textbf{(Bottom)} Normal MAE (in degrees, lower is better). Both axes are in $\log_2$ scale, with the x-axis indicating the number of directional lighting used per view.}
    \label{fig.geo_quan_eval}
    \vspace{-1em}
\end{figure*}

\begin{figure}
\scriptsize
    \centering
    \begin{tabular}{@{}c@{}c@{}c@{}c@{}c@{}c@{}c@{}c@{}}
     & \multicolumn{7}{c}{
            96 \hfill 32 \hfill 16 \hfill 4 \hfill 2
            \hfill 1 \hfill GT
        }
    \\
    \rotatebox{90}{\quad \quad \dpir} & \multicolumn{7}{c}{
    \includegraphics[width=0.95\linewidth]{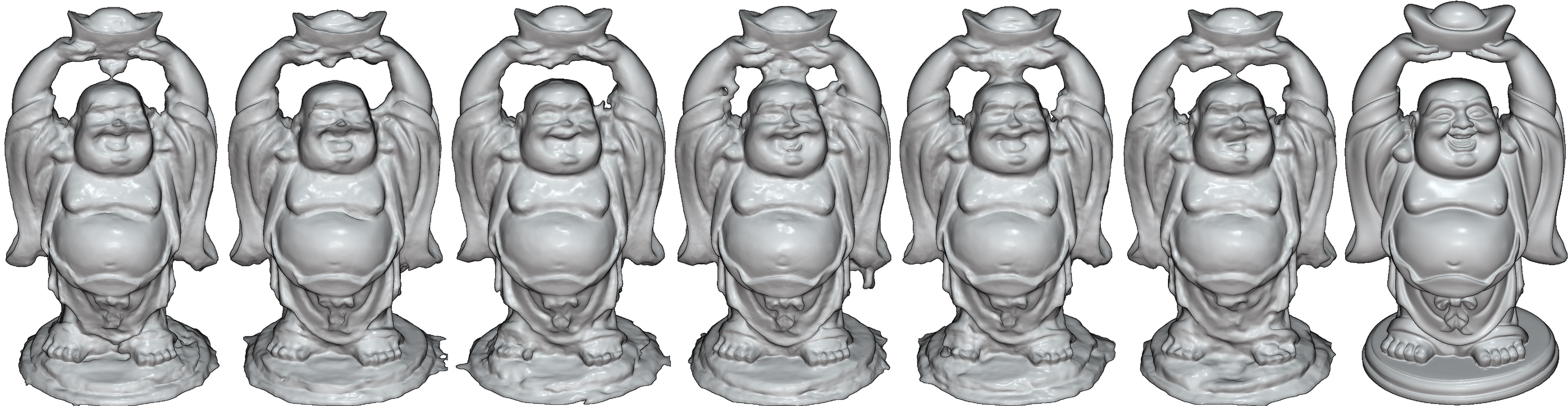}}
    \\
    \rotatebox{90}{\quad \supernormal} & \multicolumn{7}{c}{
    \includegraphics[width=0.95\linewidth]{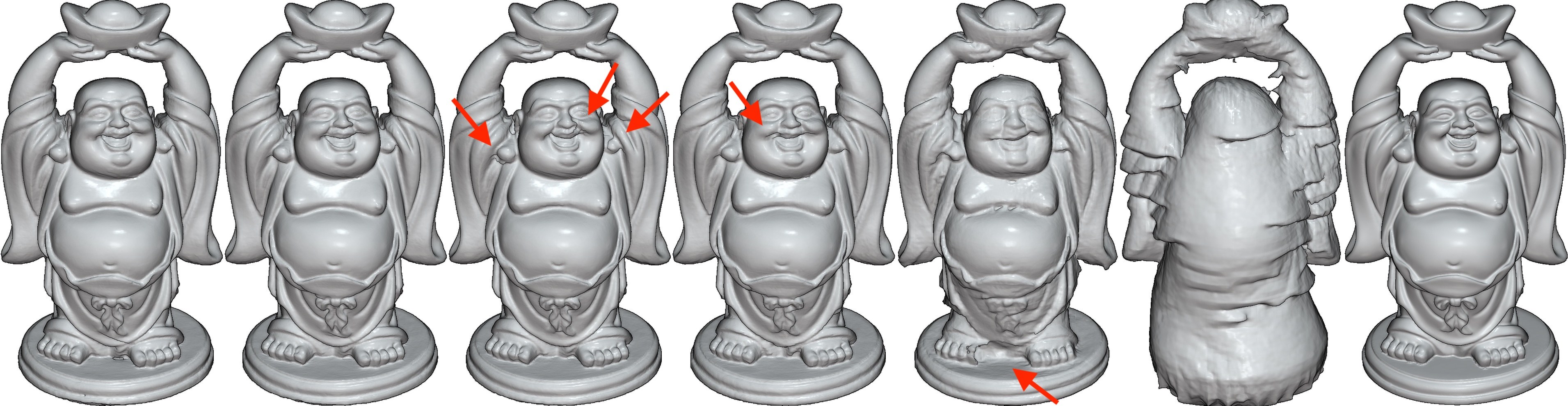}}
    \\
    \rotatebox{90}{\quad \quad \quad Ours} & \multicolumn{7}{c}{
    \includegraphics[width=0.95\linewidth]{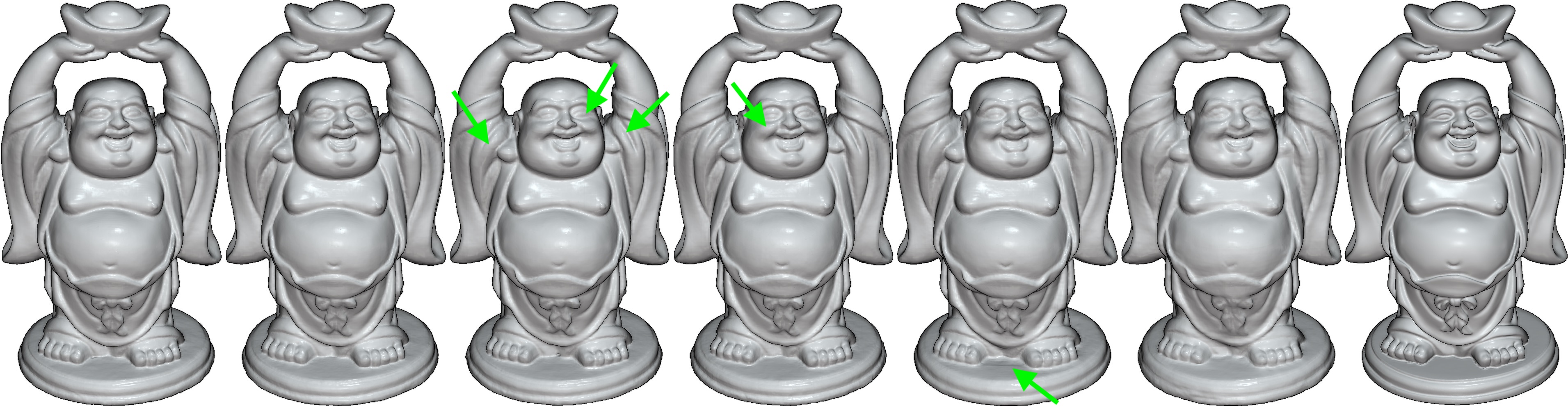}}
    \end{tabular}
\vspace{-1em}
    \caption{\textbf{Qualitative comparison of meshes.} 
    The numbers in the header row indicate the number of directional lights used per view.}
    \label{fig.mesh_qual_comparison}
    \vspace{-1em}
\end{figure}
\begin{figure}
\scriptsize
    \centering
    \begin{tabular}{@{}c@{}c@{}c@{}c@{}c@{}}
    \dpir & \sdmunips & \supernormal & Ours & GT \\
    \includegraphics[width=0.2\linewidth]{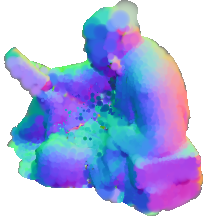} &
    \includegraphics[width=0.2\linewidth]{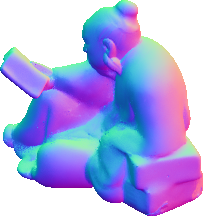} &
    \includegraphics[width=0.2\linewidth]{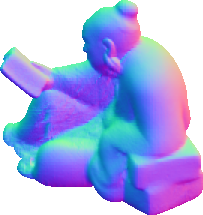} &
    \includegraphics[width=0.2\linewidth]{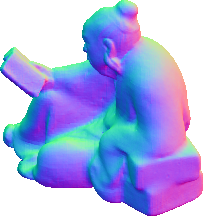} &
    \includegraphics[width=0.2\linewidth]{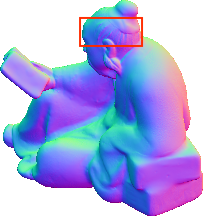} \\
    \includegraphics[width=0.2\linewidth]{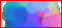} &
    \includegraphics[width=0.2\linewidth]{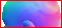} &
    \includegraphics[width=0.2\linewidth]{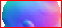} &
    \includegraphics[width=0.2\linewidth]{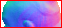} &
    \includegraphics[width=0.2\linewidth]{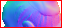} \\
    \includegraphics[width=0.2\linewidth]{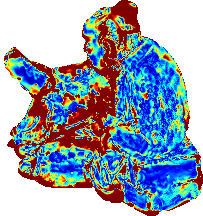} &
    \includegraphics[width=0.2\linewidth]{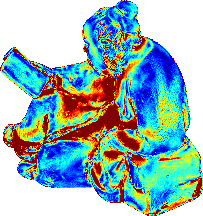} &
    \includegraphics[width=0.2\linewidth]{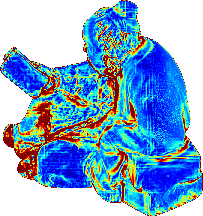} &
    \includegraphics[width=0.2\linewidth]{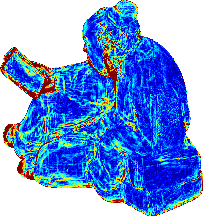} &
    \colorbartwo{0.1}{$\geq 20^\circ$} \\
    MAE:  $13.81^\circ$ & $10.96^\circ$ & $8.68^\circ$ & \textbf{$6.55^\circ$} &  \\
    \end{tabular}
    \vspace{-1.5em}
    \caption{\textbf{Normal accuracy.} Normal maps and angular error maps recovered from 4L20V input images.}
    \label{fig.normal_qual_comparison}
    \vspace{-1em}
\end{figure}

\vspace{0.5em}
\subsubsection{Geometry Reconstruction}
\label{sec.geo_eval}
We evaluate the geometry reconstruction quality against \dpir and \supernormal on the five objects in \diligentmv. 
We use all $20$ views as input for all methods while varying the number of OLAT images per view.
Lighting conditions are randomly sampled and kept consistent across views, and all methods use the same set of OLAT images as input.
\Cref{fig.geo_quan_eval} reports the L2 Chamfer distance (CD) of reconstructed meshes and mean angular error (MAE) of rendered normal maps under different numbers of lighting conditions.
\Cref{fig.mesh_qual_comparison} shows reconstructed meshes, and \cref{fig.normal_qual_comparison} displays normal and angular error maps.

\vspace{0.5em}
\noindent
\textbf{DPIR}~\cite{Chung_2024_CVPR} applies point-based rendering to reproduce the input images, where each point's normal vector is queried from a neural SDF.
\dpir enforces the consistency between the point set and the zero-level set of the neural SDF, but lacks a mechanism to propagate the estimated normals back into the neural SDF.
As a result, even though the normals can be optimized via photometric constraints, the loose coupling between geometry and photometric supervision leads to a noisy zero-level set (\cref{fig.mesh_qual_comparison}).
Moreover, the point-based rendering may introduce “disk-aliasing” artifacts (\cref{fig.normal_qual_comparison}, left column).

\vspace{0.5em}
\noindent
\textbf{SuperNormal}~\cite{Cao_2024_CVPR} is a stage-by-stage method that first estimates per-view normal maps from each OLAT stack~\cite{ikehata2023sdmunips} and then reconstructs a neural SDF whose rendered normals are consistent with PS normal maps.
Under dense lighting conditions, the PS method~\cite{ikehata2023sdmunips} produces high-quality normal maps, enabling \supernormal to achieve the best shape reconstruction quality on objects \textsc{Bear} and \textsc{Pot2}.
However, under sparse lighting (\eg, $<4$ light sources), errors in normal estimation significantly degrade the final reconstruction quality, as shown in \cref{fig.geo_quan_eval}.
Moreover, estimating each view's normal maps independently fails to enforce cross-view normal consistency, resulting in ``crack" artifacts in the reconstructed meshes (\cref{fig.mesh_qual_comparison}, red arrows).
\Cref{fig.normal_qual_comparison} further shows a normal map predicted by the PS method \sdmunips, which serves as input to \supernormal.
Despite improving overall normal accuracy (\cref{fig.normal_qual_comparison}, smaller MAE), fusing multi-view normal maps may smooth out high-frequency surface details when inconsistencies exist across views.

\vspace{0.5em}
\noindent
\textbf{Ours} achieves the best shape and normal reconstruction quality on the three objects \textsc{Buddha}, \textsc{Cow}, and \textsc{Reading}, regardless of the number of lighting conditions (\cref{fig.geo_quan_eval}).
In contrast to \dpir, our method tightly couples the geometry representation with photometric supervision through normals, enabling more accurate shape and normal estimation.
Benefiting from end-to-end optimization that exploits raw pixel measurements, our approach recovers finer surface details (\cref{fig.normal_qual_comparison}) and avoids the ``crack" artifacts present in \supernormal results (\cref{fig.mesh_qual_comparison}, red v.s. green arrows).

\subsubsection{Reflectance Recovery}
\begin{figure}
    \centering
\begin{tabular}{@{}c@{}c@{}}
\includegraphics[height=0.2\linewidth]{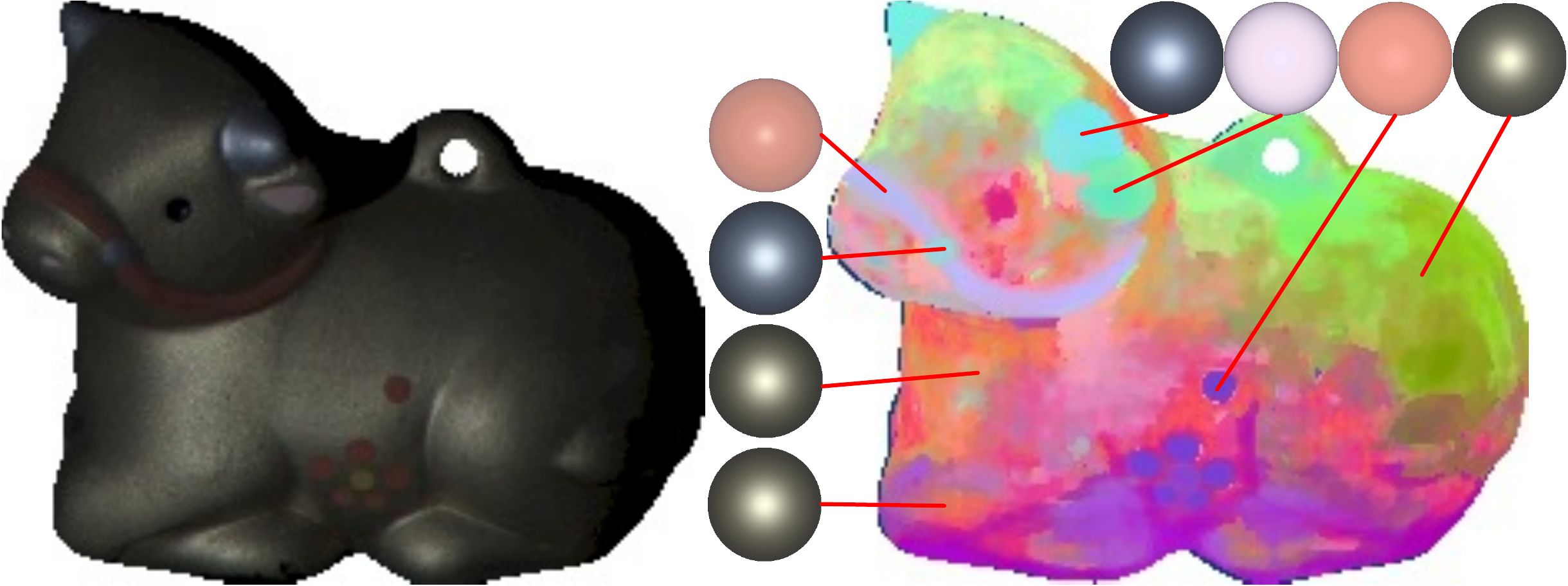} &
\includegraphics[height=0.2\linewidth]{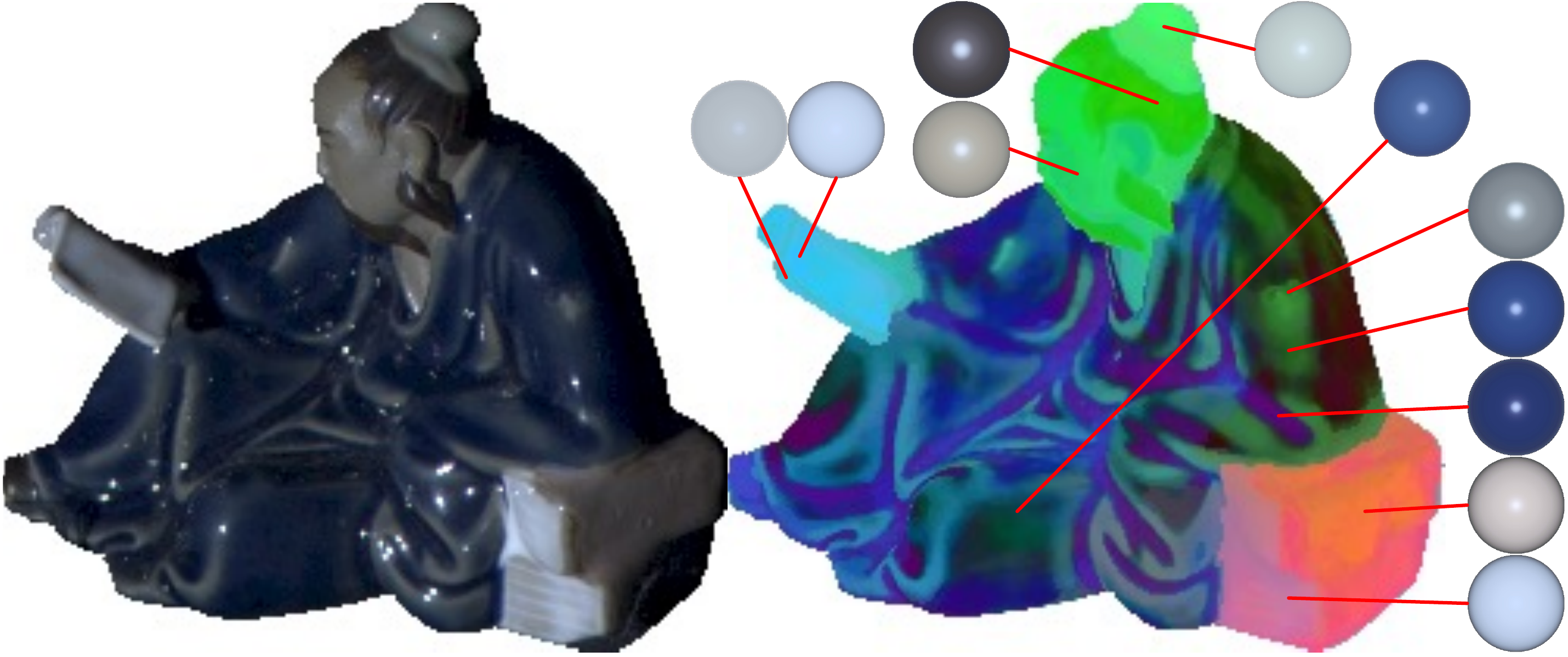} 
\end{tabular}
\vspace{-1em}
\caption{Visualization of BRDF latent maps and representative BRDF spheres for the objects \textsc{Cow} and \textsc{Reading}. Surface points with similar BRDFs are mapped to nearby latent codes.}
\vspace{-0.5em}
\label{fig.brdf_representative}
\end{figure}

\Cref{fig.brdf_representative} shows the BRDF latent map of the input image and representative BRDF spheres.
To visualize the BRDF latent codes, we reduce their dimensionality to $3$ and color-code them in RGB after training.
Thereby, similar colors indicate close latent vectors.
To render the BRDF sphere, we fix the viewing and light directions as $[0, 0, -1]^\top$ and query the BRDF values using normals from a hemisphere.
A full BRDF map displaying per-pixel BRDF spheres is provided in the supplementary, \cref{fig.brdf_map_reading}.

The BRDF latent map reveals that our method's effectiveness lies in how the spatial and BRDF MLP jointly uncover a low-dimensional reflectance structure from multi-view OLAT images.
The spatial MLP assigns similar latent codes to surface points with similar BRDFs,
enabling the BRDF MLP to learn one BRDF per latent shared across surface points, rather than fitting one BRDF per point with sparse measurements.

\begin{figure}
\newcommand{\figwidthps}{0.2}
\scriptsize
    \centering
    \begin{tabular}{@{}c@{}c@{}c@{}c@{}c@{}}
     \dpir-32 & \dpir-4 & Ours-32 & Ours-4 & GT \\
    \includegraphics[width=\figwidthps\linewidth]{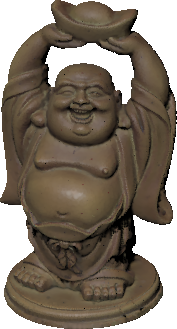} &
    \includegraphics[width=\figwidthps\linewidth]{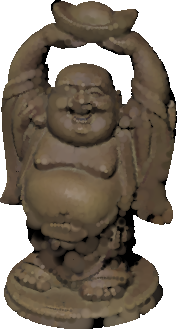} &
    \includegraphics[width=\figwidthps\linewidth]{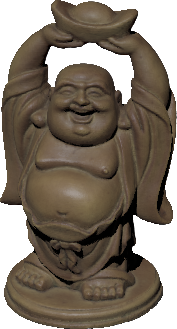} &
    \includegraphics[width=\figwidthps\linewidth]{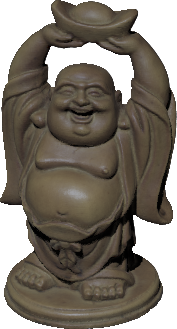} &
    \includegraphics[width=\figwidthps\linewidth]{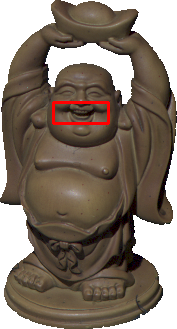} 
\\
\includegraphics[width=\figwidthps\linewidth]{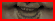} &
    \includegraphics[width=\figwidthps\linewidth]{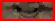} &
    \includegraphics[width=\figwidthps\linewidth]{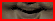} &
    \includegraphics[width=\figwidthps\linewidth]{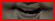} &
    \includegraphics[width=\figwidthps\linewidth]{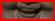} 
\\
PSNR: \textbf{35.94} & 32.14 & 29.20 & 28.95
\vspace{0.5em}
\\
\includegraphics[width=\figwidthps\linewidth]{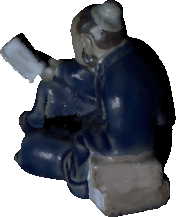} &
    \includegraphics[width=\figwidthps\linewidth]{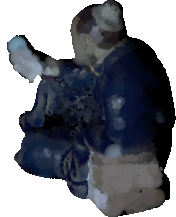} &
    \includegraphics[width=\figwidthps\linewidth]{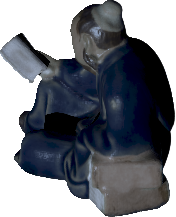} &
    \includegraphics[width=\figwidthps\linewidth]{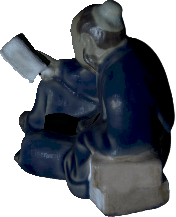} &
    \includegraphics[width=\figwidthps\linewidth]{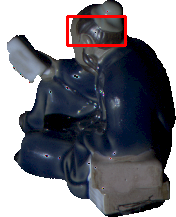} 
    \\
    \includegraphics[width=\figwidthps\linewidth]{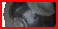} &
    \includegraphics[width=\figwidthps\linewidth]{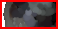} &
    \includegraphics[width=\figwidthps\linewidth]{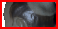} &
    \includegraphics[width=\figwidthps\linewidth]{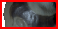} &
    \includegraphics[width=\figwidthps\linewidth]{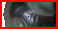} \\
    PSNR: \textbf{30.30} & 27.53 & 21.08 & 22.90 & 
    \end{tabular}
    \vspace{-1.5em}
    \caption{\textbf{Rendering quality evaluation.} Images are rendered from test viewpoints under a test light source. The suffix denotes the number of lighting conditions used per view for training.
    }
    \vspace{-1em}
\label{fig.novel_render_comparison}
\end{figure}

To quantitatively evaluate reflectance accuracy, we evaluate the images rendered at test viewpoints under test light sources.
\Cref{fig.novel_render_comparison} compares the test rendering quality against \dpir.
We train both methods using 32L18V or 4L18V input images, with the remaining views reserved for testing.

Although \dpir achieves a higher PSNR, our rendered images are perceptually more accurate.
This discrepancy arises because \dpir minimizes L2 loss, which favors PSNR, while our method minimizes weighted L1 loss, which is less correlated with PSNR.
Moreover, \dpir introduces disk-aliasing artifacts under sparse lighting conditions, and our method renders more realistic shadowed regions.

\subsubsection{Lighting Estimation}
\Cref{tab.light_direction_accuracy} quantitatively compares the light direction estimation accuracy of our method and \psnerf.
\psnerf uses a pre-trained network~\cite{chen2019SDPS_Net} to initialize light directions.
Nevertheless, our approach achieves a consistently lower MAE despite being optimized from scratch.

\begin{table}[]
    \centering
    \scriptsize
    \caption{MAE ($\downarrow$) of estimated light directions.}
    \vspace{-1em}
\resizebox{\linewidth}{!}{
    \begin{tabular}{@{}lc|ccccc@{}}
    \toprule
    Method & \#lights & \textsc{Bear} & \textsc{Buddha} & \textsc{Cow} & \textsc{Pot2} & \textsc{Reading}  \\
    \midrule
    \psnerf & 96 & 2.27 & 2.75 & 2.59&2.89& 4.26  \\
    \psnerf & 32 & 2.93 & 2.86 & 2.93 &3.34& 6.21   \\
    \psnerf & 4 & 3.25 & 2.74 & 4.79 &3.95& 5.89   \\
    \midrule
    Ours & 96 &1.87&1.89&2.51&2.34& 2.09  \\
    Ours & 32 & 2.11 & 1.61 &2.39& 1.73 & 1.22 \\
    Ours & 4  &2.51&1.92&1.99&1.64&1.16 \\
    Ours & 1 &2.18&2.49&2.02&1.26&1.16 \\
    \bottomrule
    \end{tabular}
  }  \label{tab.light_direction_accuracy}
  \vspace{-1em}
\end{table}

\subsubsection{Ablation Study}
\label{sec.ablation}
To evaluate the effectiveness of weighted L1 color loss and AE, we conduct an ablation study by removing AE and replacing the color loss with three alternatives: L2, weighted L2, and L1.
Without AE, we apply spherical harmonics encoding to the normal-view-light directions.

\begin{figure}
\tiny
    \centering
\begin{tabular}
{@{}c@{}c@{}c@{}c@{}c@{}c@{}c@{}}
L2 & W-L2 & L1 & W-L1 w/o AE & W-L1 w/ AE & GT & \\
\includegraphics[width=0.16\linewidth]{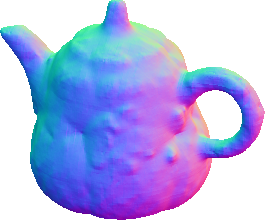} &
\includegraphics[width=0.16\linewidth]{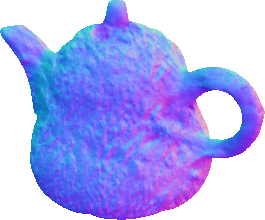} &
\includegraphics[width=0.16\linewidth]{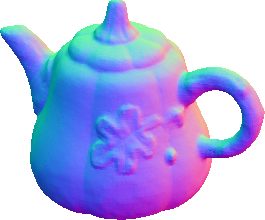} &
\includegraphics[width=0.16\linewidth]{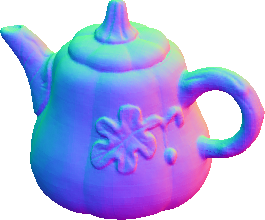} &
\includegraphics[width=0.16\linewidth]{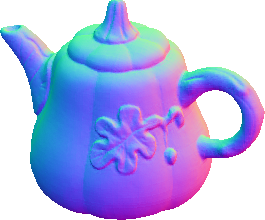} &
\includegraphics[width=0.16\linewidth]{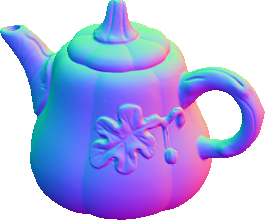} &
\\
\includegraphics[width=0.16\linewidth]{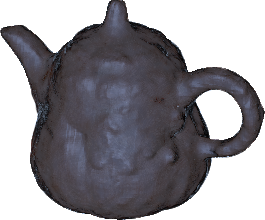} &
\includegraphics[width=0.16\linewidth]{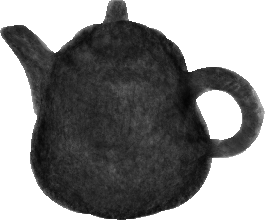} &
\includegraphics[width=0.16\linewidth]{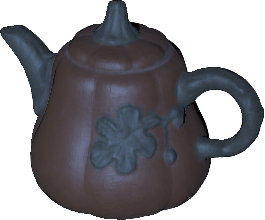} &
\includegraphics[width=0.16\linewidth]{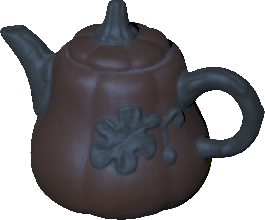} &
\includegraphics[width=0.16\linewidth]{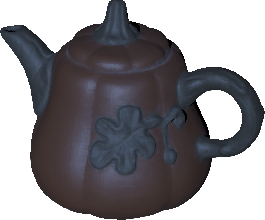} &
\includegraphics[width=0.16\linewidth]{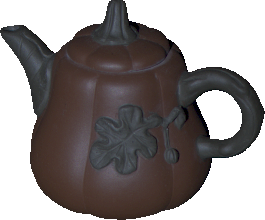} &
\\
\includegraphics[width=0.1\linewidth]{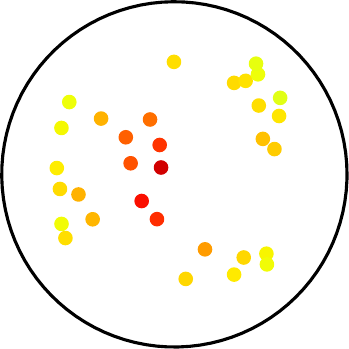} &
\includegraphics[width=0.1\linewidth]{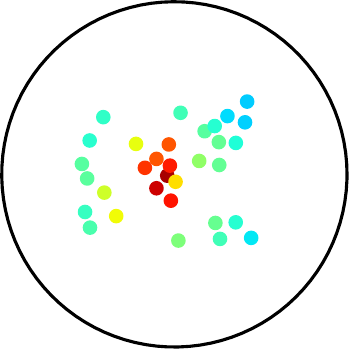} &
\includegraphics[width=0.1\linewidth]{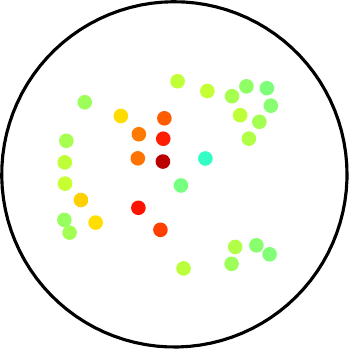} &
\includegraphics[width=0.1\linewidth]{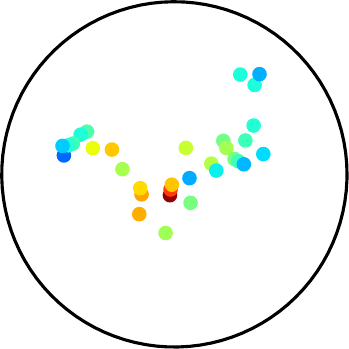} &
\includegraphics[width=0.1\linewidth]{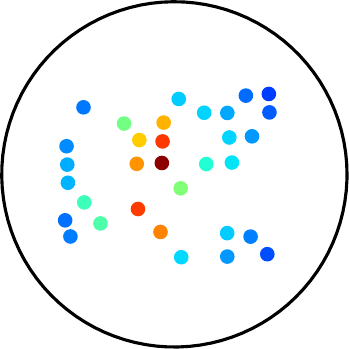} &
\includegraphics[width=0.1\linewidth]{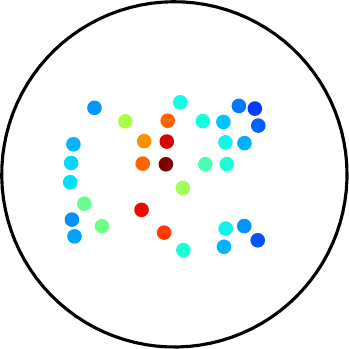}
&
\colorbartwo{0.05}{1}
\end{tabular} 
    \vspace{-1.5em}
\caption{\textbf{Qualitative results of ablation study.} ``W" denotes ``Weighted." Top to bottom: Normal maps, test-view test-light renderings, and estimated light directions and relative intensities.}
    \label{fig.ablation_qual}
    \vspace{-1em}
\end{figure}

\Cref{tab.ablation} reports quantitative evaluations, and \cref{fig.ablation_qual} presents qualitative results using 32L18V input images of \textsc{Pot2}.
To visualize light directions and intensities, we scatter plot the $x$ and $y$ components of the unit light directions,  with each point color-coded by its relative light intensity.
The maximum light intensity is normalized to $1$.
The results indicate that the combination of the weighted L1 loss and AE enhances overall reconstruction quality.
Removing AE increases light direction error from $2.67^\circ$ to over $16^\circ$.
Although using the L2 loss achieves the second-best training PSNR, the perceptual quality remains inferior, as minimizing the L2 loss is correlated to maximizing PSNR but does not guarantee perceptual fidelity.

\begin{table}[]
    \centering
\normalsize
    \caption{\textbf{Ablation study.}}
    \vspace{-1em}
    \resizebox{\linewidth}{!}{
    \begin{tabular}{@{}lc|ccccccc@{}}  
    \toprule
    Color Loss & AE & Mesh CD $\downarrow$ & Normal MAE $\downarrow$ &PSNR (Train) $\uparrow$ & PSNR (Test) $\uparrow$ & Light MAE $\downarrow$ &
    Light MSE $\downarrow$ \\
    \midrule
    L2 & \cmark & 0.38 & 11.45 & 29.68& 25.31& 11.04 & 0.398\\
    Weighted-L2 & \cmark & 0.67  & 24.97 & 24.27 & 21.85 & 5.19 & 0.167 \\
    L1 & \cmark & 0.27 & 6.41  & 24.67 & 23.84 & 5.34 & 0.351 \\
    Weighted-L1 & \xmark & 0.24& 5.66 & 26.87 & 24.98& 16.79 & 0.269 \\
    \midrule
    Weighted-L1 & \cmark & \textbf{0.16} & \textbf{4.86} & \textbf{35.38} & \textbf{32.40}  & \textbf{2.67} & \textbf{0.037} \\
    \bottomrule
    \vspace{-3.5em}
    \end{tabular}
    }
    \label{tab.ablation}
\end{table}
\begin{figure}
    \centering
\includegraphics[width=\linewidth]{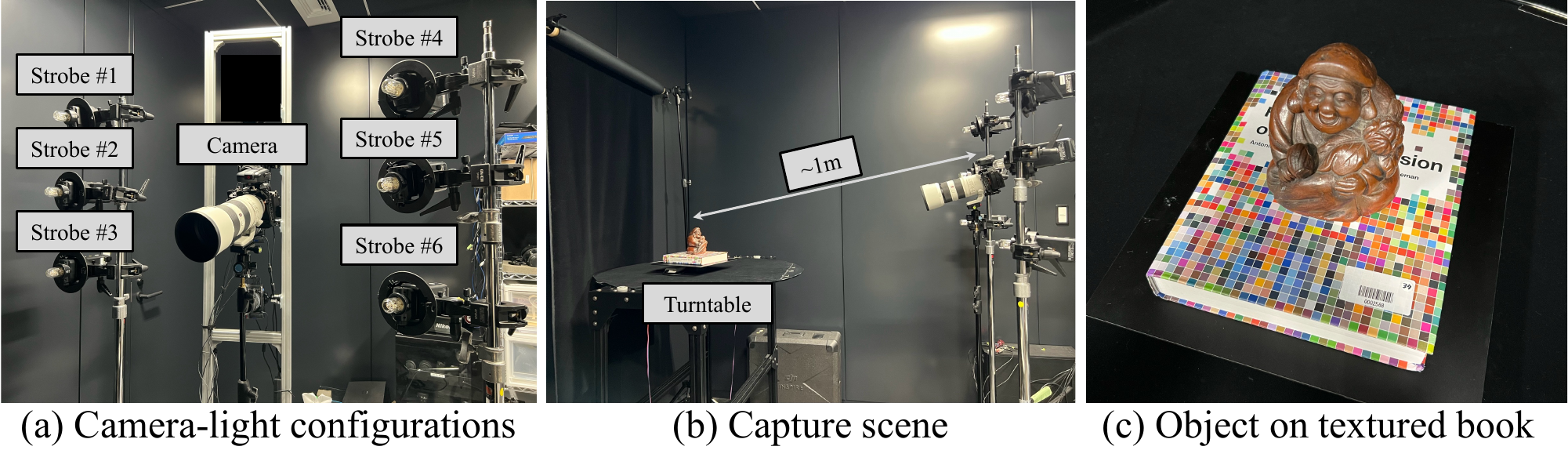}
    \vspace{-2em}
    \caption{\textbf{Our capture setup.} \textbf{(a)} A DSLR camera and six strobes are mounted on tripods, being fixed to each other.
    \textbf{(b)} The centimeter-scale object is positioned on a turntable approximately \SI{1}{m} from the strobes. \textbf{(c)} A textured book is placed beneath the object to assist with camera calibration.}
    \label{fig.capture_setup_real}
    \vspace{-1em}
\end{figure}

\begin{figure*}
\newcommand{\figwidthR}{0.128}
\newcommand{\figwidthBRDF}{0.024}
\newcommand{\figwidthG}{0.128}
\scriptsize
    \centering
    \begin{tabular}{@{}c@{}c@{}c@{}c@{}c@{}c@{}c@{}c@{}c@{}c@{}}
Shape & Normal &  \textbf{Capture} & Rendering (Shadowed) & Unshadowed & BRDF & Shadow map & & Relighting  & Lighting \\
\includegraphics[width=\figwidthR\linewidth]{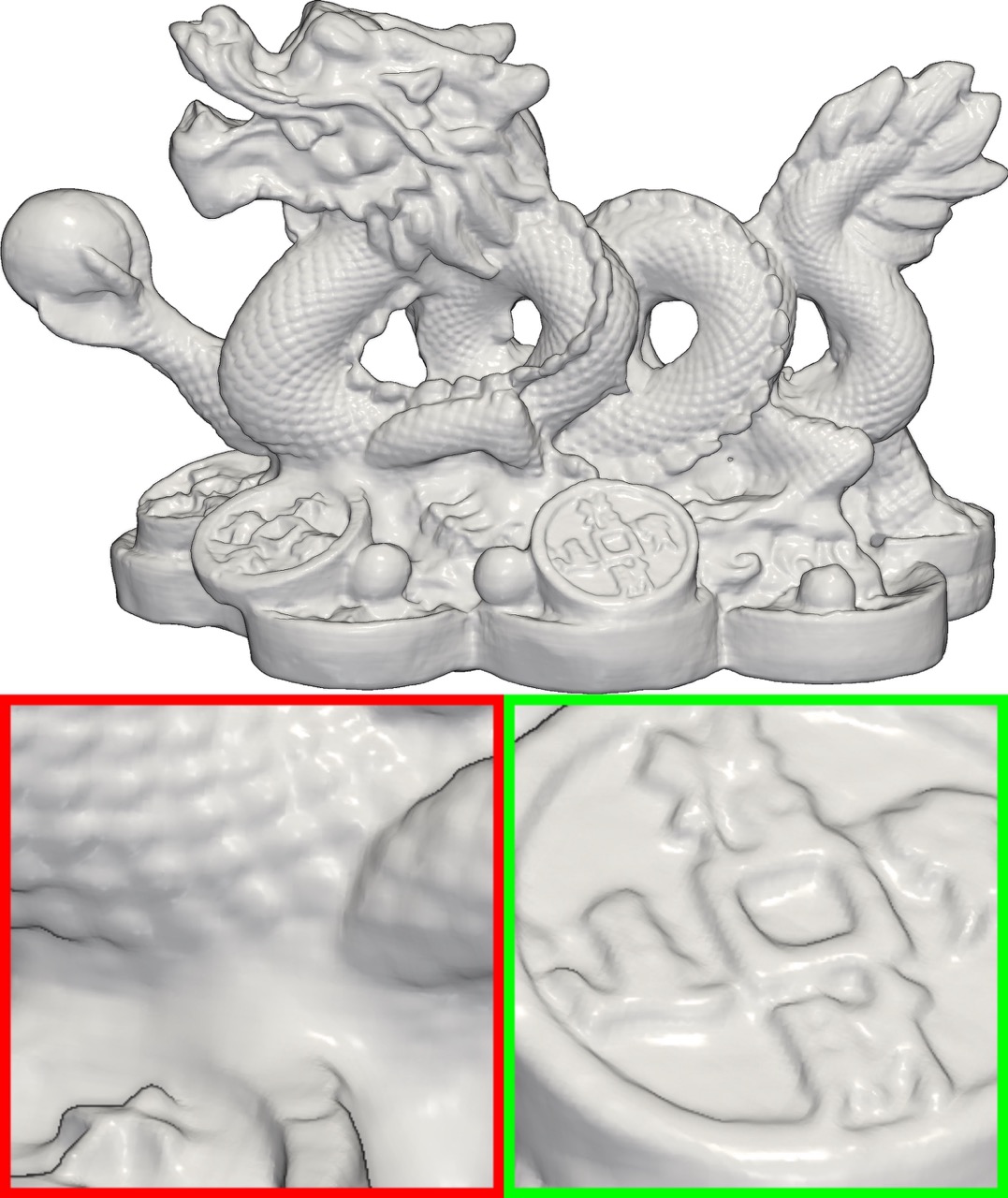} & 
\includegraphics[width=\figwidthR\linewidth]{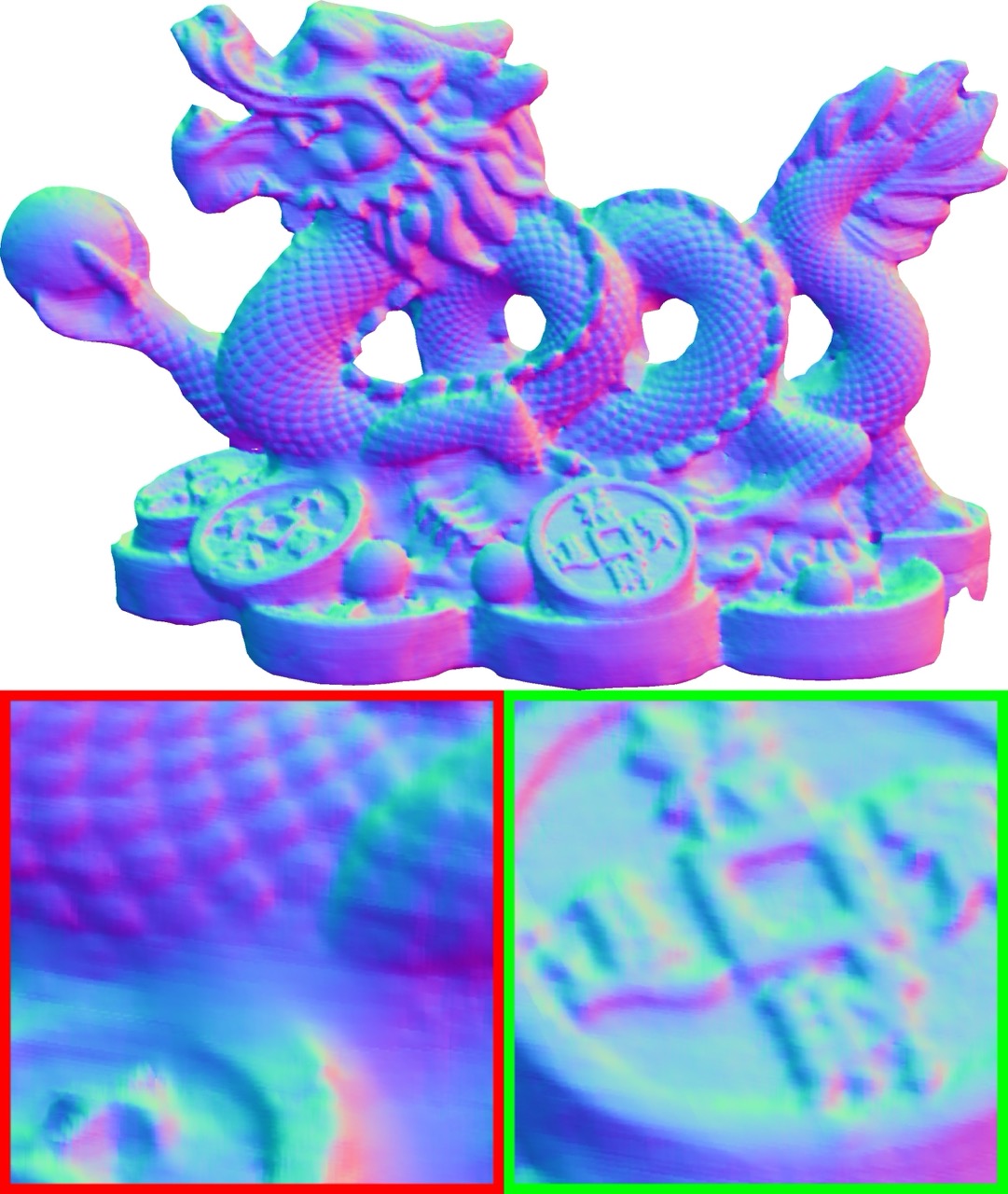} &
\includegraphics[width=\figwidthR\linewidth]{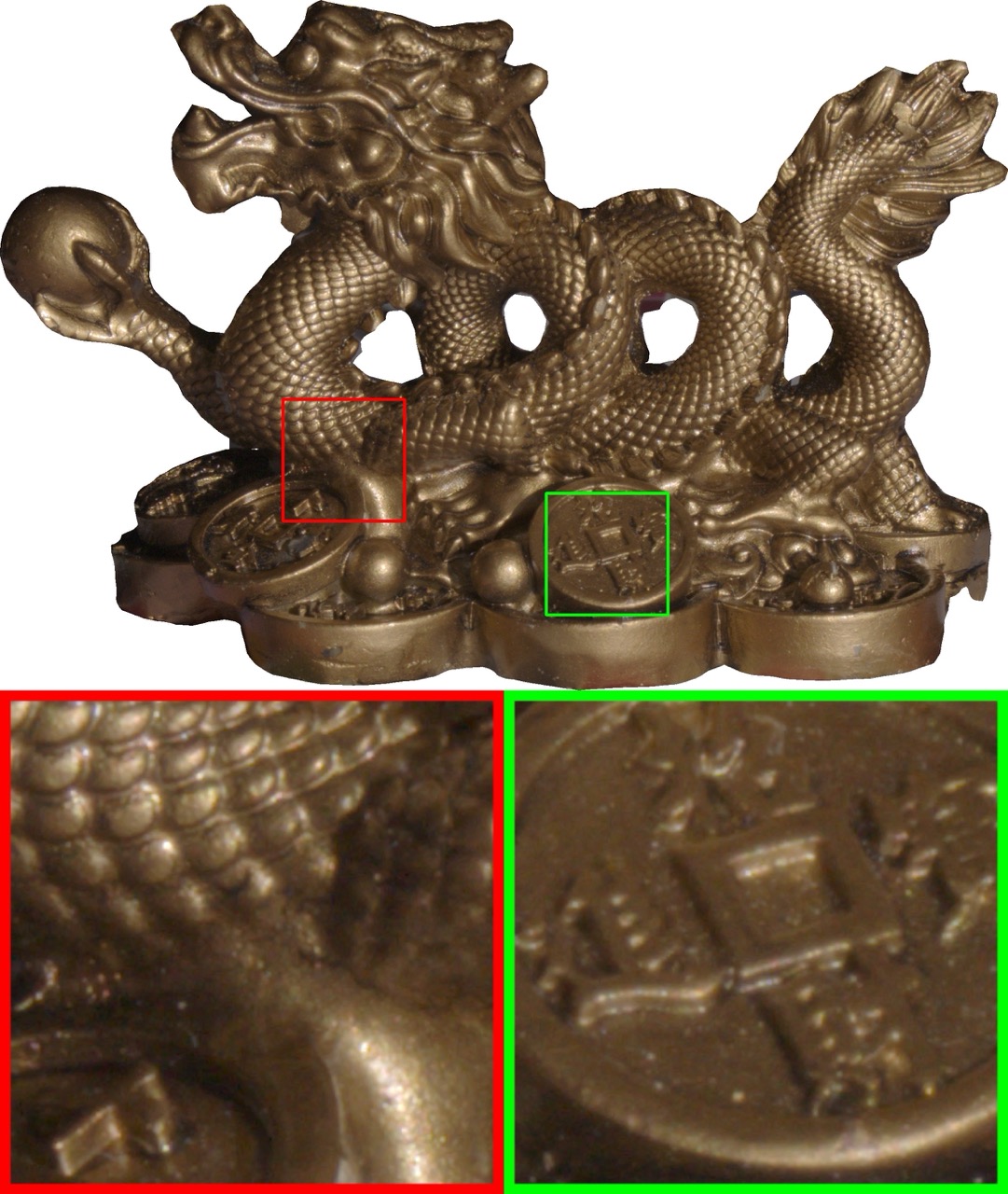} &
\includegraphics[width=\figwidthR\linewidth]{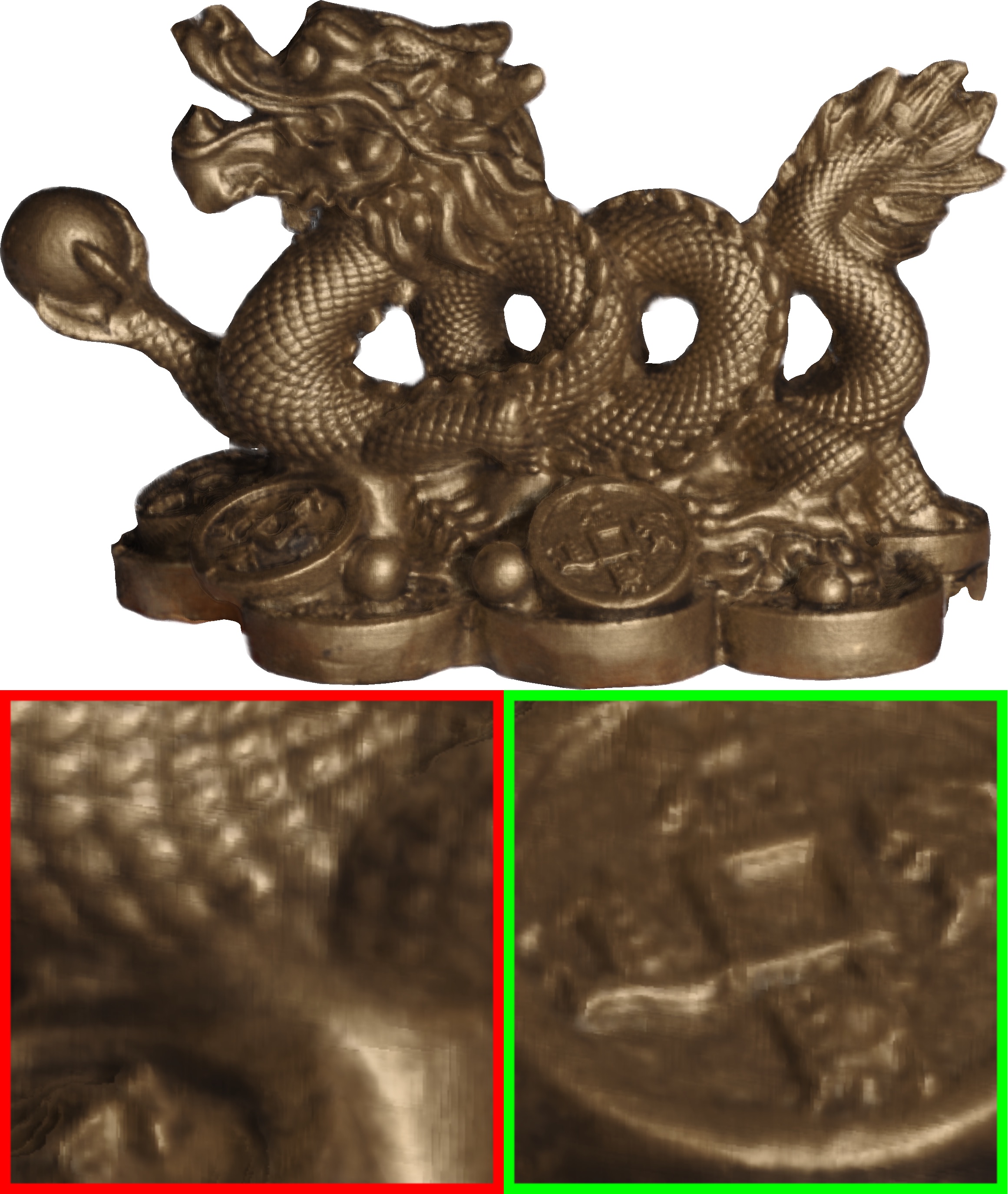} &
\includegraphics[width=\figwidthR\linewidth]{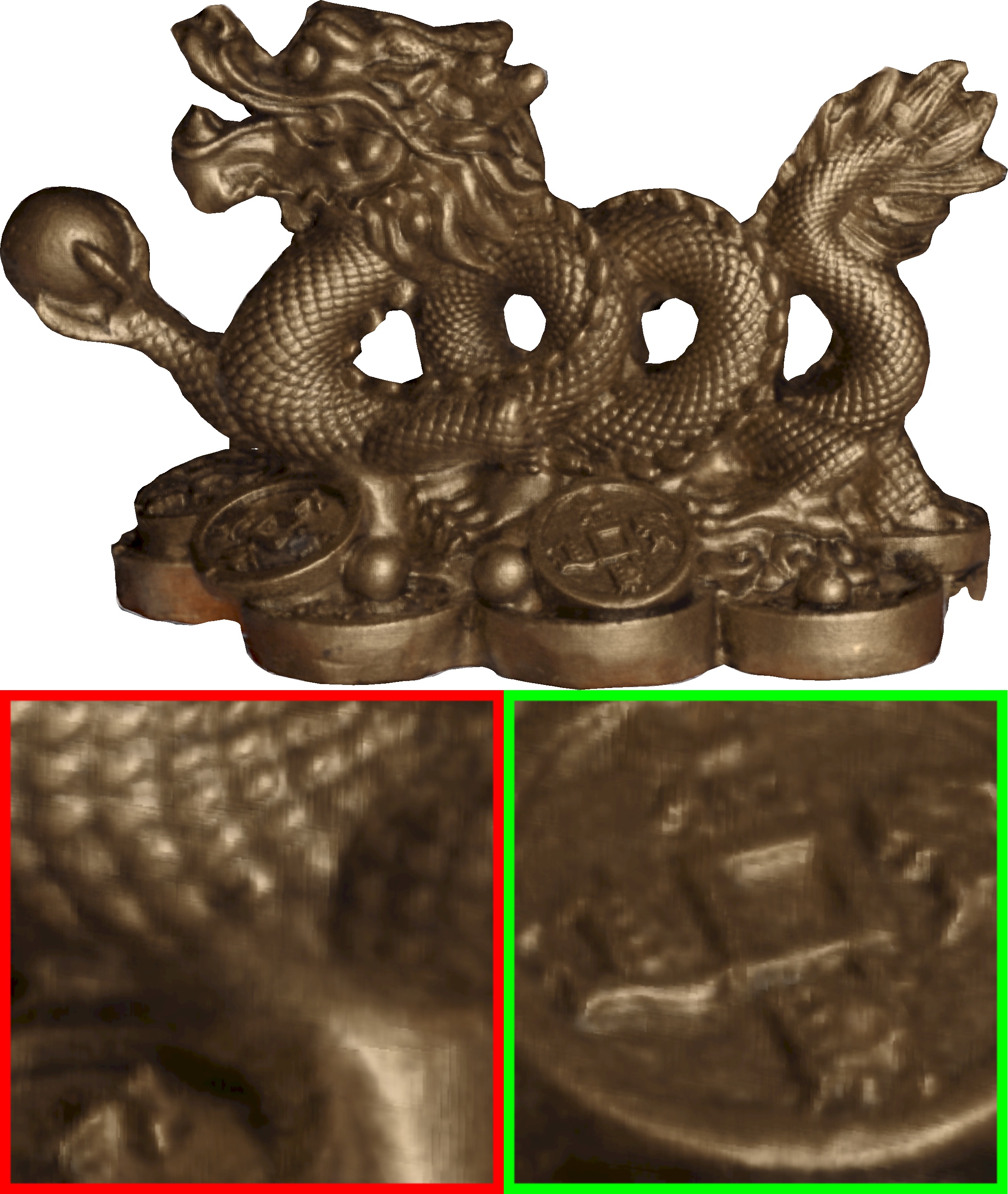} &
\begin{tabular}[b]{@{}c@{}}
  \includegraphics[width=\figwidthBRDF\linewidth]{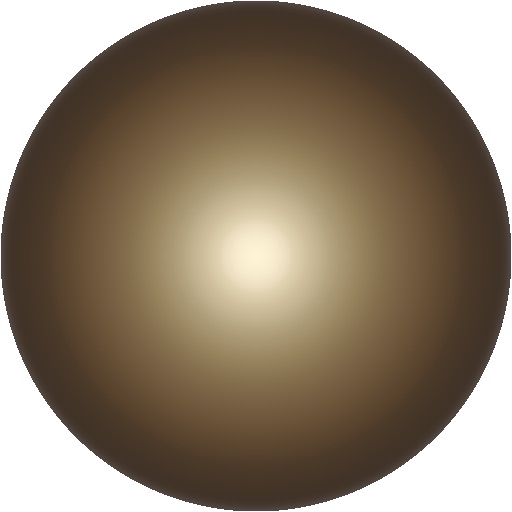}     \\
   \includegraphics[width=\figwidthBRDF\linewidth]{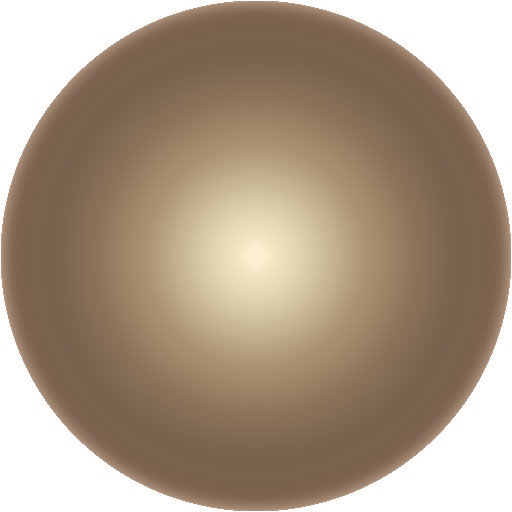} 
\end{tabular}&
\includegraphics[width=\figwidthR\linewidth]{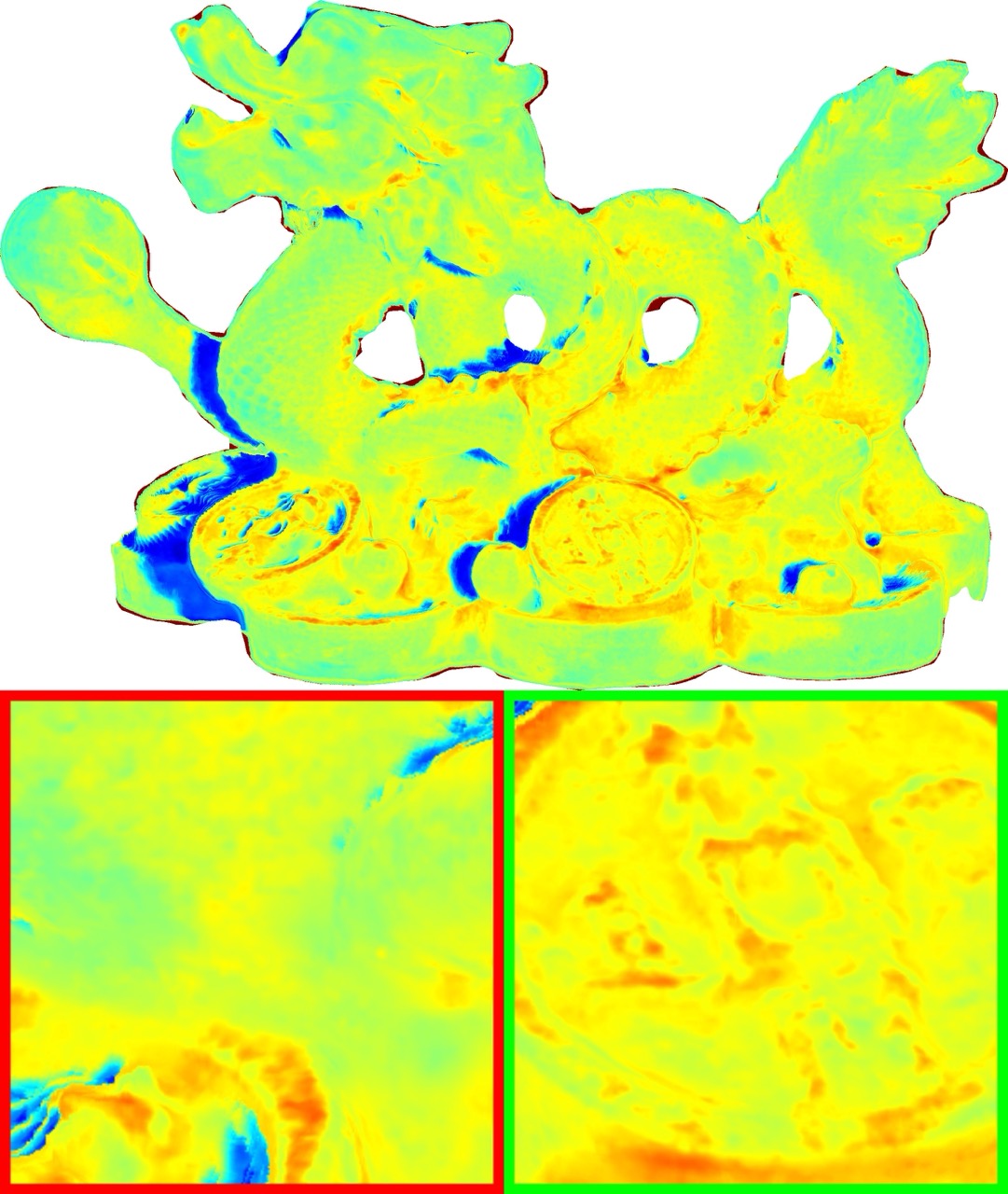} &
\colorbartwo{0.04}{1} &
\includegraphics[width=\figwidthR\linewidth]{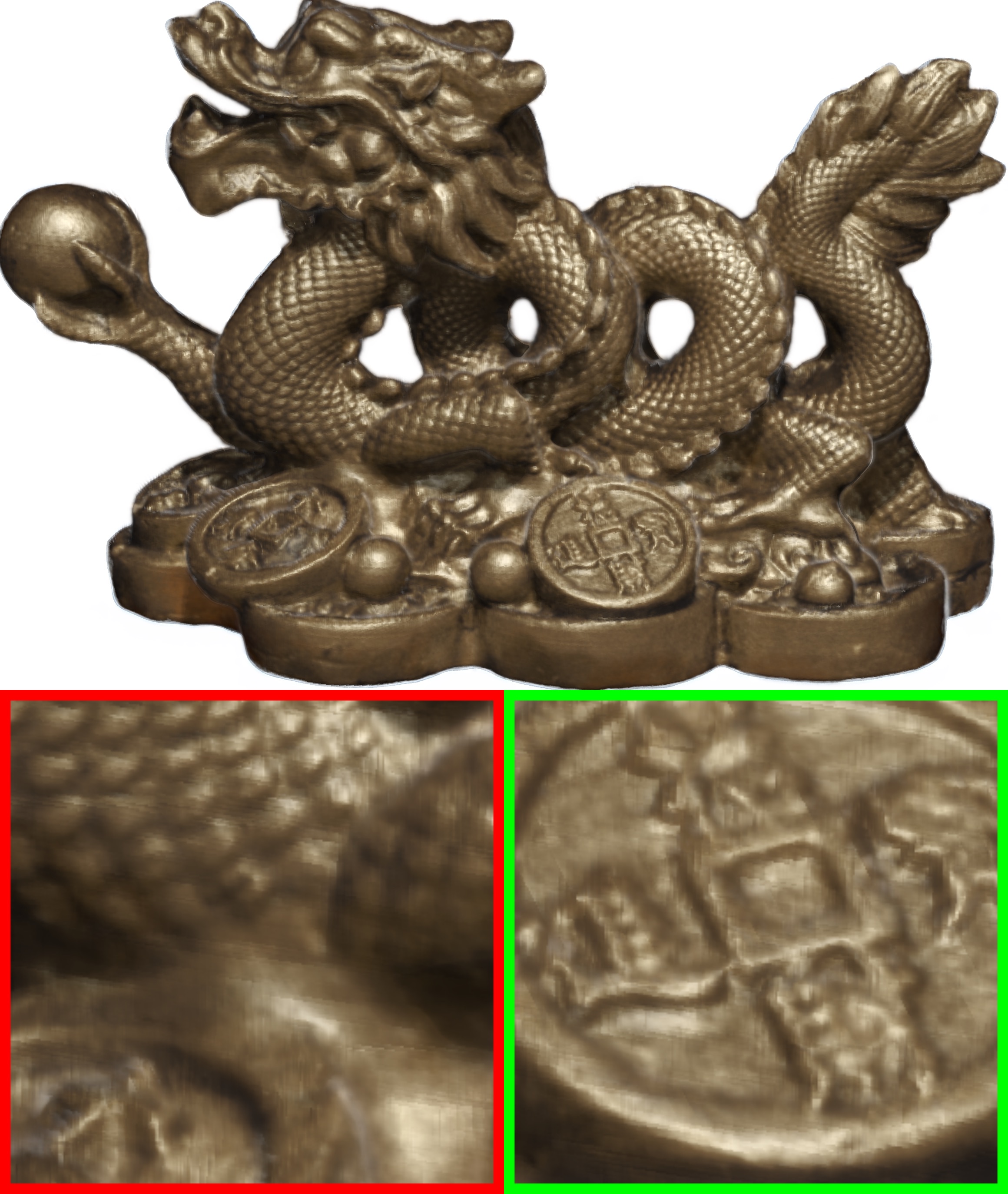} &
\includegraphics[width=0.04\linewidth]{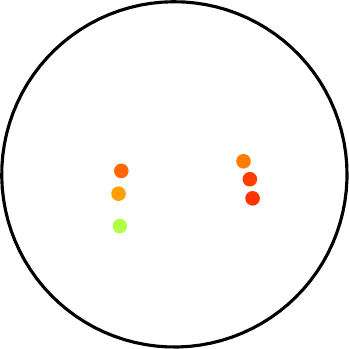}
\vspace{0.5em}
\\
\includegraphics[width=\figwidthG\linewidth]{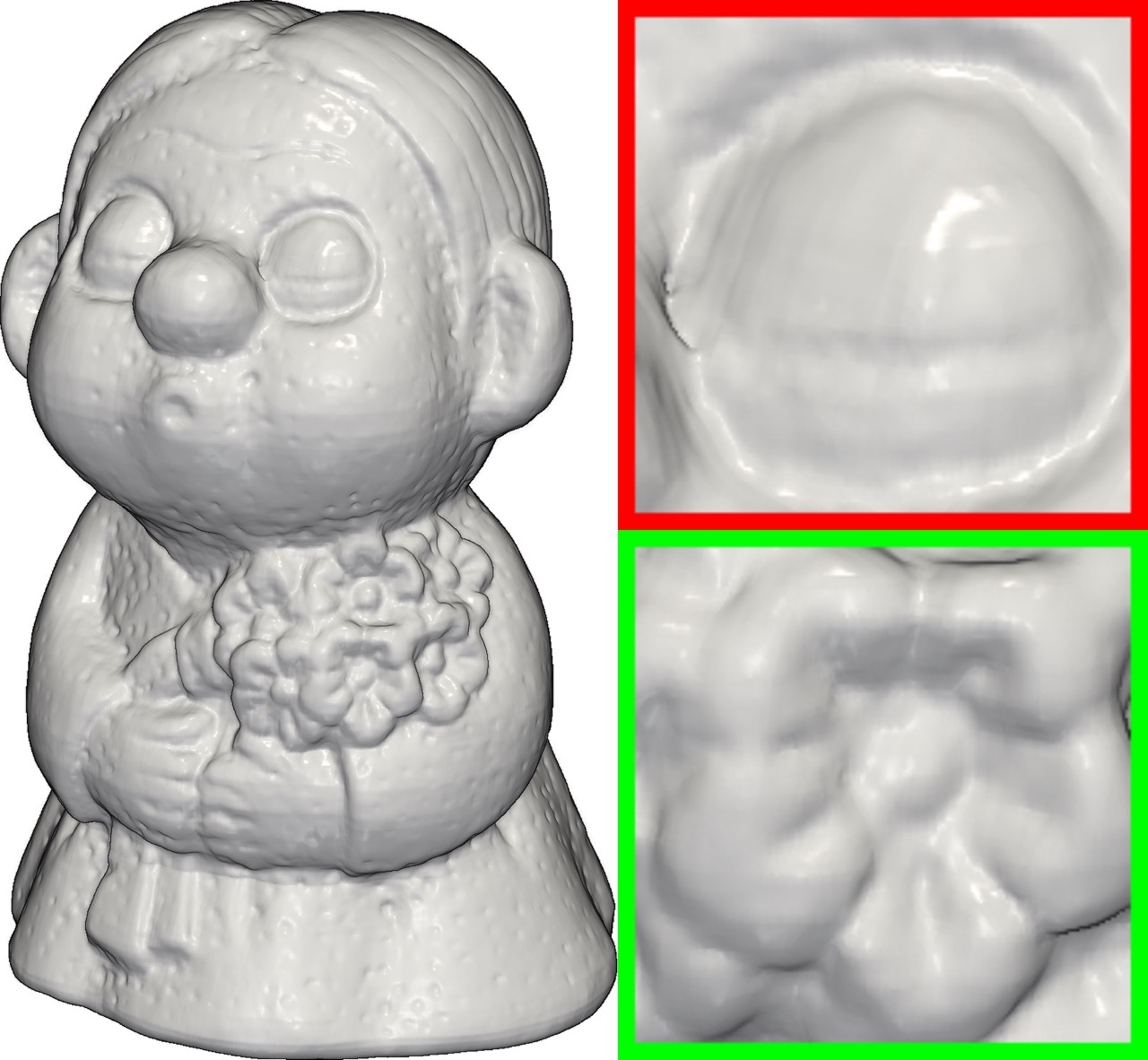} &
\includegraphics[width=\figwidthG\linewidth]{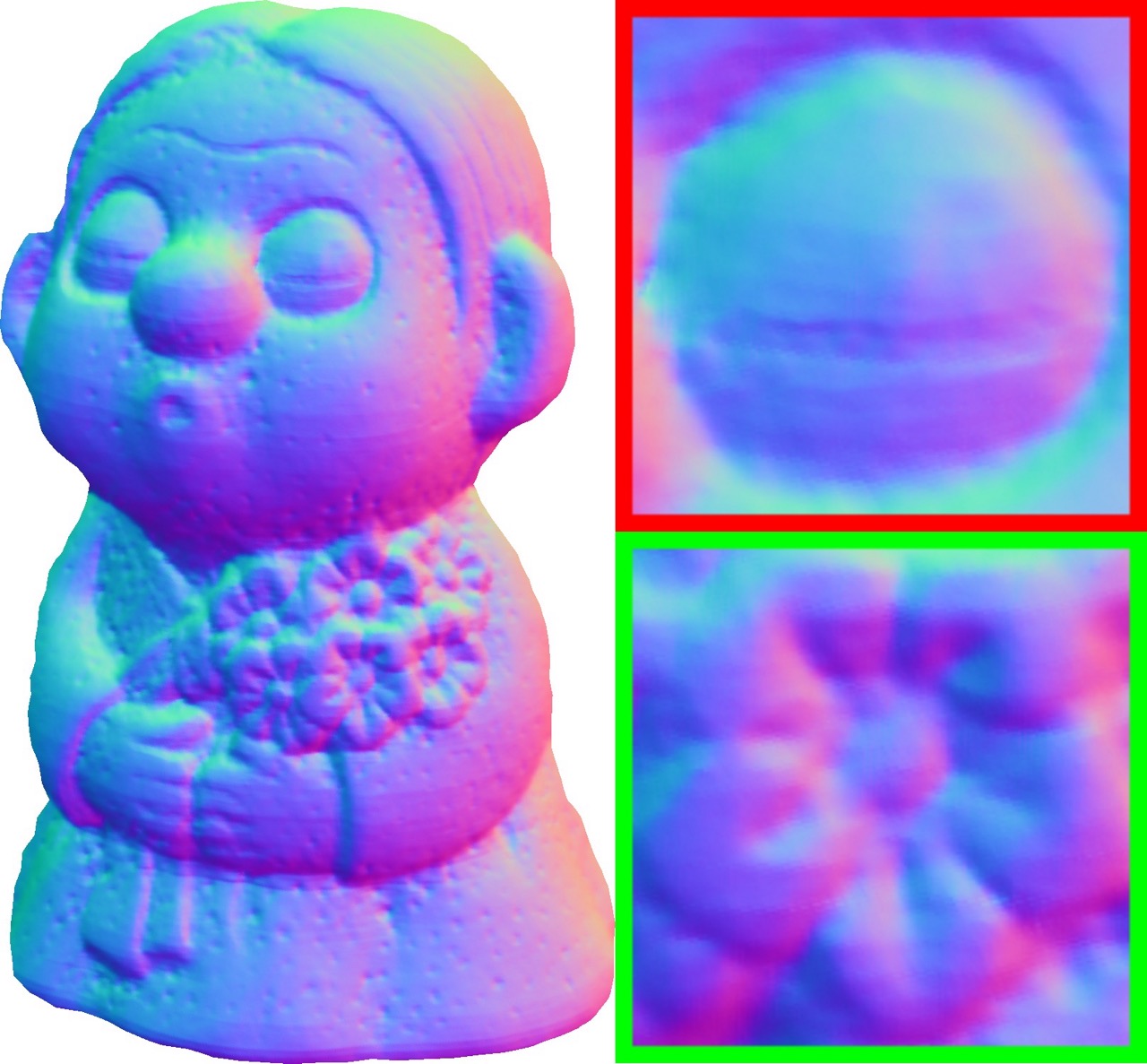} &
\includegraphics[width=\figwidthG\linewidth]{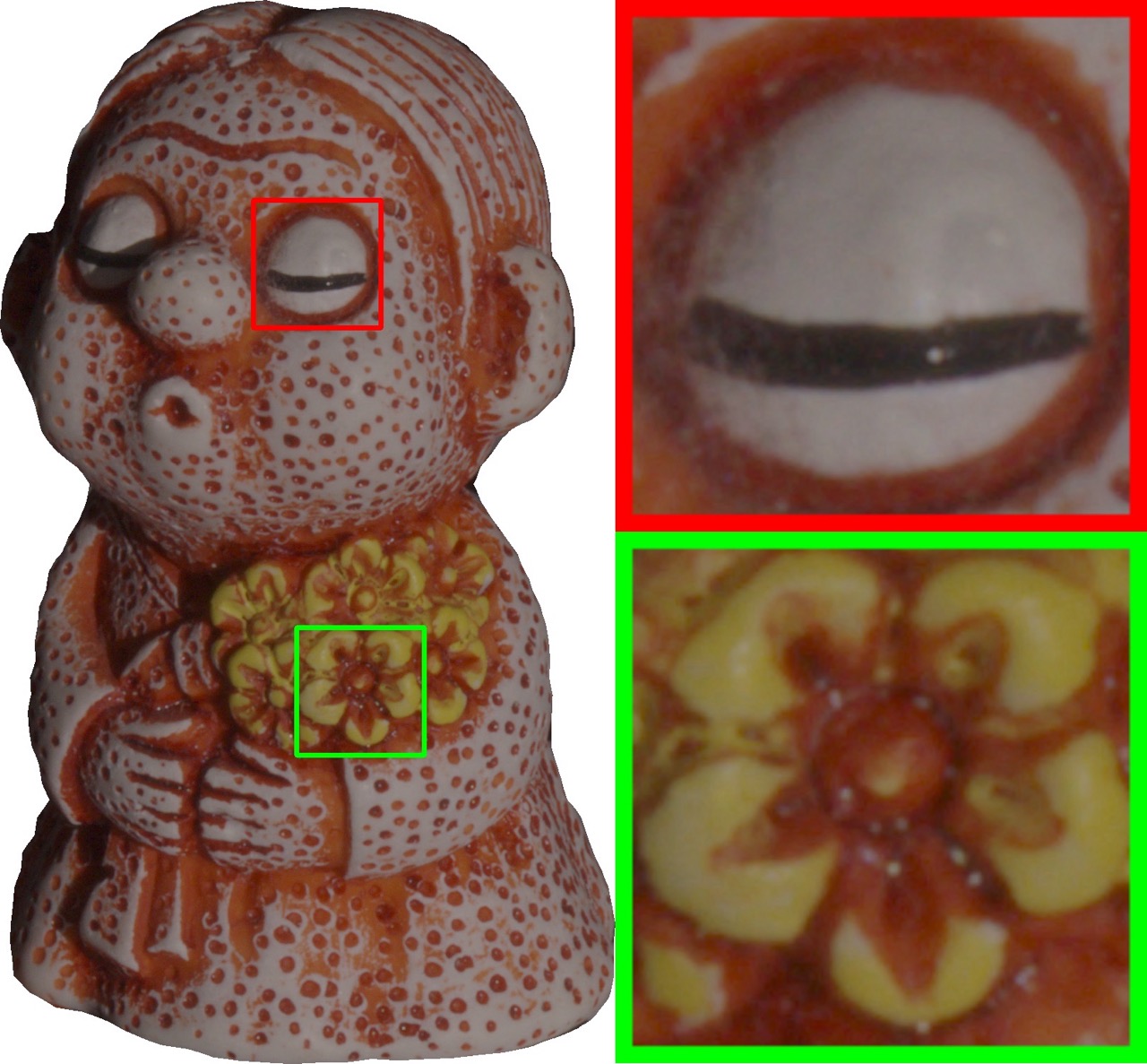} &
\includegraphics[width=\figwidthG\linewidth]{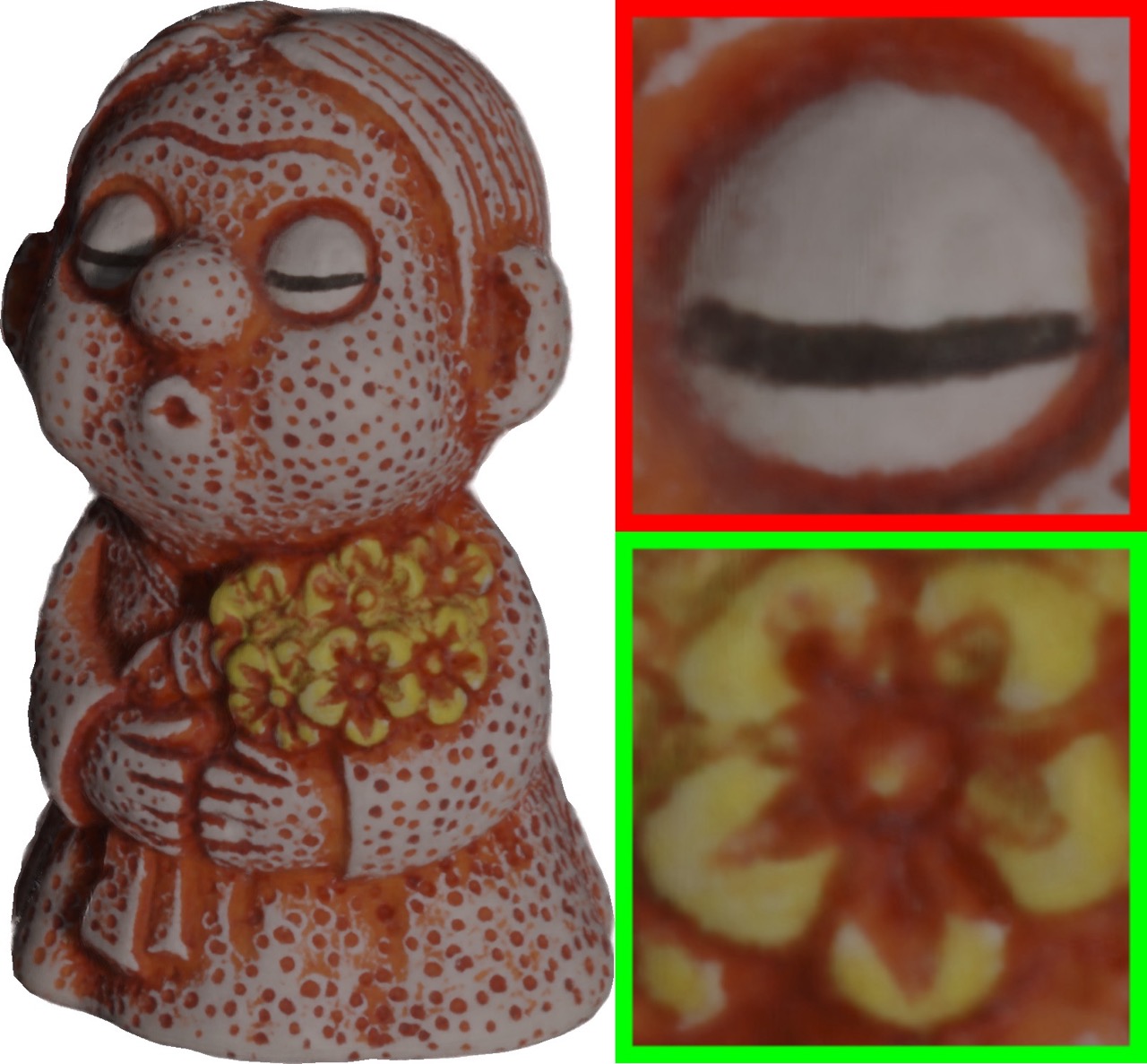} &
\includegraphics[width=\figwidthG\linewidth]{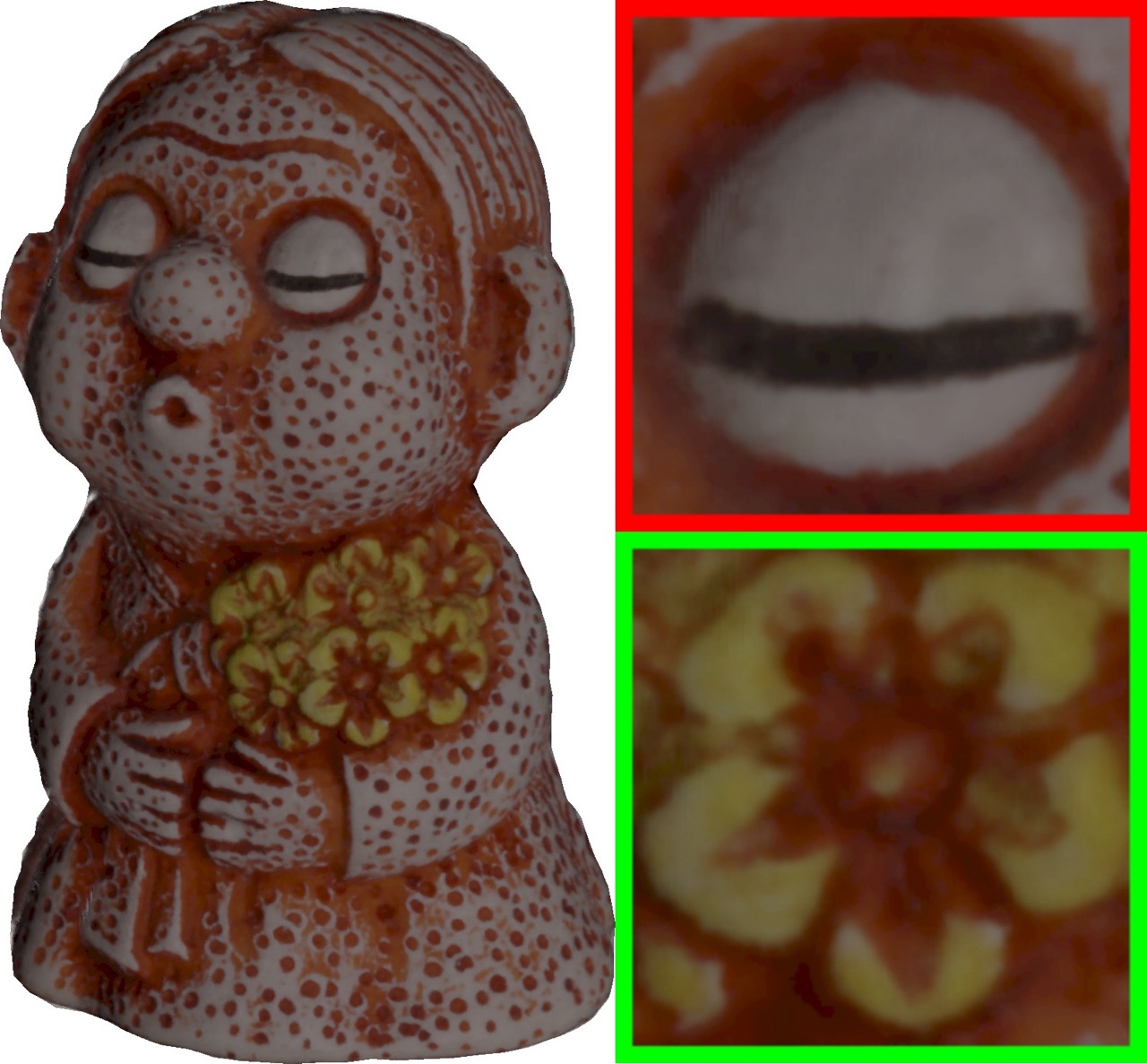} &
\begin{tabular}[b]{@{}c@{}}
      \includegraphics[width=\figwidthBRDF\linewidth]{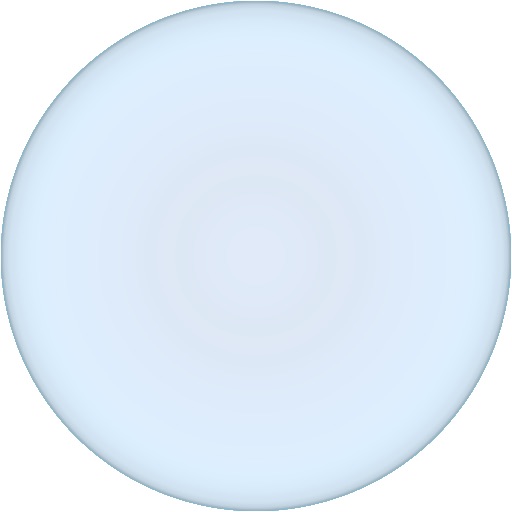} \\
    \includegraphics[width=\figwidthBRDF\linewidth]{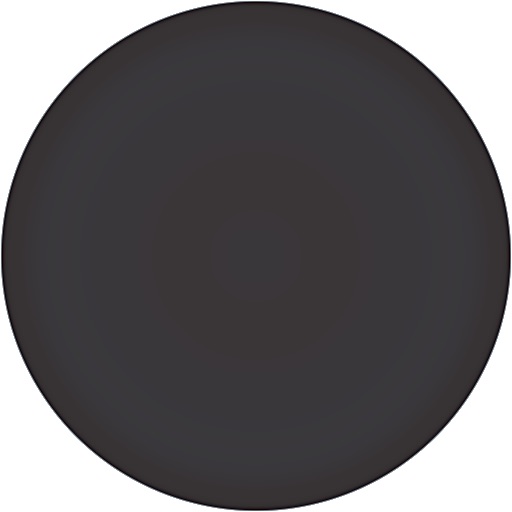}
    \\
  \includegraphics[width=\figwidthBRDF\linewidth]{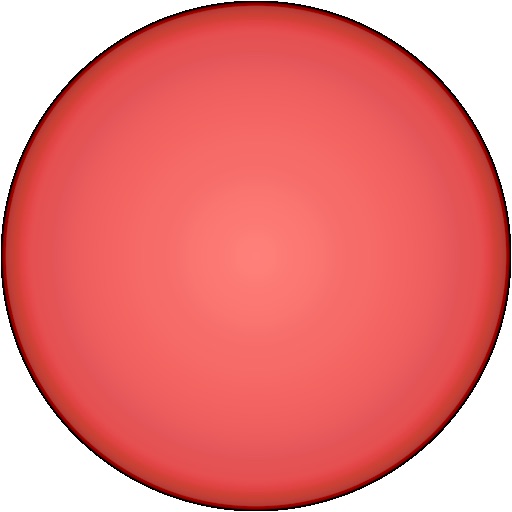}     \\
   \includegraphics[width=\figwidthBRDF\linewidth]{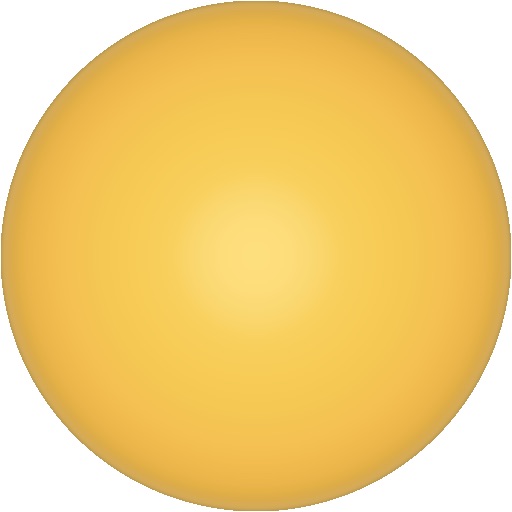}
\end{tabular}&
\includegraphics[width=\figwidthG\linewidth]{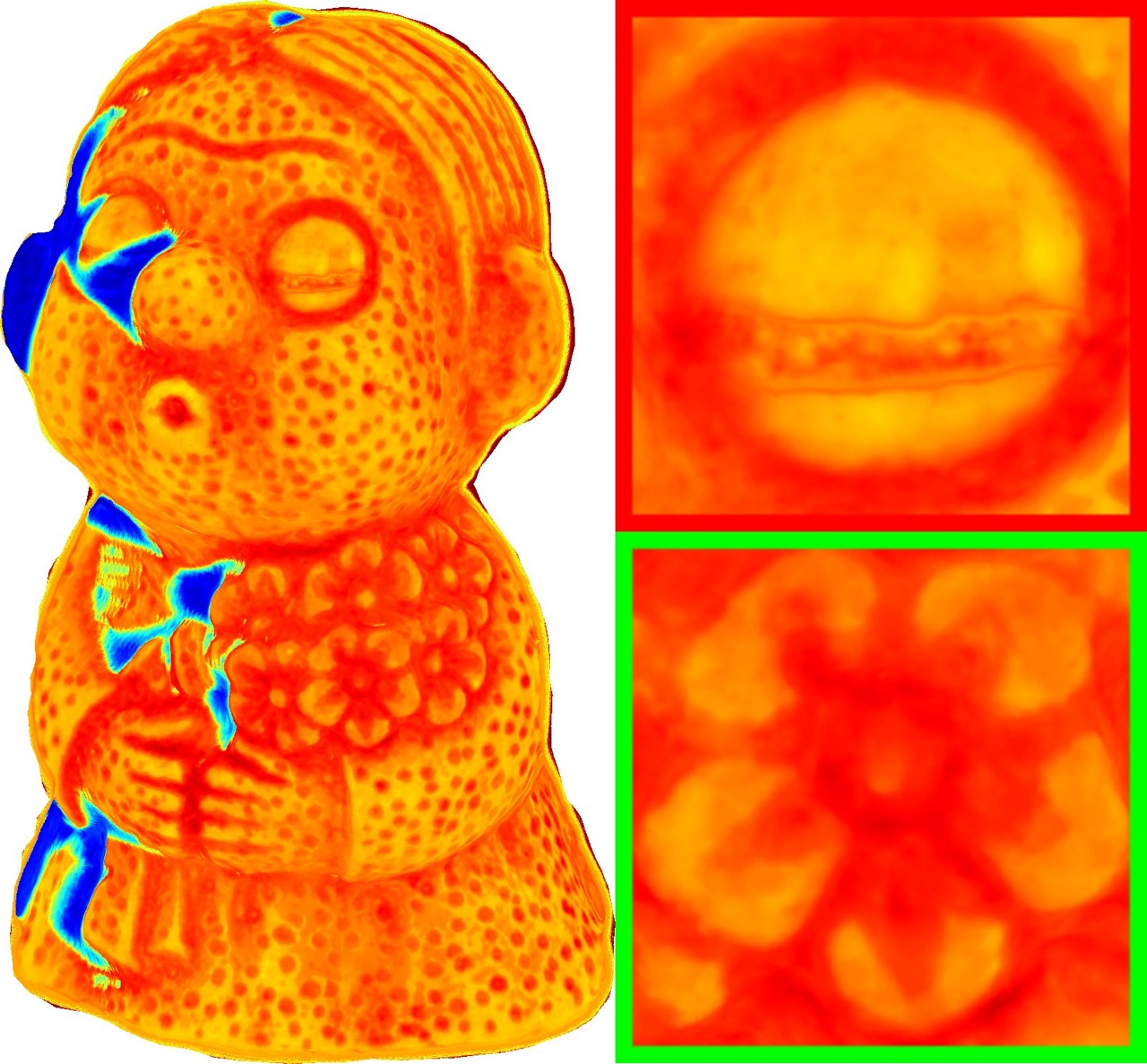} &
\colorbartwo{0.04}{1} &
\includegraphics[width=\figwidthG\linewidth]{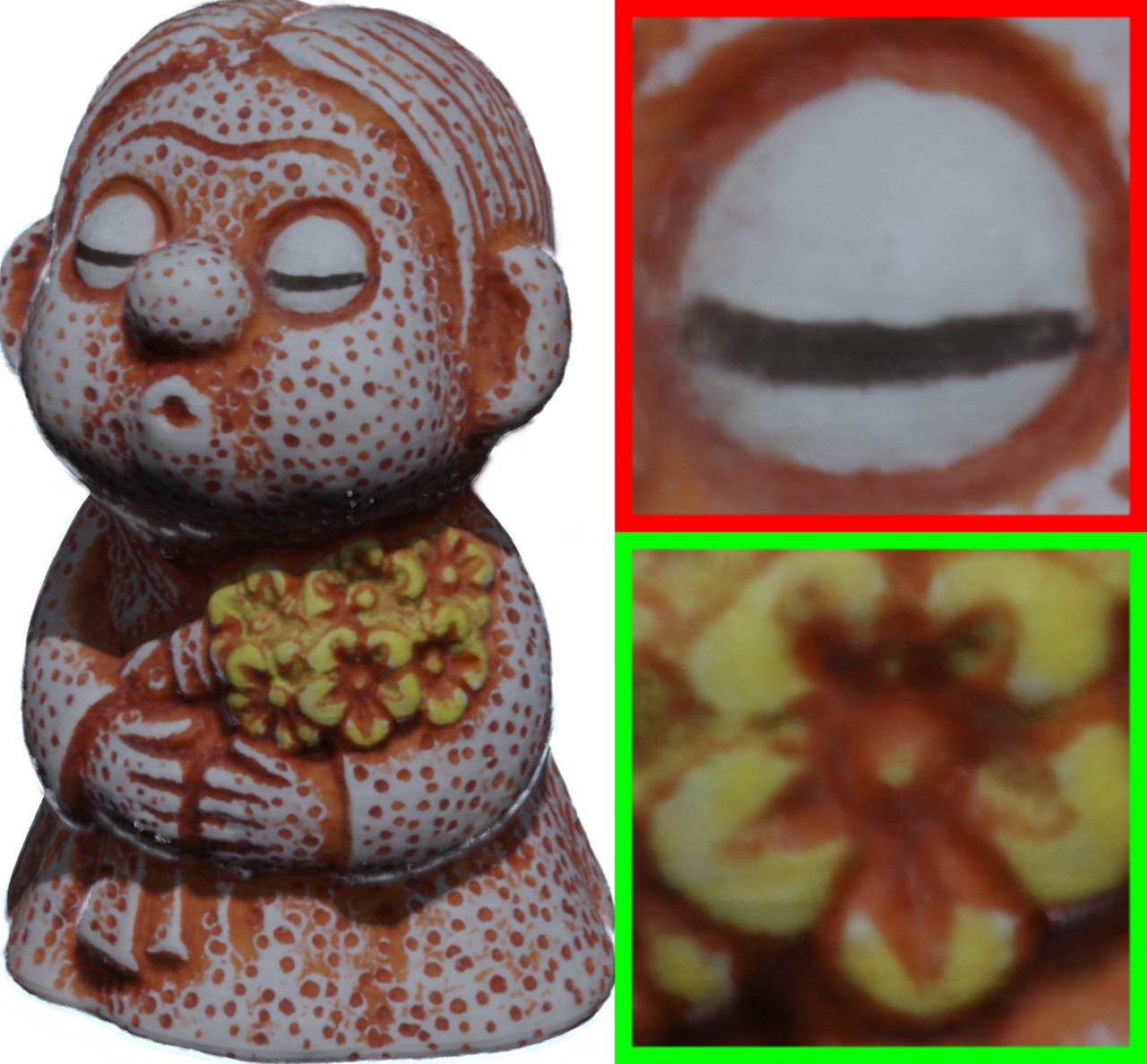} &
\includegraphics[width=0.04\linewidth]{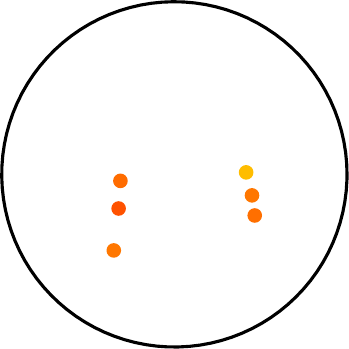} \vspace{0.5em}
\\
\includegraphics[width=\figwidthG\linewidth]{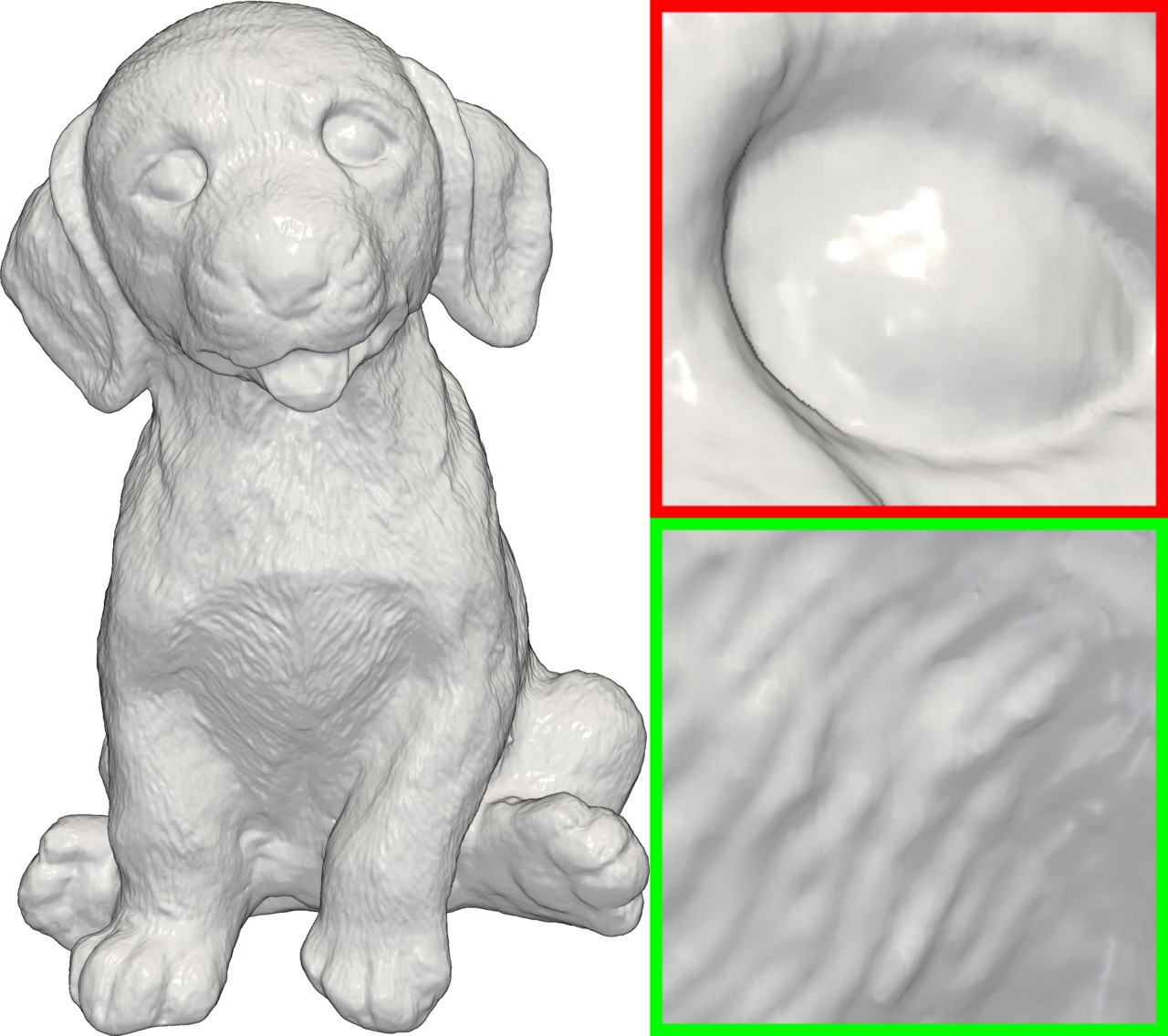} &
\includegraphics[width=\figwidthG\linewidth]{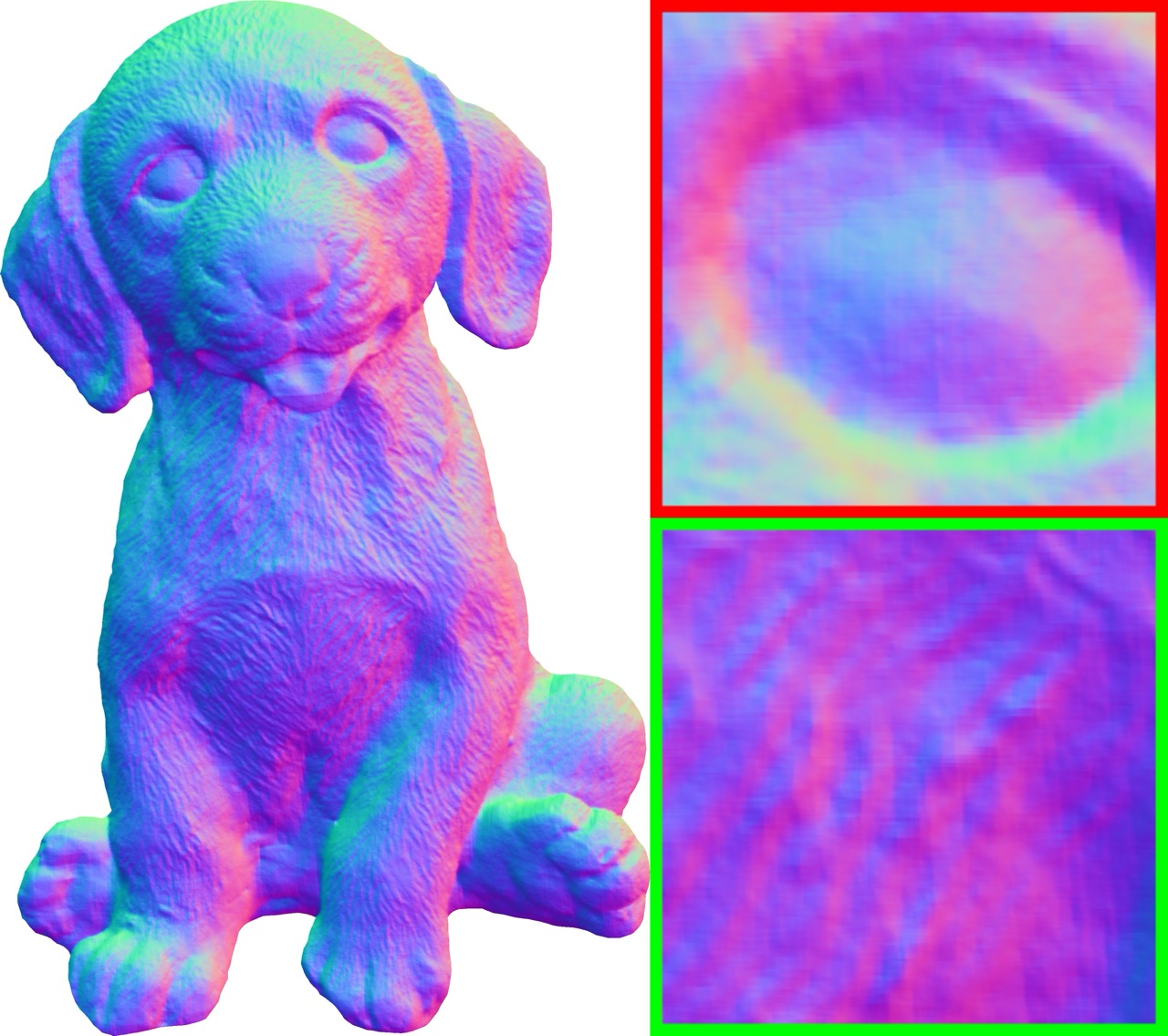} &
\includegraphics[width=\figwidthG\linewidth]{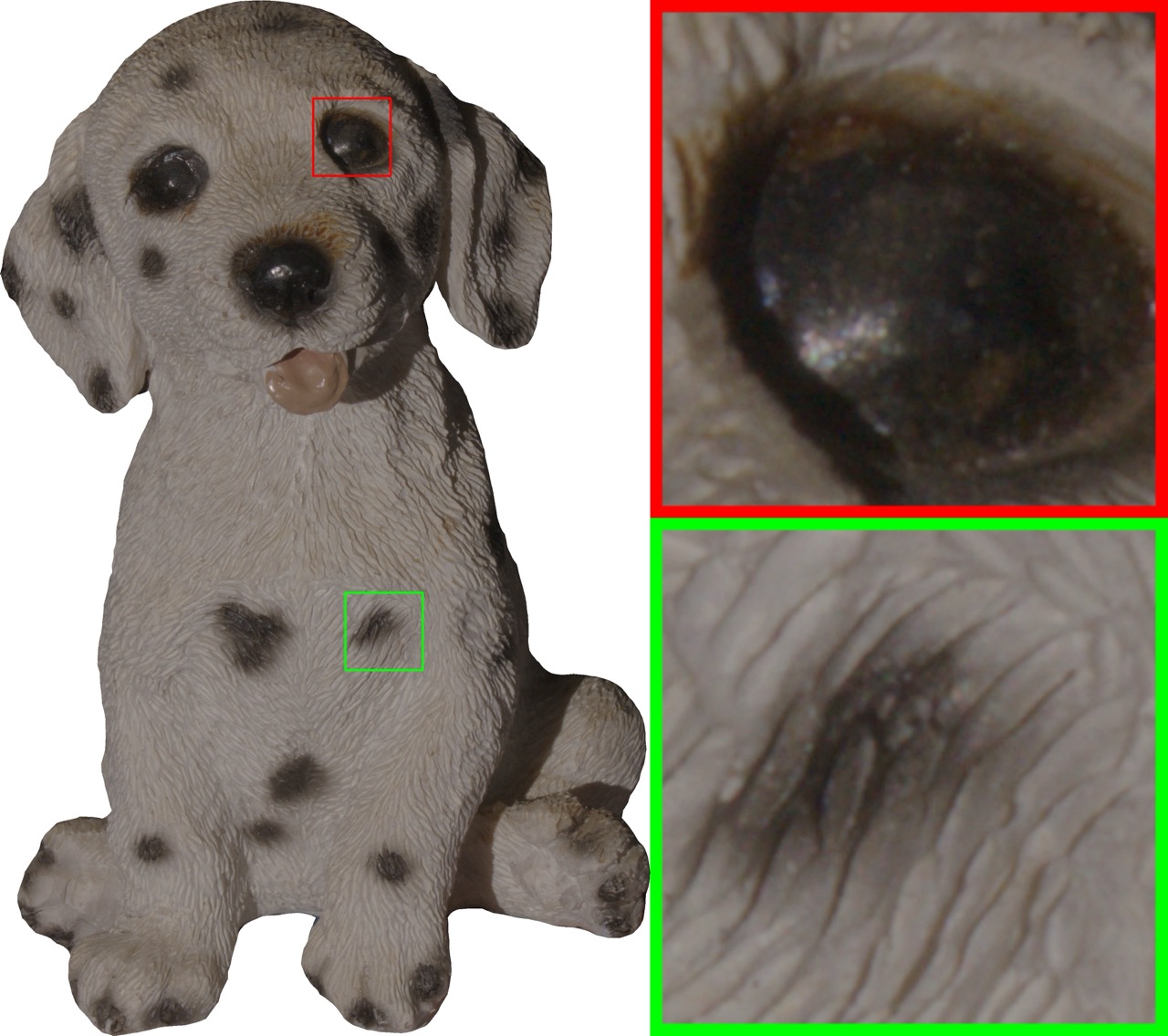} &
\includegraphics[width=\figwidthG\linewidth]{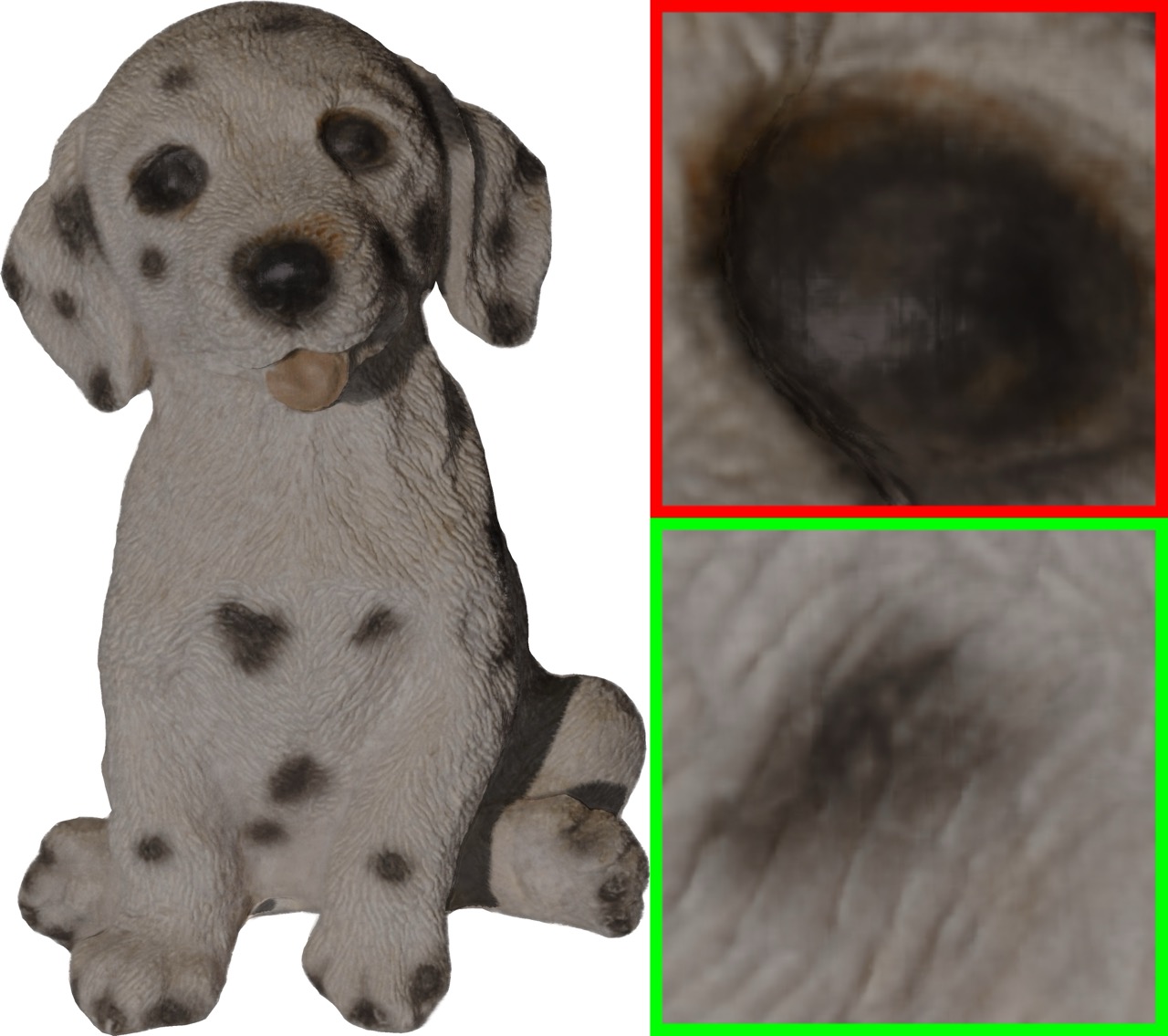} &
\includegraphics[width=\figwidthG\linewidth]{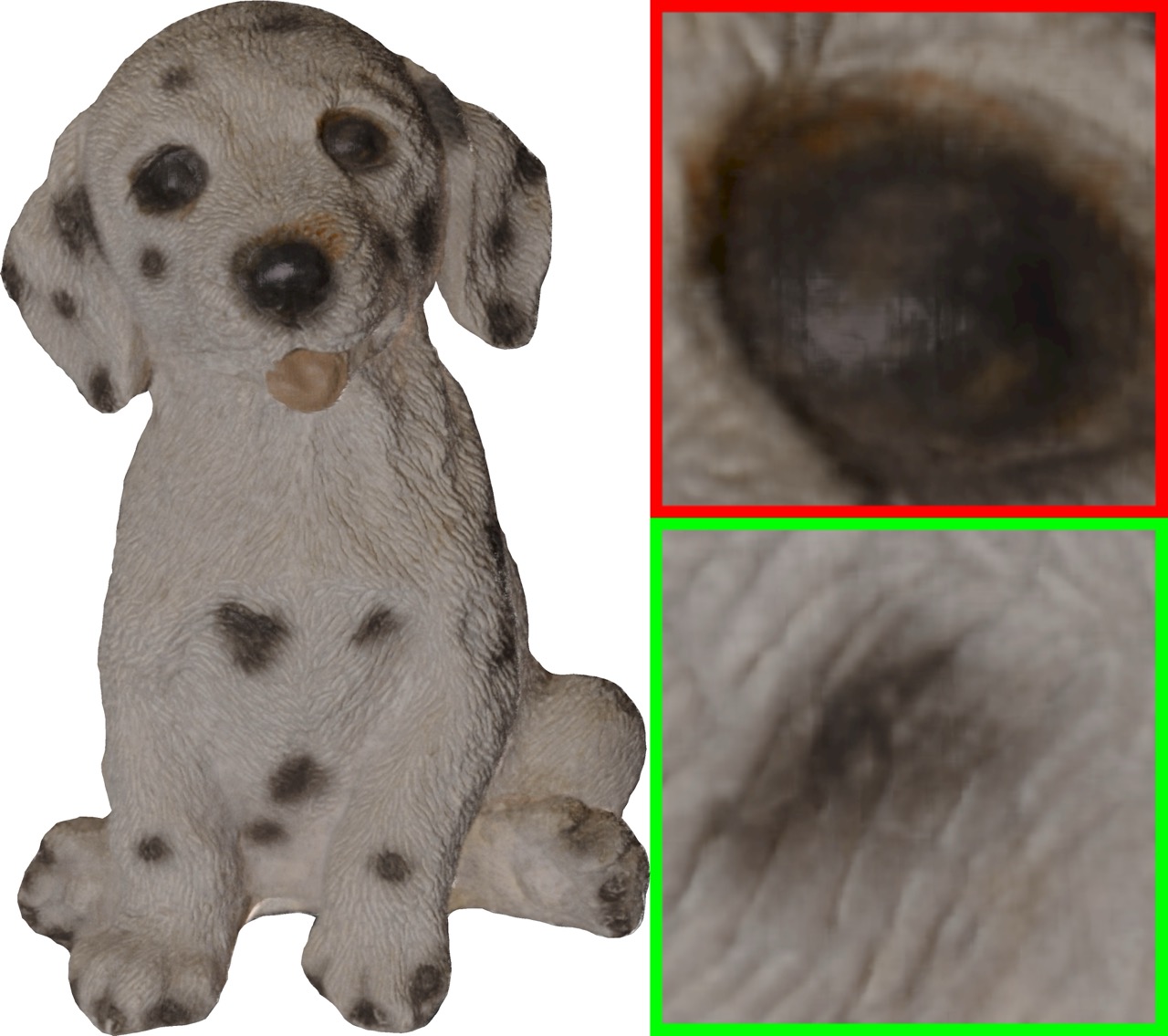} &
\begin{tabular}[b]{@{}c@{}}
  \includegraphics[width=\figwidthBRDF\linewidth]{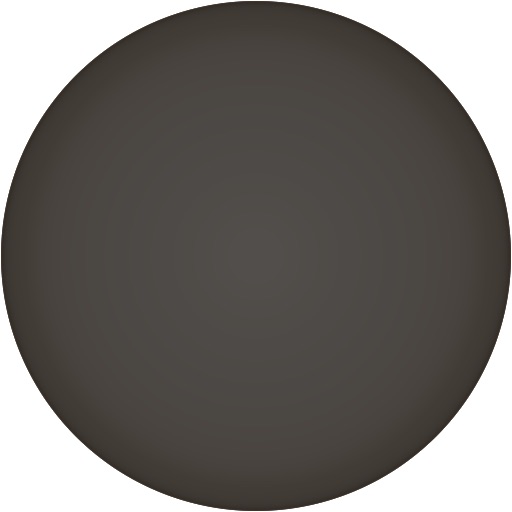}     \\
        \includegraphics[width=\figwidthBRDF\linewidth]{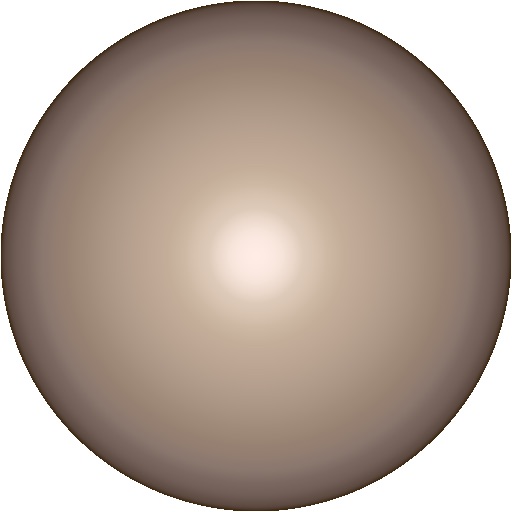} \\
   \includegraphics[width=\figwidthBRDF\linewidth]{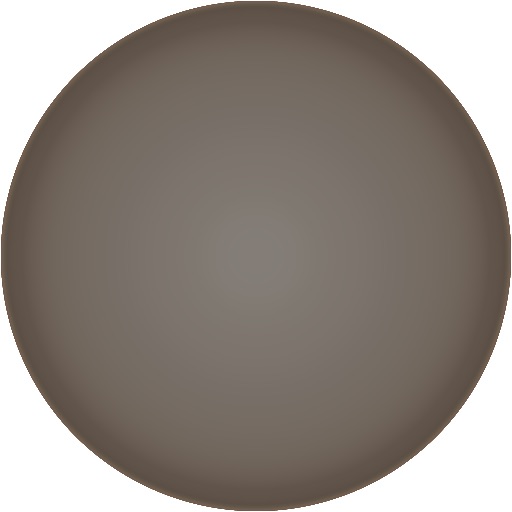}  \\
      \includegraphics[width=\figwidthBRDF\linewidth]{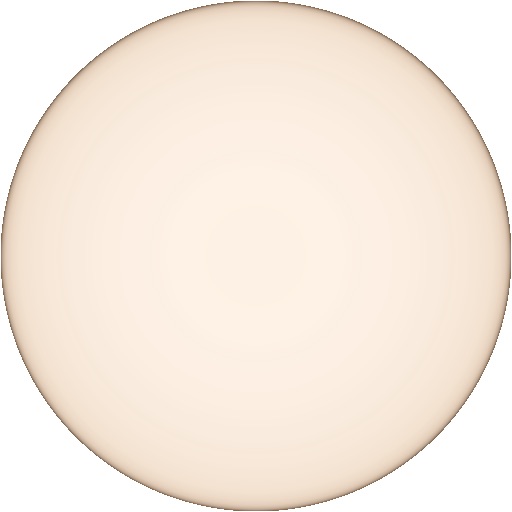} 
\end{tabular}&
\includegraphics[width=\figwidthG\linewidth]{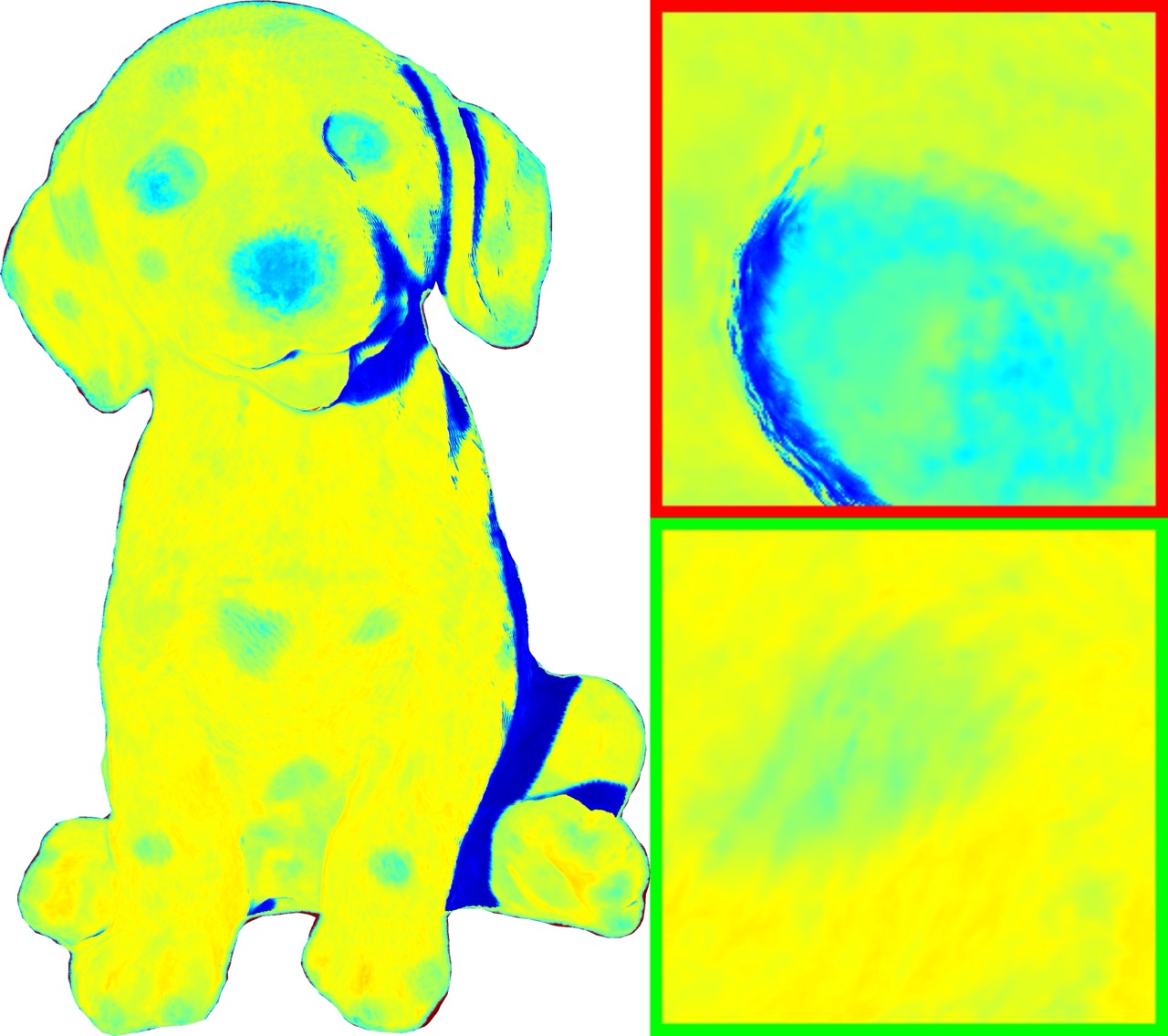} &
\colorbartwo{0.04}{1} &
\includegraphics[width=\figwidthG\linewidth]{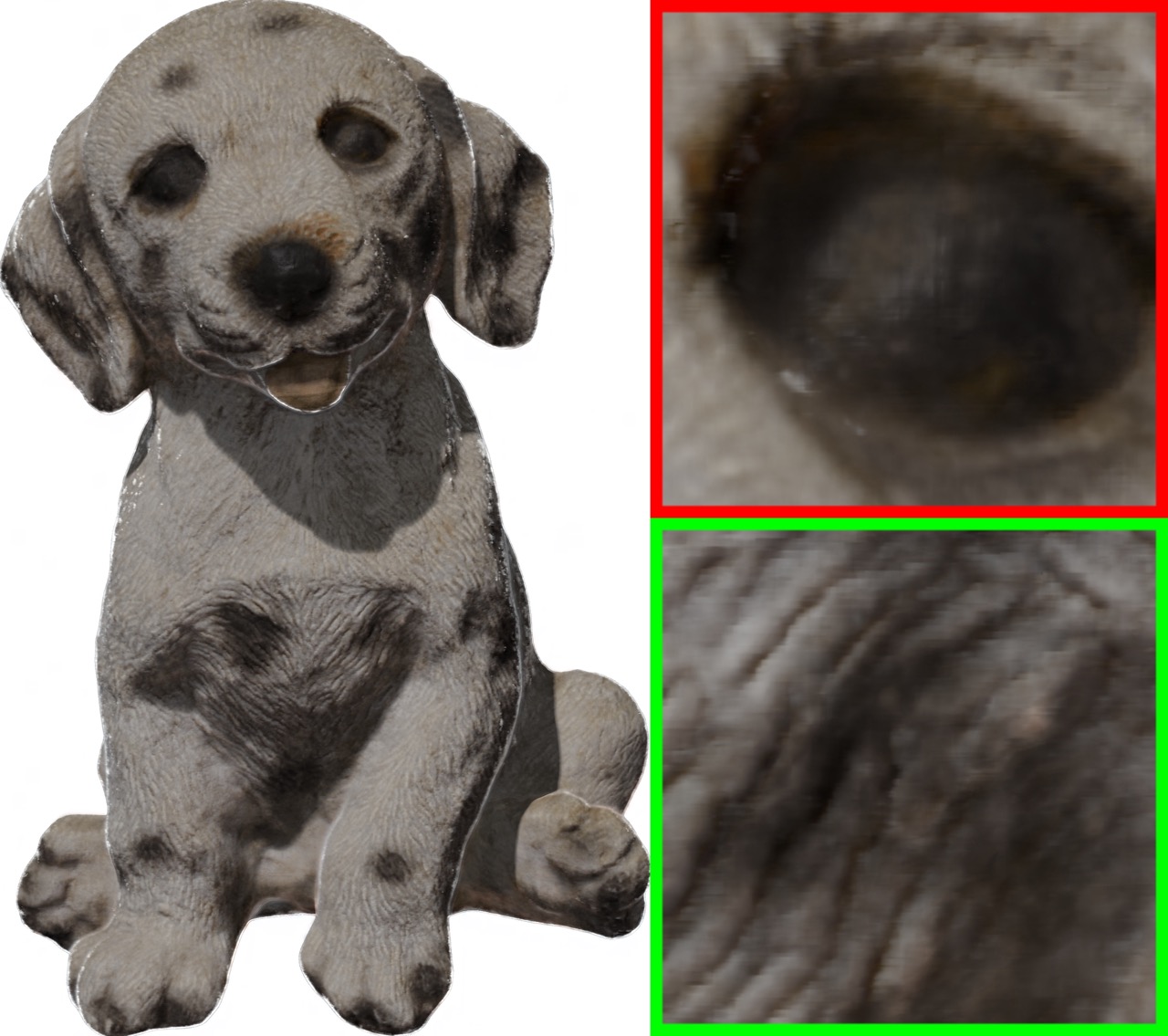} &
\includegraphics[width=0.04\linewidth]{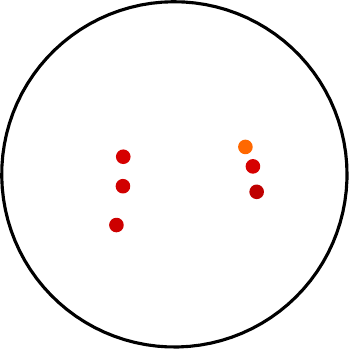}
    \end{tabular}
    \vspace{-1.2em}
    \caption{\textbf{Qualitative results using view-unaligned OLAT images.}  We place the real-world captured images in the third column to facilitate visual comparison with rendered normal maps and color images. Relighting uses a top-down directional light source.
    } \label{fig.real_world_results}
    \vspace{-1em}
\end{figure*}
\begin{figure}
\scriptsize
    \centering
    \begin{tabular}{@{}c@{}c@{}c@{}c@{}c@{}}
Shape & Normal & \textbf{Capture} & Rendering & BRDF \\
\includegraphics[width=0.232\linewidth]{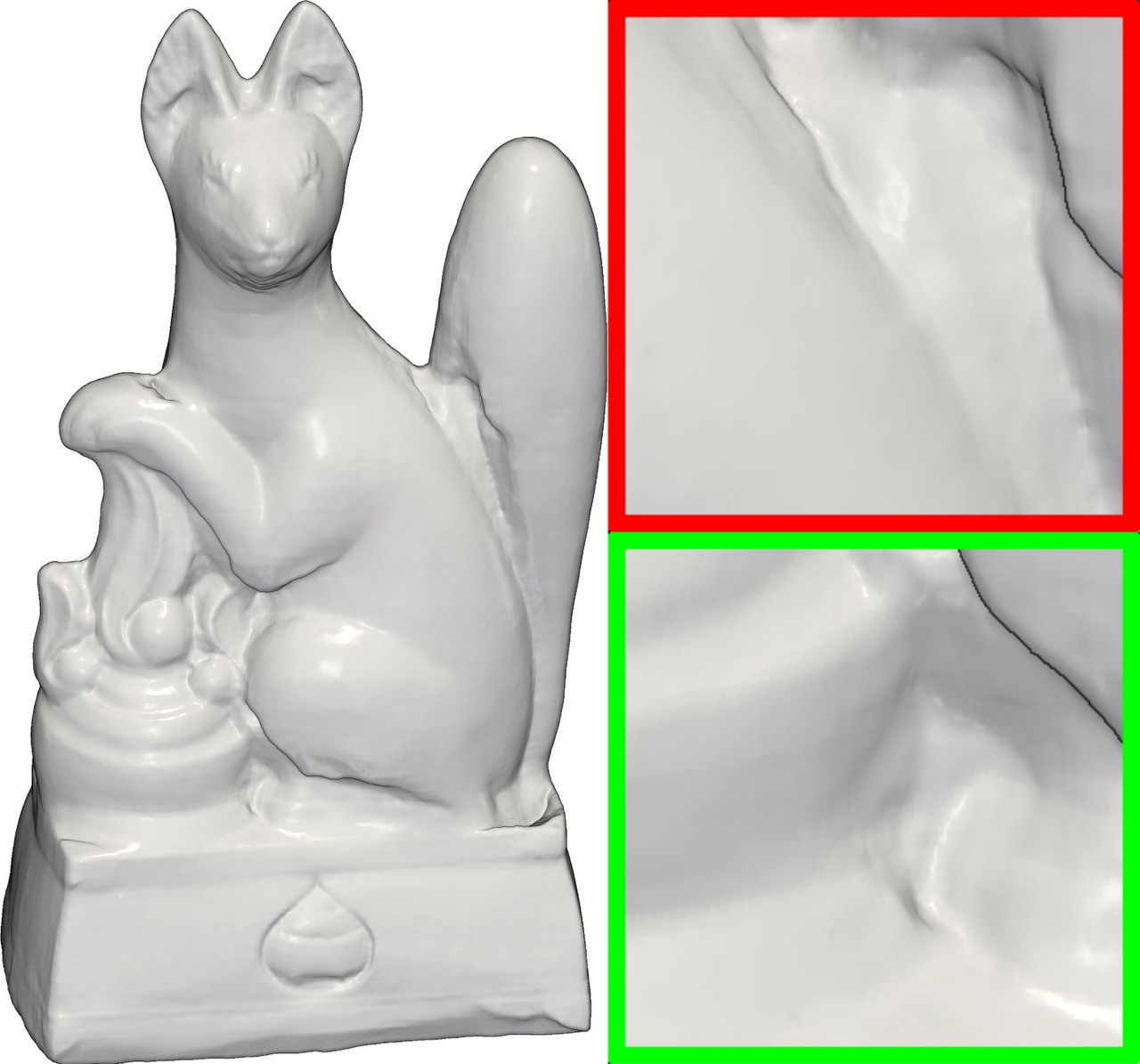}&
\includegraphics[width=0.232\linewidth]{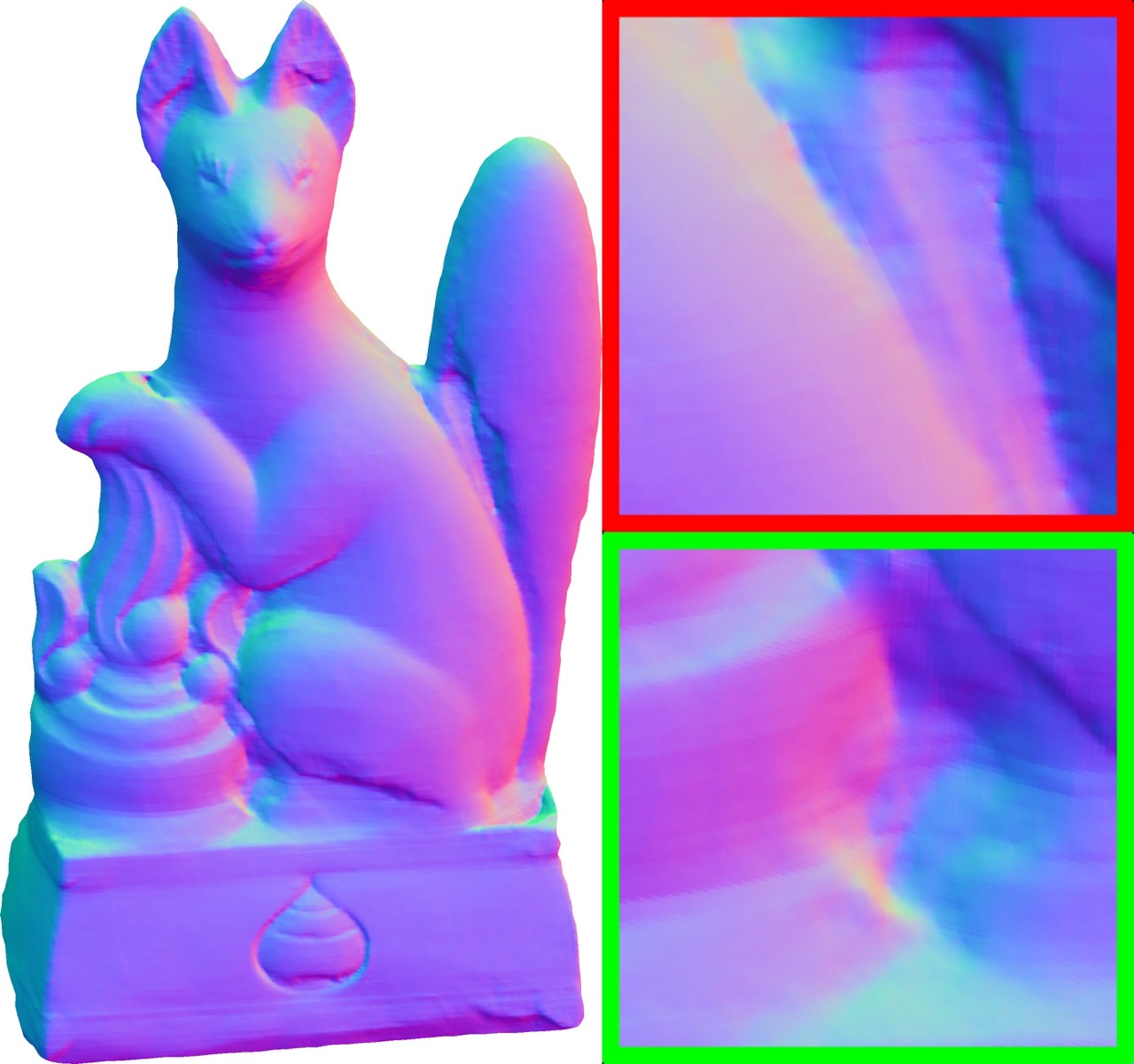} &
\includegraphics[width=0.232\linewidth]{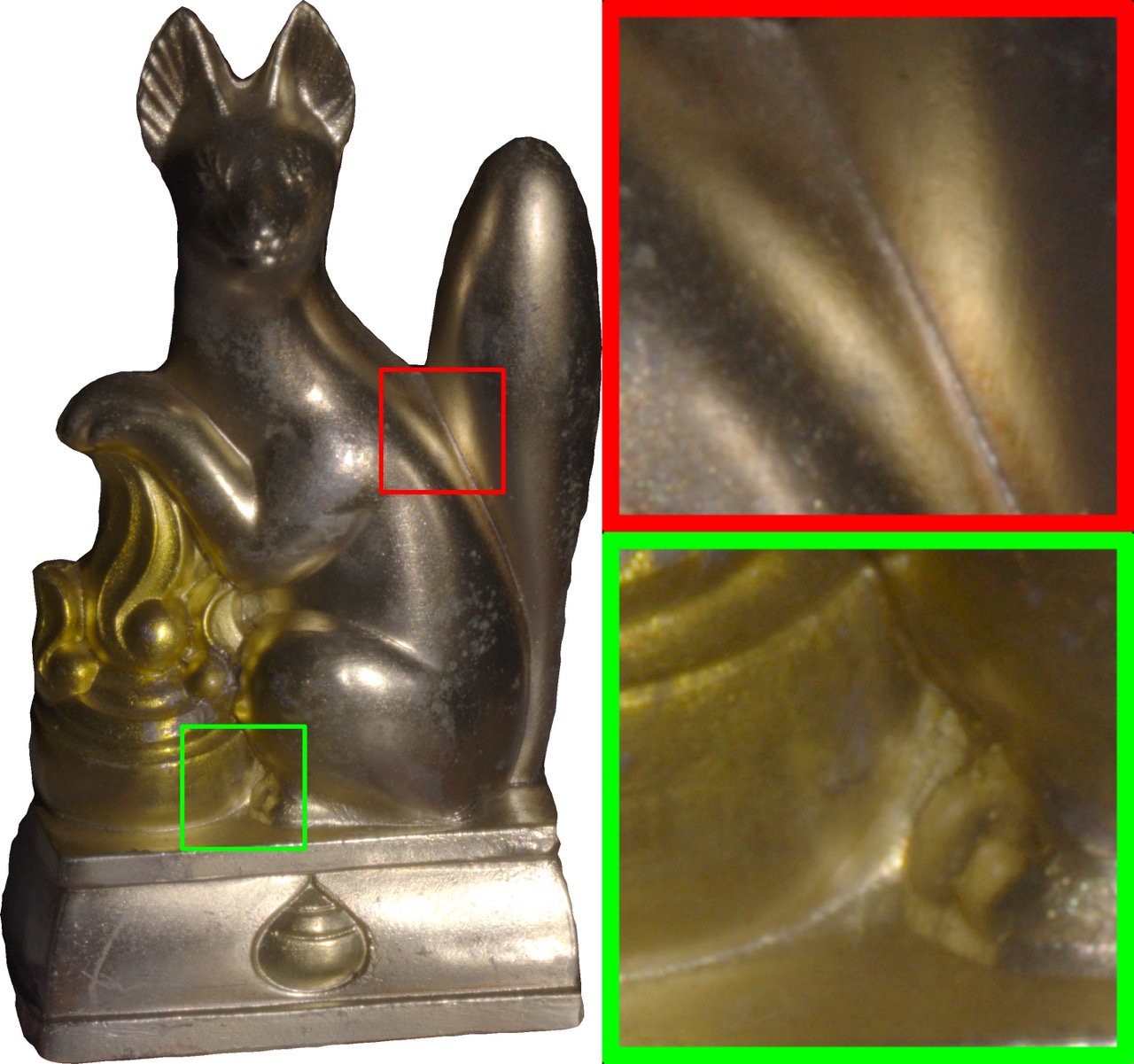} & 
\includegraphics[width=0.232\linewidth]{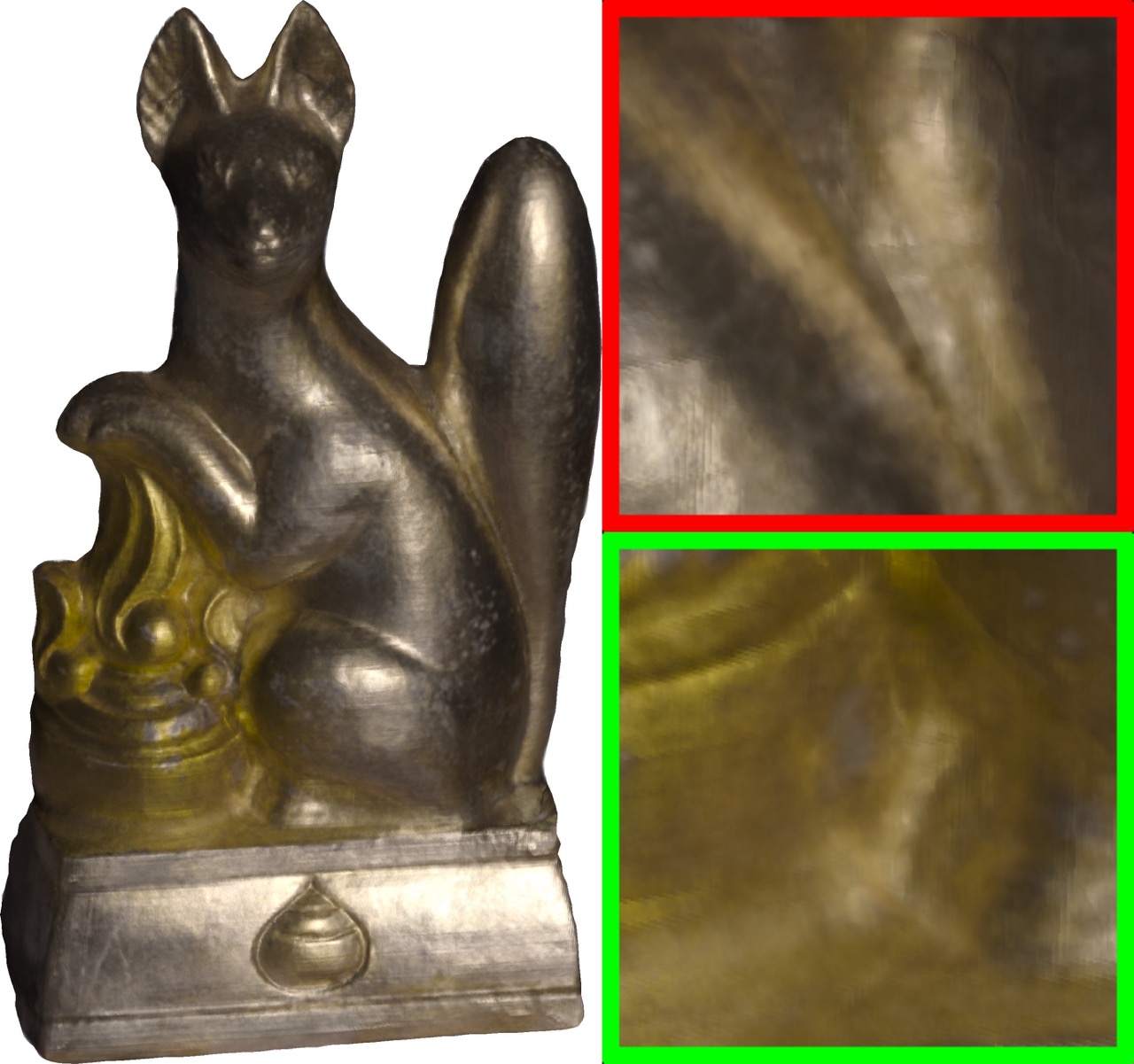} &
\begin{tabular}[b]{@{}c@{}}
   \includegraphics[width=0.06\linewidth]{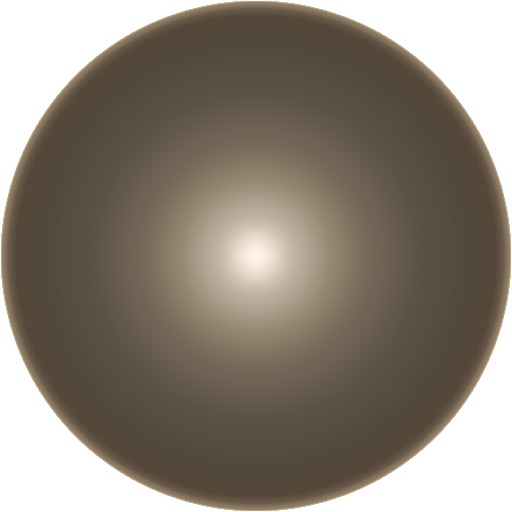}  \\
  \includegraphics[width=0.06\linewidth]{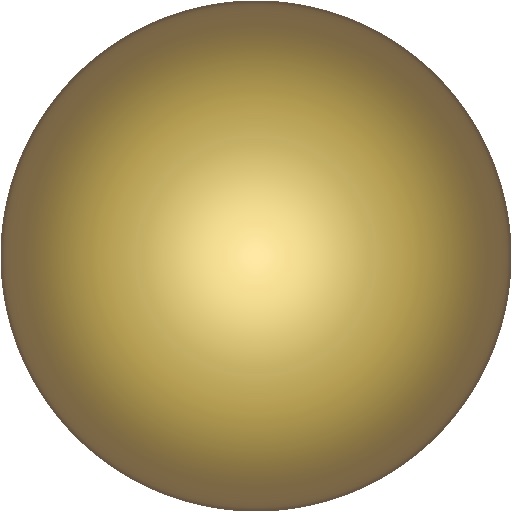}    
\end{tabular}
    \end{tabular}
    \vspace{-1.2em}
    \caption{\textbf{Failure case.} Our method breaks down in concave regions of reflective surfaces, since inter-reflections are not explicitly modeled in our pipeline.}
    \vspace{-1em}
    \label{fig.limitation}
\end{figure}

\subsection{Generalization to View-unaligned OLAT}
\label{sec.application}
This section presents qualitative results on our self-collected view-unaligned OLAT images.
No baseline comparison is provided, as no publicly available MVPS implementation supports this setup.

\vspace{0.5em}
\noindent
\textbf{Capture setup.}
As shown in \cref{fig.capture_setup_real}, we prepare six strobe lights and a camera to capture the target object placed on a motorized turntable.
For each light, we capture a set of multi-view images by rotating the object and then switch to the next light for another set. 
A total of $144$ OLAT images (6L24V) are captured per object (\cref{fig.teaser}).
We calibrate camera parameters using RealityCapture~\cite{realitycapture}, and segment multi-view foreground masks using SAM2~\cite{ravi2024sam2}.
No light calibration is conducted.
See supplementary for details.

\vspace{0.5em}
\noindent
\textbf{Results.}
\Cref{fig.real_world_results,fig.teaser} show qualitative results using our captured view-unaligned OLAT images. 
Our method reconstructs perceptually accurate surfaces with fine-grained details.
Benefitting from the neural BRDF representation, our method is effective for diverse reflectance types, including bronzed metallic (\cref{fig.real_world_results}, first row), hybrid diffuse and specular (\cref{fig.real_world_results}, second and third rows), and ceramic surfaces (\cref{fig.teaser}).
Our rendering pipeline faithfully reproduces the acquired OLAT images and synthesizes unshadowed images that are infeasible to directly capture.
The estimated light directions align well with our camera-light setup and stay consistent across different scenes.
The lighting estimation accuracy is further supported by the rendered shadow maps, where shadowed regions align with those observed in the captured images.

\vspace{0.5em}
\noindent
\textbf{Limitations.}
As shown in \cref{fig.limitation}, our method performs less reliably in concave regions of reflective surfaces, as inter-reflections are not explicitly modeled in our pipeline.
Moreover, our method struggles to reproduce high-frequency color details and sharp specular lobes, which is evident when zooming into the rendered images.

\section{Conclusions}
We have presented a neural inverse rendering approach for recovering geometry, SVBRDFs, and per-light directions and relative intensities from multi-view OLAT images.
By discarding intermediate PS cues and optimizing from raw pixel values, our method outperforms stage-by-stage MVPS methods and generalizes to view-unaligned OLAT images.
In principle, view-unaligned OLAT images can be captured by a handheld camera-light setup, and we hope this work encourages the development of MVPS-based handheld 3D scanning systems.

\vspace{0.35em}
\noindent
\textbf{Future Work.}
Future research may relax the directional lighting assumption to support more general point light sources.
While our angular encoding design is empirically effective, it remains heuristic and requires further theoretical analysis.
Another direction is to model inter-reflections and enhance the learning of high-frequency color details.

\clearpage
{\small
\bibliographystyle{ieee_fullname}
\bibliography{main}

\begin{thebibliography}{56}
\providecommand{\natexlab}[1]{#1}
\providecommand{\url}[1]{\texttt{#1}}
\expandafter\ifx\csname urlstyle\endcsname\relax
  \providecommand{\doi}[1]{doi: #1}\else
  \providecommand{\doi}{doi: \begingroup \urlstyle{rm}\Url}\fi

\bibitem[rea()]{realitycapture}
Reality capture.
\newblock \url{https://www.capturingreality.com}.

\bibitem[Atzmon and Lipman(2020)]{atzmon2020sal}
Matan Atzmon and Yaron Lipman.
\newblock Sal: {S}ign agnostic learning of shapes from raw data.
\newblock In \emph{{Proc. of Computer Vision and Pattern Recognition (CVPR)}},
  pages 2565--2574, 2020.

\bibitem[Blinn(1977)]{blinn1977models}
James~F Blinn.
\newblock Models of light reflection for computer synthesized pictures.
\newblock In \emph{Proceedings of the 4th annual conference on Computer
  graphics and interactive techniques}, pages 192--198, 1977.

\bibitem[Boss et~al.(2021)Boss, Braun, Jampani, Barron, Liu, and
  Lensch]{boss2021nerd}
Mark Boss, Raphael Braun, Varun Jampani, Jonathan~T Barron, Ce Liu, and Hendrik
  Lensch.
\newblock {NeRD: N}eural reflectance decomposition from image collections.
\newblock In \emph{{Proc. of International Conference on Computer Vision
  (ICCV)}}, pages 12684--12694, 2021.

\bibitem[Brahimi et~al.(2024)Brahimi, Haefner, Ye, Goldluecke, and
  Cremers]{Brahimi_2024_CVPR}
Mohammed Brahimi, Bjoern Haefner, Zhenzhang Ye, Bastian Goldluecke, and Daniel
  Cremers.
\newblock Sparse views near light: A practical paradigm for uncalibrated
  point-light photometric stereo.
\newblock In \emph{{Proc. of Computer Vision and Pattern Recognition (CVPR)}},
  pages 11862--11872, 2024.

\bibitem[Brument et~al.(2024)Brument, Bruneau, Qu{\'e}au, M{\'e}lou, Lauze,
  Durou, and Calvet]{brument2024rnb}
Baptiste Brument, Robin Bruneau, Yvain Qu{\'e}au, Jean M{\'e}lou,
  Fran{\c{c}}ois~Bernard Lauze, Jean-Denis Durou, and Lilian Calvet.
\newblock Rnb-neus: Reflectance and normal-based multi-view 3d reconstruction.
\newblock In \emph{{Proc. of Computer Vision and Pattern Recognition (CVPR)}},
  pages 5230--5239, 2024.

\bibitem[Bruneau et~al.(2025)Bruneau, Brument, Qu{\'e}au, M{\'e}lou, Lauze,
  Durou, and Calvet]{bruneau2025multi}
Robin Bruneau, Baptiste Brument, Yvain Qu{\'e}au, Jean M{\'e}lou,
  Fran{\c{c}}ois~Bernard Lauze, Jean-Denis Durou, and Lilian Calvet.
\newblock Multi-view surface reconstruction using normal and reflectance cues.
\newblock \emph{arXiv preprint arXiv:2506.04115}, 2025.

\bibitem[Cao and Taketomi(2024)]{Cao_2024_CVPR}
Xu Cao and Takafumi Taketomi.
\newblock Supernormal: {N}eural surface reconstruction via multi-view normal
  integration.
\newblock In \emph{{Proc. of Computer Vision and Pattern Recognition (CVPR)}},
  pages 20581--20590, 2024.

\bibitem[Cao et~al.(2023)Cao, Santo, Okura, and Matsushita]{mvas2023cao}
Xu Cao, Hiroaki Santo, Fumio Okura, and Yasuyuki Matsushita.
\newblock Multi-view azimuth stereo via tangent space consistency.
\newblock In \emph{{Proc. of Computer Vision and Pattern Recognition (CVPR)}},
  2023.

\bibitem[Chang et~al.(2007)Chang, Lee, and Lee]{chang2007multiview}
Ju~Yong Chang, Kyoung~Mu Lee, and Sang~Uk Lee.
\newblock Multiview normal field integration using level set methods.
\newblock In \emph{{Proc. of Computer Vision and Pattern Recognition (CVPR)}},
  pages 1--8. IEEE, 2007.

\bibitem[Chen et~al.(2019)Chen, Han, Shi, Matsushita, and
  Wong]{chen2019SDPS_Net}
Guanying Chen, Kai Han, Boxin Shi, Yasuyuki Matsushita, and Kwan-Yee~K. Wong.
\newblock {SDPS-Net: S}elf-calibrating deep photometric stereo networks.
\newblock In \emph{{Proc. of Computer Vision and Pattern Recognition (CVPR)}},
  2019.

\bibitem[Cheng et~al.(2023)Cheng, Li, and Li]{cheng2023wildlight}
Ziang Cheng, Junxuan Li, and Hongdong Li.
\newblock Wildlight: {I}n-the-wild inverse rendering with a flashlight.
\newblock In \emph{{Proc. of Computer Vision and Pattern Recognition (CVPR)}},
  pages 4305--4314, 2023.

\bibitem[Chung et~al.(2024{\natexlab{a}})Chung, Choi, and
  Baek]{Chung_2024_CVPR}
Hoon-Gyu Chung, Seokjun Choi, and Seung-Hwan Baek.
\newblock Differentiable point-based inverse rendering.
\newblock In \emph{Proceedings of the IEEE/CVF Conference on Computer Vision
  and Pattern Recognition (CVPR)}, pages 4399--4409, 2024{\natexlab{a}}.

\bibitem[Chung et~al.(2024{\natexlab{b}})Chung, Choi, and
  Baek]{chung2024differentiable}
Hoon-Gyu Chung, Seokjun Choi, and Seung-Hwan Baek.
\newblock Differentiable inverse rendering with interpretable basis brdfs.
\newblock In \emph{{Proc. of Computer Vision and Pattern Recognition (CVPR)}},
  2024{\natexlab{b}}.

\bibitem[Cook and Torrance(1982)]{cook1982}
R.~L. Cook and K.~E. Torrance.
\newblock A reflectance model for computer graphics.
\newblock \emph{ACM Trans. Graph.}, 1\penalty0 (1):\penalty0 7–24, 1982.

\bibitem[Dave et~al.(2022)Dave, Zhao, and Veeraraghavan]{dave2022pandora}
Akshat Dave, Yongyi Zhao, and Ashok Veeraraghavan.
\newblock {Pandora: P}olarization-aided neural decomposition of radiance.
\newblock In \emph{European conference on computer vision}, pages 538--556.
  Springer, 2022.

\bibitem[Esteban et~al.(2008)Esteban, Vogiatzis, and
  Cipolla]{esteban2008multiview}
Carlos~Hernandez Esteban, George Vogiatzis, and Roberto Cipolla.
\newblock Multiview photometric stereo.
\newblock \emph{{IEEE Transactions on Pattern Analysis and Machine Intelligence
  (PAMI)}}, 30\penalty0 (3):\penalty0 548--554, 2008.

\bibitem[Gropp et~al.(2020)Gropp, Yariv, Haim, Atzmon, and Lipman]{igr2020icml}
Amos Gropp, Lior Yariv, Niv Haim, Matan Atzmon, and Yaron Lipman.
\newblock Implicit geometric regularization for learning shapes.
\newblock In \emph{Proceedings of Machine Learning and Systems 2020}, pages
  3569--3579. 2020.

\bibitem[Hofherr et~al.(2025)Hofherr, Haefner, and Cremers]{hofherr2025neural}
Florian Hofherr, Bjoern Haefner, and Daniel Cremers.
\newblock On {Neural BRDFs: A} thorough comparison of state-of-the-art
  approaches.
\newblock In \emph{2025 IEEE/CVF Winter Conference on Applications of Computer
  Vision (WACV)}, pages 1785--1794. IEEE, 2025.

\bibitem[Ikehata(2023)]{ikehata2023sdmunips}
Satoshi Ikehata.
\newblock Scalable, detailed and mask-free universal photometric stereo.
\newblock In \emph{{Proc. of Computer Vision and Pattern Recognition (CVPR)}},
  2023.

\bibitem[Kaya et~al.(2022{\natexlab{a}})Kaya, Kumar, Oliveira, Ferrari, and
  Van~Gool]{kaya2022uncertainty}
Berk Kaya, Suryansh Kumar, Carlos Oliveira, Vittorio Ferrari, and Luc Van~Gool.
\newblock Uncertainty-aware deep multi-view photometric stereo.
\newblock 2022{\natexlab{a}}.

\bibitem[Kaya et~al.(2022{\natexlab{b}})Kaya, Kumar, Sarno, Ferrari, and
  Van~Gool]{kaya2022neural}
Berk Kaya, Suryansh Kumar, Francesco Sarno, Vittorio Ferrari, and Luc Van~Gool.
\newblock Neural radiance fields approach to deep multi-view photometric
  stereo.
\newblock In \emph{{Proc. of IEEE/CVF Winter Conference on Applications of
  Computer Vision (WACV)}}, pages 1965--1977, 2022{\natexlab{b}}.

\bibitem[Kaya et~al.(2023)Kaya, Kumar, Oliveira, Ferrari, and
  Van~Gool]{kaya2023multi}
Berk Kaya, Suryansh Kumar, Carlos Oliveira, Vittorio Ferrari, and Luc Van~Gool.
\newblock Multi-view photometric stereo revisited.
\newblock In \emph{{Proc. of IEEE/CVF Winter Conference on Applications of
  Computer Vision (WACV)}}, pages 3126--3135, 2023.

\bibitem[Li et~al.(2024)Li, Ono, Uemori, Mihara, Gatto, Nagahara, and
  Moriuchi]{li2024neisf}
Chenhao Li, Taishi Ono, Takeshi Uemori, Hajime Mihara, Alexander Gatto, Hajime
  Nagahara, and Yusuke Moriuchi.
\newblock {NeISF: N}eural incident stokes field for geometry and material
  estimation.
\newblock In \emph{{Proc. of Computer Vision and Pattern Recognition (CVPR)}},
  pages 21434--21445, 2024.

\bibitem[Li et~al.(2025)Li, Ono, Uemori, Nitta, Mihara, Gatto, Nagahara, and
  Moriuchi]{li2025neisf++}
Chenhao Li, Taishi Ono, Takeshi Uemori, Sho Nitta, Hajime Mihara, Alexander
  Gatto, Hajime Nagahara, and Yusuke Moriuchi.
\newblock {NeISF++: N}eural incident stokes field for polarized inverse
  rendering of conductors and dielectrics.
\newblock In \emph{{Proc. of Computer Vision and Pattern Recognition (CVPR)}},
  pages 26493--26503, 2025.

\bibitem[Li and Li(2022)]{li2022selfps}
Junxuan Li and Hongdong Li.
\newblock Self-calibrating photometric stereo by neural inverse rendering.
\newblock In \emph{{Proc. of European Conference on Computer Vision (ECCV)}},
  pages 166--183. Springer, 2022.

\bibitem[Li et~al.(2020)Li, Zhou, Wu, Shi, Diao, and Tan]{li2020multi}
Min Li, Zhenglong Zhou, Zhe Wu, Boxin Shi, Changyu Diao, and Ping Tan.
\newblock Multi-view photometric stereo: {A} robust solution and benchmark
  dataset for spatially varying isotropic materials.
\newblock \emph{IEEE Transactions on Image Processing}, 29:\penalty0
  4159--4173, 2020.

\bibitem[Li et~al.(2023{\natexlab{a}})Li, Gao, Tancik, and
  Kanazawa]{nerfacc_2023_ICCV}
Ruilong Li, Hang Gao, Matthew Tancik, and Angjoo Kanazawa.
\newblock {NerfAcc: E}fficient sampling accelerates nerfs.
\newblock In \emph{{Proc. of International Conference on Computer Vision
  (ICCV)}}, pages 18537--18546, 2023{\natexlab{a}}.

\bibitem[Li et~al.(2023{\natexlab{b}})Li, M\"uller, Evans, Taylor, Unberath,
  Liu, and Lin]{li2023neuralangelo}
Zhaoshuo Li, Thomas M\"uller, Alex Evans, Russell~H Taylor, Mathias Unberath,
  Ming-Yu Liu, and Chen-Hsuan Lin.
\newblock Neuralangelo: {H}igh-fidelity neural surface reconstruction.
\newblock In \emph{IEEE Conference on Computer Vision and Pattern Recognition
  ({CVPR})}, 2023{\natexlab{b}}.

\bibitem[Logothetis et~al.(2019)Logothetis, Mecca, and
  Cipolla]{Logothetis_2019_ICCV}
Fotios Logothetis, Roberto Mecca, and Roberto Cipolla.
\newblock A differential volumetric approach to multi-view photometric stereo.
\newblock In \emph{{Proc. of International Conference on Computer Vision
  (ICCV)}}, 2019.

\bibitem[Logothetis et~al.(2024)Logothetis, Budvytis, and
  Cipolla]{logothetis2024nplmv}
Fotios Logothetis, Ignas Budvytis, and Roberto Cipolla.
\newblock Nplmv-ps: {N}eural point-light multi-view photometric stereo.
\newblock \emph{arXiv preprint arXiv:2405.12057}, 2024.

\bibitem[Lorensen and Cline(1987)]{lorensen1987marching}
William~E Lorensen and Harvey~E Cline.
\newblock Marching cubes: {A} high resolution 3{D} surface construction
  algorithm.
\newblock \emph{ACM siggraph computer graphics}, 21\penalty0 (4):\penalty0
  163--169, 1987.

\bibitem[Loshchilov and Hutter(2019)]{loshchilov2018decoupled}
Ilya Loshchilov and Frank Hutter.
\newblock Decoupled weight decay regularization.
\newblock In \emph{International Conference on Learning Representations}, 2019.

\bibitem[Matusik(2003)]{Matusik2003ADR}
Wojciech Matusik.
\newblock A data-driven reflectance model.
\newblock \emph{ACM SIGGRAPH 2003 Papers}, 2003.

\bibitem[Mildenhall et~al.(2022)Mildenhall, Hedman, Martin-Brualla, Srinivasan,
  and Barron]{mildenhall2022rawnerf}
Ben Mildenhall, Peter Hedman, Ricardo Martin-Brualla, Pratul~P. Srinivasan, and
  Jonathan~T. Barron.
\newblock {NeRF} in the dark: {H}igh dynamic range view synthesis from noisy
  raw images.
\newblock \emph{CVPR}, 2022.

\bibitem[M\"uller(2021)]{tiny-cuda-nn}
Thomas M\"uller.
\newblock {tiny-cuda-nn}, 2021.

\bibitem[M\"uller et~al.(2022)M\"uller, Evans, Schied, and
  Keller]{mueller2022instant}
Thomas M\"uller, Alex Evans, Christoph Schied, and Alexander Keller.
\newblock Instant neural graphics primitives with a multiresolution hash
  encoding.
\newblock \emph{ACM Trans. Graph.}, 41\penalty0 (4):\penalty0 102:1--102:15,
  2022.

\bibitem[Nicodemus(1965)]{nicodemus1965directional}
Fred~E Nicodemus.
\newblock Directional reflectance and emissivity of an opaque surface.
\newblock \emph{Applied optics}, 4\penalty0 (7):\penalty0 767--775, 1965.

\bibitem[Park et~al.(2013)Park, Sinha, Matsushita, Tai, and
  Kweon]{park2013multiview}
Jaesik Park, Sudipta~N Sinha, Yasuyuki Matsushita, Yu-Wing Tai, and In~So
  Kweon.
\newblock Multiview photometric stereo using planar mesh parameterization.
\newblock In \emph{{Proc. of International Conference on Computer Vision
  (ICCV)}}, pages 1161--1168, 2013.

\bibitem[Park et~al.(2016)Park, Sinha, Matsushita, Tai, and
  Kweon]{park2016robust}
Jaesik Park, Sudipta~N Sinha, Yasuyuki Matsushita, Yu-Wing Tai, and In~So
  Kweon.
\newblock Robust multiview photometric stereo using planar mesh
  parameterization.
\newblock \emph{{IEEE Transactions on Pattern Analysis and Machine Intelligence
  (PAMI)}}, 39\penalty0 (8):\penalty0 1591--1604, 2016.

\bibitem[Phong(1975)]{Phong1975IlluminationFC}
Bui~Tuong Phong.
\newblock Illumination for computer generated pictures.
\newblock \emph{Seminal graphics: pioneering efforts that shaped the field},
  1975.

\bibitem[Ravi et~al.(2025)Ravi, Gabeur, Hu, Hu, Ryali, Ma, Khedr, R{\"a}dle,
  Rolland, Gustafson, Mintun, Pan, Alwala, Carion, Wu, Girshick, Dollar, and
  Feichtenhofer]{ravi2024sam2}
Nikhila Ravi, Valentin Gabeur, Yuan-Ting Hu, Ronghang Hu, Chaitanya Ryali,
  Tengyu Ma, Haitham Khedr, Roman R{\"a}dle, Chloe Rolland, Laura Gustafson,
  Eric Mintun, Junting Pan, Kalyan~Vasudev Alwala, Nicolas Carion, Chao-Yuan
  Wu, Ross Girshick, Piotr Dollar, and Christoph Feichtenhofer.
\newblock {SAM} 2: {S}egment anything in images and videos.
\newblock In \emph{The Thirteenth International Conference on Learning
  Representations (ICLR)}, 2025.

\bibitem[Rusinkiewicz(1998)]{rusinkiewicz1998new}
Szymon~M Rusinkiewicz.
\newblock A new change of variables for efficient {BRDF} representation.
\newblock In \emph{Rendering Techniques’ 98: Proceedings of the Eurographics
  Workshop in Vienna, Austria, June 29—July 1, 1998 9}, pages 11--22.
  Springer, 1998.

\bibitem[Santo et~al.(2024)Santo, Okura, and Matsushita]{Santo_2024_CVPR}
Hiroaki Santo, Fumio Okura, and Yasuyuki Matsushita.
\newblock {MVCPS-NeuS: M}ulti-view constrained photometric stereo for neural
  surface reconstruction.
\newblock In \emph{{Proc. of Computer Vision and Pattern Recognition (CVPR)}},
  pages 20475--20484, 2024.

\bibitem[Srinivasan et~al.(2021)Srinivasan, Deng, Zhang, Tancik, Mildenhall,
  and Barron]{srinivasan2021nerv}
Pratul~P Srinivasan, Boyang Deng, Xiuming Zhang, Matthew Tancik, Ben
  Mildenhall, and Jonathan~T Barron.
\newblock {Nerv: N}eural reflectance and visibility fields for relighting and
  view synthesis.
\newblock In \emph{{Proc. of Computer Vision and Pattern Recognition (CVPR)}},
  pages 7495--7504, 2021.

\bibitem[Wang et~al.(2021)Wang, Liu, Liu, Theobalt, Komura, and
  Wang]{wang2021neus}
Peng Wang, Lingjie Liu, Yuan Liu, Christian Theobalt, Taku Komura, and Wenping
  Wang.
\newblock {NeuS: L}earning neural implicit surfaces by volume rendering for
  multi-view reconstruction.
\newblock \emph{{Advances in Neural Information Processing Systems (NeurIPS)}},
  2021.

\bibitem[Woodham(1980)]{woodham1980photometric}
Robert~J Woodham.
\newblock Photometric method for determining surface orientation from multiple
  images.
\newblock \emph{Optical engineering}, 19\penalty0 (1):\penalty0 139--144, 1980.

\bibitem[Wu et~al.(2010)Wu, Liu, Dai, and Wilburn]{wu2010fusing}
Chenglei Wu, Yebin Liu, Qionghai Dai, and Bennett Wilburn.
\newblock Fusing multiview and photometric stereo for 3d reconstruction under
  uncalibrated illumination.
\newblock \emph{IEEE transactions on visualization and computer graphics},
  17\penalty0 (8):\penalty0 1082--1095, 2010.

\bibitem[Yang et~al.(2022)Yang, Chen, Chen, Chen, and Wong]{yang2022psnerf}
Wenqi Yang, Guanying Chen, Chaofeng Chen, Zhenfang Chen, and Kwan-Yee~K. Wong.
\newblock Ps-nerf: {N}eural inverse rendering for multi-view photometric
  stereo.
\newblock In \emph{{Proc. of European Conference on Computer Vision (ECCV)}},
  2022.

\bibitem[Yao et~al.(2022)Yao, Zhang, Liu, Qu, Fang, McKinnon, Tsin, and
  Quan]{yao2022neilf}
Yao Yao, Jingyang Zhang, Jingbo Liu, Yihang Qu, Tian Fang, David McKinnon,
  Yanghai Tsin, and Long Quan.
\newblock Neilf: {N}eural incident light field for physically-based material
  estimation.
\newblock In \emph{European conference on computer vision}, pages 700--716.
  Springer, 2022.

\bibitem[Yariv et~al.(2020)Yariv, Kasten, Moran, Galun, Atzmon, Ronen, and
  Lipman]{idr2020multiview}
Lior Yariv, Yoni Kasten, Dror Moran, Meirav Galun, Matan Atzmon, Basri Ronen,
  and Yaron Lipman.
\newblock Multiview neural surface reconstruction by disentangling geometry and
  appearance.
\newblock \emph{{Advances in Neural Information Processing Systems (NeurIPS)}},
  33, 2020.

\bibitem[Yariv et~al.(2021)Yariv, Gu, Kasten, and Lipman]{yariv2021volume}
Lior Yariv, Jiatao Gu, Yoni Kasten, and Yaron Lipman.
\newblock Volume rendering of neural implicit surfaces.
\newblock In \emph{Thirty-Fifth Conference on Neural Information Processing
  Systems}, 2021.

\bibitem[Zhang et~al.(2022)Zhang, Luan, Li, and Snavely]{zhang2022iron}
Kai Zhang, Fujun Luan, Zhengqi Li, and Noah Snavely.
\newblock {IRON: I}nverse rendering by optimizing neural sdfs and materials
  from photometric images.
\newblock In \emph{{Proc. of Computer Vision and Pattern Recognition (CVPR)}},
  pages 5565--5574, 2022.

\bibitem[Zhang et~al.(2021)Zhang, Srinivasan, Deng, Debevec, Freeman, and
  Barron]{zhang2021nerfactor}
Xiuming Zhang, Pratul~P Srinivasan, Boyang Deng, Paul Debevec, William~T
  Freeman, and Jonathan~T Barron.
\newblock Nerfactor: {N}eural factorization of shape and reflectance under an
  unknown illumination.
\newblock \emph{ACM Transactions on Graphics (ToG)}, 40\penalty0 (6):\penalty0
  1--18, 2021.

\bibitem[Zhao et~al.(2023)Zhao, Lichy, Perrin, Frahm, and
  Sengupta]{zhao2023mvpsnet}
Dongxu Zhao, Daniel Lichy, Pierre-Nicolas Perrin, Jan-Michael Frahm, and
  Soumyadip Sengupta.
\newblock Mvpsnet: {F}ast generalizable multi-view photometric stereo.
\newblock In \emph{Proceedings of the IEEE/CVF International Conference on
  Computer Vision}, pages 12525--12536, 2023.

\bibitem[Zhou et~al.(2013)Zhou, Wu, and Tan]{zhou2013multi}
Zhenglong Zhou, Zhe Wu, and Ping Tan.
\newblock Multi-view photometric stereo with spatially varying isotropic
  materials.
\newblock In \emph{{Proc. of Computer Vision and Pattern Recognition (CVPR)}},
  pages 1482--1489, 2013.

\end{thebibliography}
}
\clearpage
\setcounter{page}{1}
\maketitlesupplementary

\begin{figure*}[h]
    \centering 
\begin{tabular}{@{}c@{}c@{}c@{}}
     \includegraphics[height=0.268\linewidth]{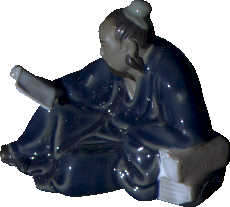} & 
    \includegraphics[height=0.268\linewidth]{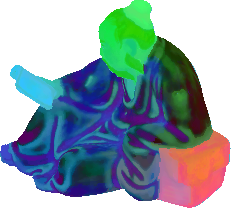} &
    \includegraphics[height=0.268\linewidth]{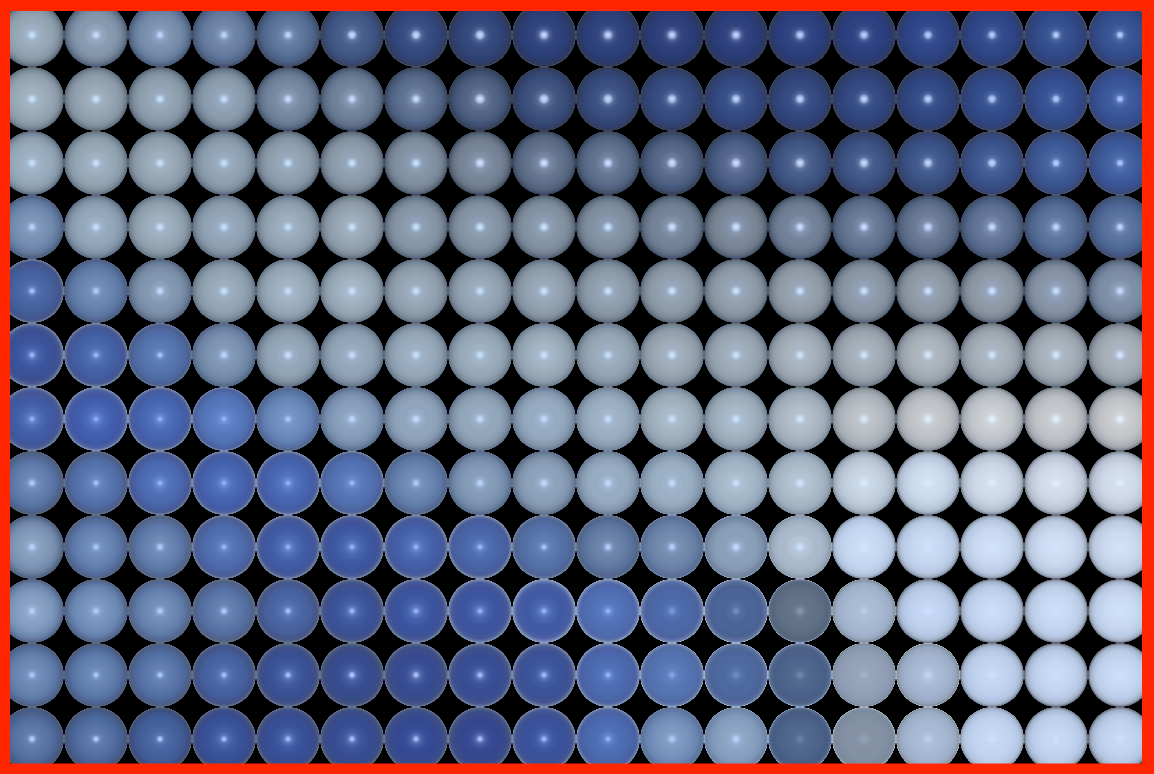} \\
    Captured image & Color-coded BRDF latent map & Close-up view
    \\
    \multicolumn{3}{@{}c@{}}{\includegraphics[width=\linewidth]{images/brdf_map/brdf_map_reading_small_bound.jpg}} \\
    \multicolumn{3}{c}{Full BRDF map}
\end{tabular}
    \caption{\textbf{BRDF map.} For each pixel in the input image, we retrieve its corresponding surface point from the spatial MLP and render its BRDF on a sphere under a colocated camera and light.}
    \label{fig.brdf_map_reading}
\end{figure*}

This supplementary material provides implementation details of our method in \cref{supp_sec.implementation_details}, real-world data collection details in \cref{sec.real_world_data}, scene normalization details in \cref{sec.scene_normalization}, and presents a visualization of the full BRDF map in \cref{fig.brdf_map_reading}.

\section{Implementation Details}
\label{supp_sec.implementation_details}

\subsection{Network Architecture}
\begin{figure*}
    \centering
    \includegraphics[width=\linewidth]{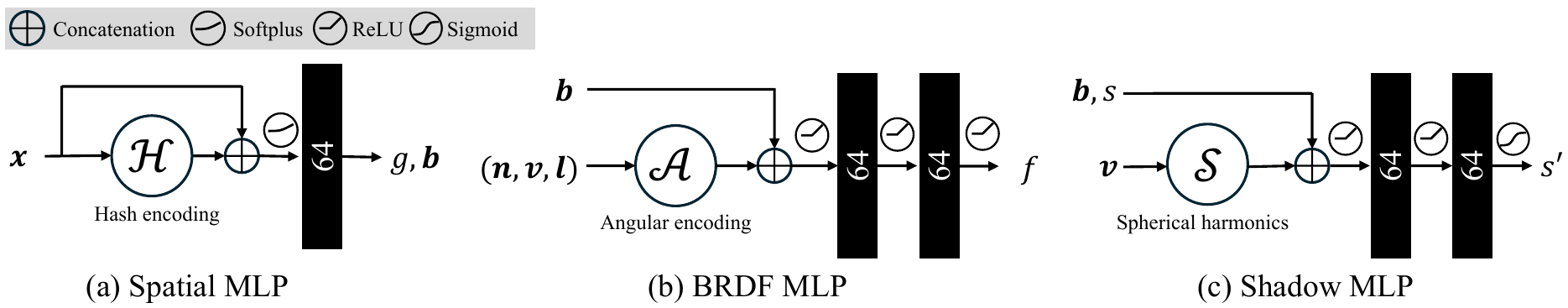}
    \vspace{-1em}
    \caption{Our MLP architectures.}
    \label{fig.mlp_archi}
\end{figure*}

As shown in \cref{fig.pipeline}, our forward rendering pipeline comprises three MLPs: the spatial MLP, the BRDF MLP, and the shadow MLP.
We detail each MLP’s architecture below and visualize them in \cref{fig.mlp_archi}. 

\paragraph{Spatial MLP.}
The spatial MLP takes as input the hash-encoded point features concatenated with the point’s coordinates and outputs a 64-dimensional vector. 
The first entry of the output vector represents the signed distance $g$, while the remaining entries serve as the BRDF latent vector $\V{b}$. 
The spatial MLP consists of one hidden layer with 64 channels using the softplus activation. 
The output layer does not employ any activation function.

Regarding hash encoding, we set the base resolution to $32$ and use $14$ levels with $2$ entries per level, yielding a $28$-dimensional feature vector.

\paragraph{BRDF MLP.}
The BRDF MLP inputs the BRDF latent vector $\V{b}(\point)$ and the angularly encoded normal-view-light directions, producing a $3$-dimensional output accounting for RGB channels. 
It comprises two hidden layers, each with 64 channels and ReLU activations.
The output layer also employs a ReLU activation to ensure non-negative BRDF values.

\paragraph{Shadow MLP.}
The shadow MLP takes as input the surface point’s latent vector $\V{b}(\point’)$, the viewing direction $\viewDir$, and the volume-rendered shadow value $s$. 
We apply a third-degree spherical harmonics encoding to the viewing direction. 
The shadow MLP consists of two hidden layers with 64 channels and ReLU activations. 
The output layer uses a sigmoid activation to ensure the output shadow factors are within the range $[0, 1]$.

\subsection{Loss Functions}
As described in \cref{sec.optimization}, the loss function comprises three terms: color loss, mask loss, and Eikonal loss. The color loss is defined in \cref{eq.color_loss} of the main paper.
Here, we provide details for the remaining two terms.
\paragraph{Mask loss.}
Given a batch of sampled pixels $\mathcal{P}$ from input images, we render the accumulated opacities $m(\pixel)$ and compare them against the input binary mask values $\hat{m}(\pixel)$ using the binary cross-entropy (BCE) loss:

\begin{equation}
\loss_{\text{mask}} = \sum_{\pixel \in \mathcal{P}} \text{BCE}\left(m(\pixel), \hat{m}(\pixel)\right).
\end{equation}

\paragraph{Eikonal loss.}
Along the viewing rays cast from the camera centers, we sample a set of points $\mathcal{X}$ via ray marching. The Eikonal loss encourages the gradient norms of the SDF at all sample points to be $1$:
\begin{equation}
    \loss_{\text{Eikonal}} = \sum_{\point \in \mathcal{X}} \left(\norm{\nabla g(\point)}_2 - 1\right)^2.
\end{equation}
Since we only sample points near the surface, the Eikonal loss enforces a unit spatial gradient only in the vicinity of the zero level set of the SDF.

\subsection{Optimization}
We implement our method using the PyTorch Lightning framework.
Tiny-cuda-nn~\cite{tiny-cuda-nn} and NerfAcc~\cite{nerfacc_2023_ICCV} are used to accelerate ray marching and volume rendering.
Training takes about \SI{20}{min} per DiLiGenT-MV object and about \SI{100}{min} per self-collected object on an NVIDIA A100 GPU.

All terms in the loss function are weighted equally.
We use the AdamW optimizer~\cite{loshchilov2018decoupled} with an initial learning rate of $1 \times 10^{-2}$ for the parameters of the spatial MLP and BRDF MLP and $1 \times 10^{-3}$ for the remaining parameters.
All parameters are jointly trained for $20{,}000$ steps on \diligentmv objects and $100{,}000$ steps on our self-collected data.
In each step, we randomly sample $4{,}096$ rays for rendering.

Using NerfAcc~\cite{nerfacc_2023_ICCV}, we update an occupancy grid every $16$ steps during optimization to skip ray marching in empty regions (grids with occupancy values within a tiny threshold). For shadow ray marching, we set $t_{\text{near}} = 1 \times 10^{-2}$ and $t_{\text{far}} = 0.5$, and uniformly sample 64 points along the shadow ray segment.

\subsection{Evaluation Metrics}

\paragraph{L2 Chamfer Distance (CD).} 
CD quantifies the similarity between two sets of points by calculating the average closest point distance from each point in one set to the other set.
Given two point sets \pointsetOne and \pointsetTwo, CD is defined as
\begin{equation}
	\begin{aligned}
\text{CD} = &\frac{1}{|\pointsetOne|} \sum_{\pointOne \in \pointsetOne} \min_{\pointTwo \in \pointsetTwo} \norm{\pointOne - \pointTwo}_2 \\
&+ \frac{1}{|\pointsetTwo|} \sum_{\pointTwo \in \pointsetTwo} \min_{\pointOne \in \pointsetOne} \norm{\pointTwo - \pointOne}_2.
	\end{aligned}
\end{equation}
Here, $\norm{\cdot}_2$ denotes the Euclidean distance, and $|\cdot|$ represents the cardinality of the point set. In this work, we measure Chamfer Distance in millimeters (\si{mm}). A lower Chamfer Distance indicates a greater similarity between the two point sets.

Following \supernormal, we compute the Chamfer Distance for points that are visible from the input views. 
To this end, we cast rays for pixels inside the foreground mask from all captured views and find their first intersection with the reconstructed or GT meshes. 

\paragraph{Mean Angular Error (MAE).}
MAE measures the average angular difference between corresponding unit vectors.
Given two sets of unit vectors $\{\V{n}_i\}$ and $\{\V{\hat{n}}_i\}$, the mean angular error is defined as 
\begin{equation}
\text{MAE} = \frac{1}{N} \sum_{i=1}^{N} \arccos{\left( \V{n}_i^\top \V{\hat{n}}_i \right)}.
\label{eq.mean_angular_error}
\end{equation}
In this work, MAE is measured in degrees. A lower MAE indicates a smaller average angular discrepancy between the vectors, signifying higher accuracy in normal or light direction estimation.

Since normal maps are rendered in multiple views, we apply \cref{eq.mean_angular_error} to normal vectors collected from all rendered views within the foreground masks. 
Because rotation transformations preserve angles, computing MAE using either world-space or camera-space normal maps yields the same value.

\paragraph{Peak Signal-to-Noise Ratio (PSNR).}
PSNR measures the similarity between the rendered images and the captured images. It is defined based on the mean squared error (MSE) as:
\begin{equation}
\text{PSNR} = 10 \log_{10} \left( \frac{\text{MAX}^2}{\text{MSE}} \right).
\end{equation}
Here, $\text{MAX}$ is the maximum possible pixel value of the image ($1$ in our case), and $\text{MSE}$ is the mean squared error between the rendered and captured images.
Higher PSNR values indicate better image quality.

\paragraph{Scale-invariant Mean Squared Error (SI-MSE).}
As the light intensities can only be estimated up to a scale, we follow previous work~\cite{li2022selfps} to use the scale-invariant relative error defined as:

\begin{equation}
\text{SI-MSE} = \frac{1}{M} \sum_{j=1}^{M} \frac{\left| s_e \hat{e}_j - e_j \right|}{e_j},
\label{eq.scale_invariant_error}
\end{equation}

where $\hat{e}_j$ and $e_j$ denote the estimated and ground truth light intensities of the $j$-th light, respectively. The scale factor $s_e$ is computed by solving the least squares problem:

\begin{equation}
s_e = \arg\min_{s_e} \sum_{j=1}^{M} \left( s_e \hat{e}_j - e_j \right)^2.
\label{eq:scale_factor}
\end{equation}

A lower scale-invariant relative error indicates a more accurate estimation of the light intensities.

\section{Real-World Data Collection}
\label{sec.real_world_data}
\subsection{Capture Setup Details}
As shown in \cref{fig.capture_setup_real}, we prepare six strobe lights that emit point flashlights and a Sony ILCE-7RM5 camera equipped with a telephoto zoom lens.
Both the camera and the strobes are mounted on tripods to ensure that each light remains fixed relative to the camera. 
Since our method does not require a light to be static to other lights, preparing $M$ light sources is equivalent to repositioning a single source $M-1$ times during capture.

Following the setup of \diligentmv, we place the stobes approximately \SI{1}{m} away from centimeter-scale objects, so that the point flashlight can be safely approximated as directional.
To ensure sufficient image coverage, we adjust the focal length between \SI{70}{mm} and \SI{100}{mm} depending on the object size.
The aperture is set between F/11 and F/22 to achieve a large depth of field, ensuring the entire object remains in focus.  
The ISO is set between 100 and 200 to minimize sensor noise.

Capture is conducted indoors.
While a darkroom is ideal for OLAT acquisition, it is not a strict requirement as long as the flash emits a burst of light that is momentarily much stronger than the ambient illumination. 
Under such lighting conditions, a short camera exposure (e.g., 1/200 sec) effectively reduces the influence of ambient light to a negligible level.

\begin{figure*}[th]
    \centering
    \begin{tabular}{@{}c@{}c@{}c@{}c@{}c@{}}
    \includegraphics[width=0.2\linewidth]{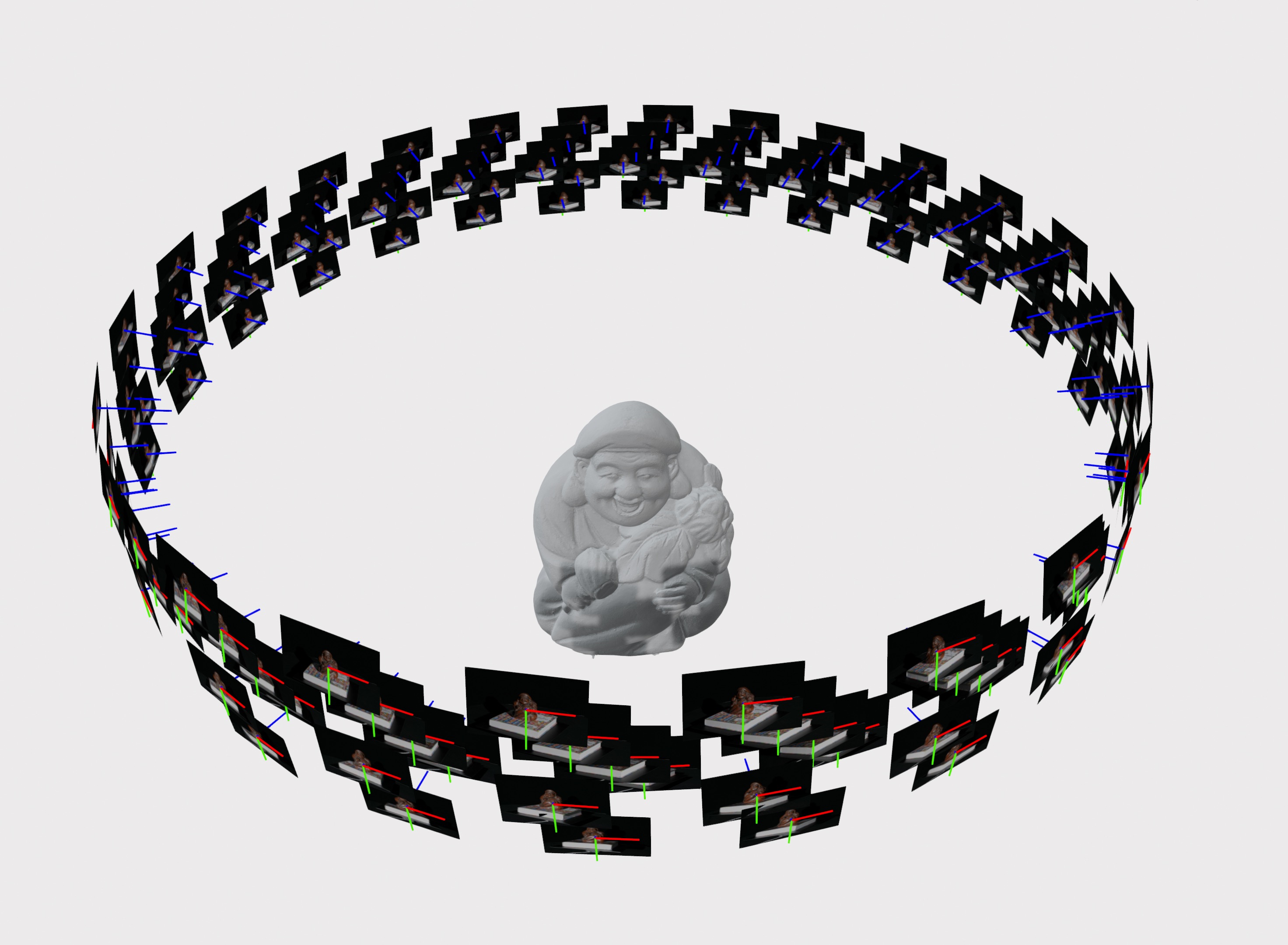} & 
    \includegraphics[width=0.2\linewidth]{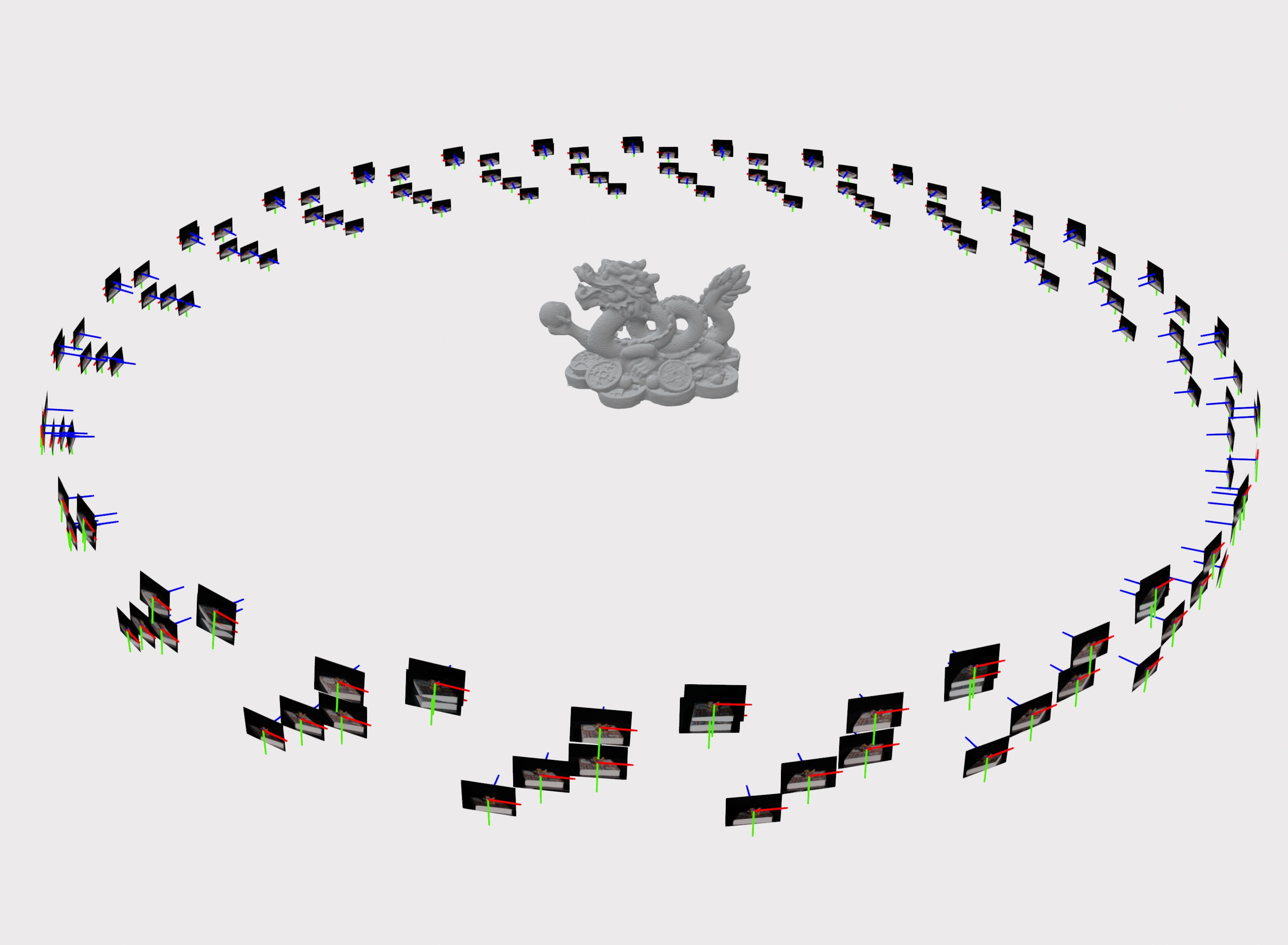}    & 
    \includegraphics[width=0.2\linewidth]{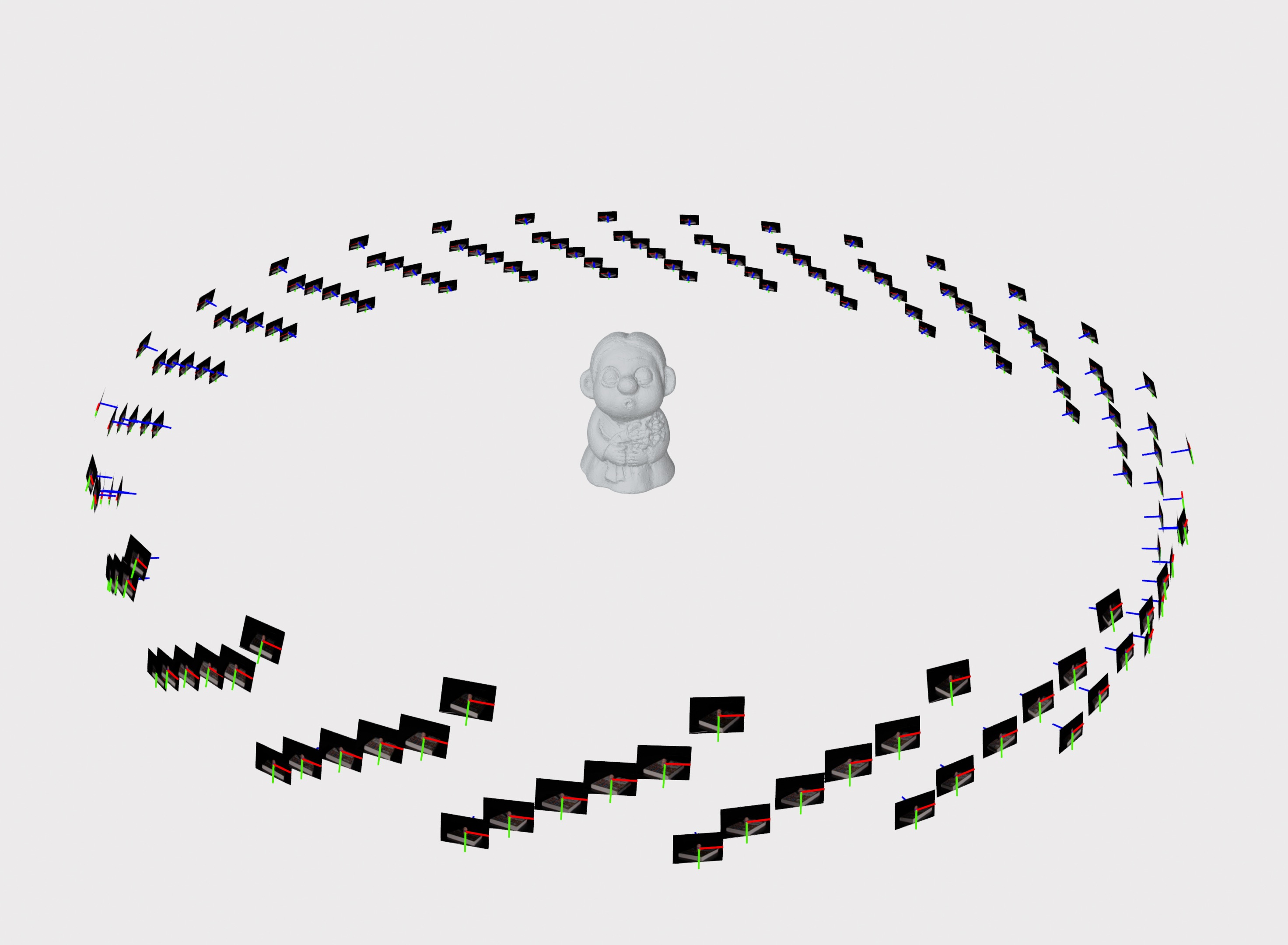} &
    \includegraphics[width=0.2\linewidth]{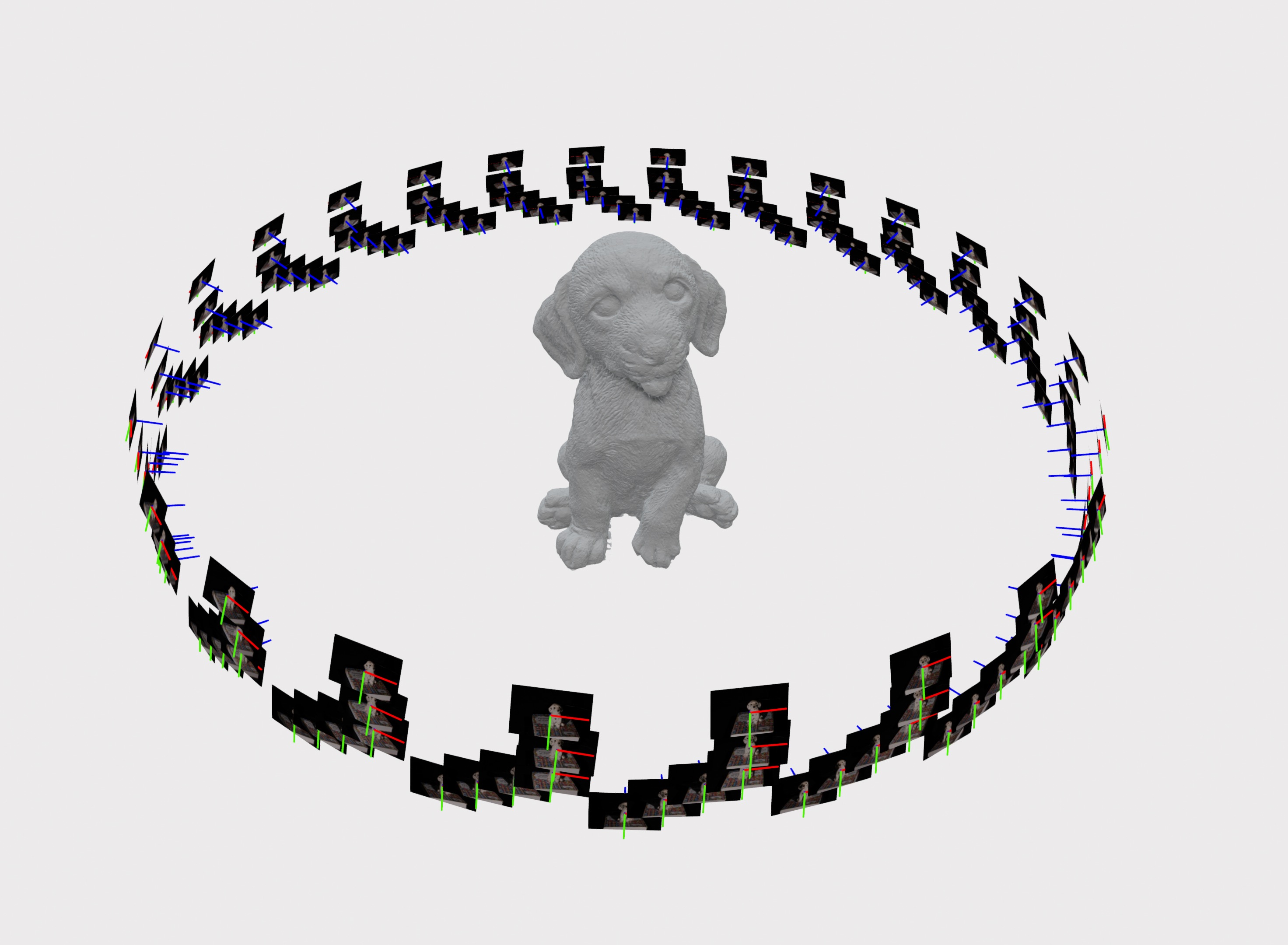} &
    \includegraphics[width=0.2\linewidth]{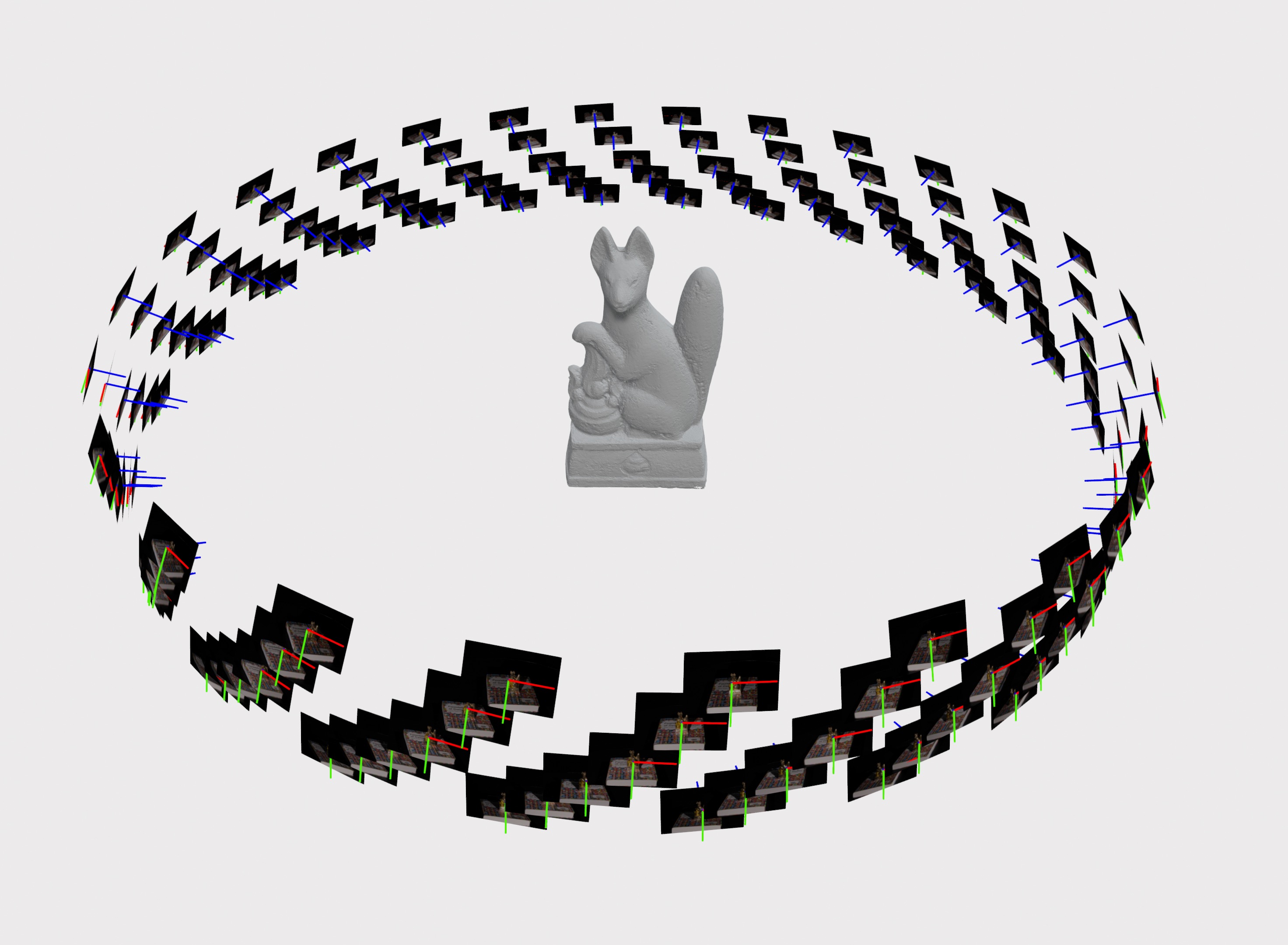}
    \end{tabular}
    \caption{Visualization of camera positions with respect to the object. Our setup does not contain aligned viewpoints across lighting, and images on the same elevation are captured under the same light source. Meshes are produced by RealityCapture~\cite{realitycapture} and enlarged for improved visibility.}
    \label{fig.real_world_setup}
\end{figure*}

\Cref{fig.real_world_setup} visualizes the distribution of camera viewpoints.
A motorized turntable rotates the object in $15^\circ$ increments, yielding $24$ uniformly distributed  multi-view OLAT images per strobe light.
Each time we switch to a different light, we adjust the camera height and elevation angles to introduce more viewpoint variations.
In total, we capture 144 OLAT images per object (6L24V) at a resolution of $9504\times6336$. 
Under our setting, the real-world resolution of each pixel is approximately \SI{0.04}{mm}.

\subsection{Image Preprocessing}
The DSLR camera produces a pair of raw and JPEG images per exposure. 
The raw images are unprocessed and directly fed into our algorithm. 
For camera calibration, we use the JPEG images and employ RealityCapture~\cite{realitycapture} due to its efficiency and robustness.
This yields a single $3 \times 4$ world-to-image projection matrix per view, encoding both the world-to-camera transformation and the perspective camera intrinsics.

Multi-view foreground masks are generated using SAM2~\cite{ravi2024sam2}, which supports efficient and automated cross-view mask propagation. 
With a few mouse clicks on a single image, SAM2~\cite{ravi2024sam2} segments the corresponding mask and propagates it to other views in seconds.
To facilitate more reliable foreground segmentation, we apply gamma correction to the JPEG images to enhance contrast between the foreground object and the background.

Finally, given the camera parameters and multi-view foreground masks, we perform scene normalization such that the target object is enclosed within a unit sphere.
This process is detailed in the next section.

\section{Scene Normalization}
\label{sec.scene_normalization}
Scene normalization applies a global scaling and translation to world coordinates such that the target object is bounded within a unit sphere~\cite{idr2020multiview}. 
This normalization facilitates the training of the neural SDF and is conducted before reconstruction.
For clarity, \textbf{we refer to the normalized world coordinates as the object coordinates}, where the unit sphere is centered at the origin.
In the following, \cref{sec.coord_transfrom} describes the coordinate system transformations involved in our rendering pipeline, and \cref{sec.o2w_trans} provides a robust scene normalization method based on camera parameters and multi-view foreground masks.

\subsection{Coordinate System Transformations}
\label{sec.coord_transfrom}
We define the object coordinates to differ from the world coordinates by an isotropic scale and translation without rotational motion. 
Specifically, for a point  $\V{x}^{\text{O}} \in \R^3$ in object coordinates, its world coordinates is given by $\V{x}^{\text{W}} = s\V{x}^{\text{O}} +\V{d}$, where $s \in \R_+$ is the object-to-world scale and $\mathbf{d} \in \R^3$ is the object-to-world translation\footnote{The symbol $s$ denotes the object-to-world scale in this section only, whereas it is referred to the shadow factor in the main paper.}. In homogenous coordinates, the object-to-world transformation $\V{T}_\text{O2W}$ is
\begin{equation}
    \tilde{\V{x}}^{\text{W}} = \underbrace{
    \left[\begin{matrix}
        s\V{I} & \V{d} \\
        \V{0}^\top & 1
    \end{matrix}\right]}_{\mathbf{T}_\text{O2W}} \tilde{\mathbf{x}}^{\text{O}}.
\end{equation}

With the object space, three  transformations are involved to project an object-space point onto the $i$-th image plane: Object space → world space → $i$-th camera space → $i$-th image plane, which can be formally described as

\begin{align}
    z_i\tilde{\V{u}_i} &= \V{K}_i\V{T}_{\text{W2C}_i} \V{T}_{\text{O2W}}\tilde{\mathbf{x}}^{\text{O}}
\\
 &= \V{K}_i\left[\begin{matrix}\V{R}_i &\V{t}_i\end{matrix}\right]\left[\begin{matrix}
        s\V{I} & \V{d} \\
        \V{0}^\top & 1
    \end{matrix}\right] \tilde{\mathbf{x}}^{\text{O}}.
\end{align}
Here, $\tilde{\V{u}_i}$ is the homogeneous coordinates of the pixel coordinates, $z_i$ is the depth of the pixel, $\V{K}_i$ is the $3 \times 3$ intrinsic matrix of the $i$-th camera, and $\V{R}_i \in SO(3)$ and $\V{t}_i \in \R^3$ represent the world-to-camera rotation and translation, respectively.

\paragraph{Camera position in object coordinates.}
In volume rendering, camera position and viewing directions are required to determine the ray from which we sample points. 
In object coordinates, the camera position $\V{c}_i^{\text{O}}$ can be obtained from its world coordinates $\V{c}_i^{\text{W}}$ as
\begin{equation}
    \V{c}_i^{\text{O}} = (\V{c}_i^{\text{W}} - \V{d})/s \quad \text{with} \quad \V{c}_i^{\text{W}} = -\V{R}_i^\top\V{t}_i.
\end{equation}
Equivalently, $\V{c}_i^{\text{O}}$ can be obtained as the null vector of object-to-camera transformation since it transforms the camera position to the origin in camera coordinates:
\begin{equation}
    \V{T}_{\text{W2C}_i} \V{T}_{\text{O2W}} \tilde{\V{c}}_i^{\text{O}} = \V{0}.
\end{equation}

\paragraph{Ray directions in object coordinates.}
The scale and translation do not affect directional vectors. 
Therefore, the viewing directions in object coordinates stay the same as in world coordinates:
\begin{equation}
    \V{v}^\text{O} = \V{v}^\text{W} = \V{R}_i^\top \V{K}_i^{-1} \tilde{\V{u}}.
 \end{equation}
The same applies to normal and light directions.

\paragraph{Mesh vertices in world coordinates.}
Since the neural SDF is trained in object coordinates, extracting its zero-level set by marching cubes yields a mesh with vertices in object coordinates. 
Converting the mesh to world coordinates is realized by 
\begin{equation}
 \V{p}^\text{W} = s\V{p}^\text{O} + \V{d},
 \label{eq.mesh_O2W}
\end{equation}
where $\V{p}^\text{O}$ indicates the object coordinates of a mesh vertex, and \cref{eq.mesh_O2W} is applied to all vertices.

\subsection{O2W Scale and Translation Estimation}
\label{sec.o2w_trans}
Estimating the object-to-world scale and translation before reconstruction requires prior knowledge about the scene.
For object-centric reconstruction, the object space origin can be defined as the point closest to the camera principal axes of all views~\cite{yariv2021volume}. 
However, this may not position the object in the center of the unit sphere well if the principal axes of the surrounding cameras do not point to the target object.
For large-scale scene reconstruction, sparse 3D points reconstructed by structure-from-motion are used~\cite{li2023neuralangelo}, and manual effort is required to segment the scene of interest.

In this work, we use the known camera parameters and foreground masks, which are readily available as input to our method, to normalize the scene.
Given that the object is assumed to lie within a unit sphere in object space, the projection of this unit sphere onto each image should ideally encompass the foreground pixels across all views.
Based on this constraint, we use a two-step process to estimate the object-to-world translation and scale.

\paragraph{O2W translation estimation.} 
We expect the object to be centered at the origin of the object space.
To this end, we estimate the O2W translation vector $\V{d}$ such that the object space origin lies as close as possible to the foreground center-of-mass rays across all views.
Let $\V{o}_i + t\V{v}_i$ be the ray in world coordinates passing through the center of mass of the foreground region in the $i$-th view, where $\V{o}_i$ is the camera center and $\V{v}_i$ is the corresponding viewing direction.
In world coordinates, the origin of the object space is located at $\V{d}$.
The squared distance from $\V{d}$ to the ray is given by
\begin{equation}
    d^2(\V{d}, \V{o}_i + t\V{v}_i) = \V{d}^\top \V{V}_i \V{d} - 2 \V{o}_i^\top \V{V}_i \V{d} + \V{o}_i^\top \V{V}_i \V{o}_i,
\end{equation}
where $\V{V}_i=\V{I}-\V{v}_i\V{v}_i^\top$.
The O2W translation is then obtained by minimizing the sum of squared distances across all views:
\begin{equation}
   \V{d}^* = \argmin_{\V{d}} \sum_i d^2(\V{d}, \V{o}_i + t\V{v}_i).
   \label{eq.o2w_trans}
\end{equation}
The objective \cref{eq.o2w_trans} is quadratic and convex, and thus attains a global optimum at the critical point where the gradient takes $0$. 
Setting the gradient of \cref{eq.o2w_trans} to zero yields the normal equation:
\begin{align}
       \V{V}^\top \V{V}\V{d} &= \V{V}^\top \V{b} \\
    \text{with} \quad \V{V} &=\sum_i \V{V}_i, \quad \V{b} = \sum_i \V{V}_i \V{o}_i. 
\label{eq.o2w_normal_eq}
\end{align}
We solve \cref{eq.o2w_normal_eq} using a least-squares solver.

\begin{figure*}[th]
    \centering
    \begin{tabular}{@{}c@{ }c@{ }c@{ }c@{ }c@{}}
        \includegraphics[width=0.2\linewidth]{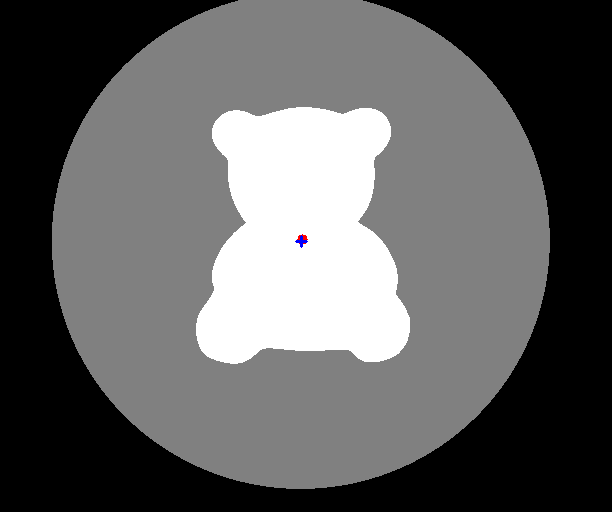} &
        \includegraphics[width=0.2\linewidth]{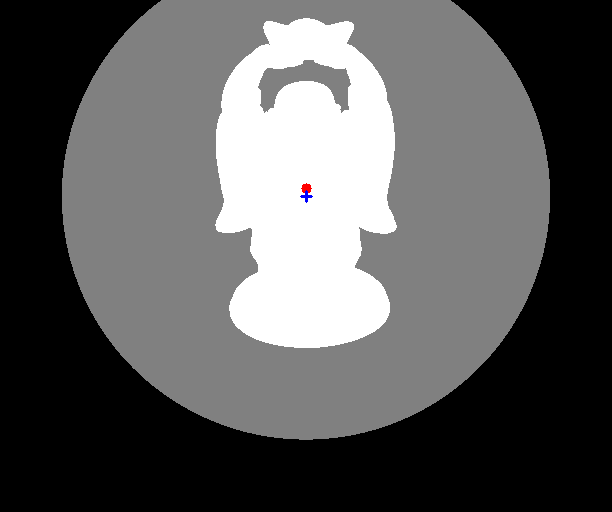} &
        \includegraphics[width=0.2\linewidth]{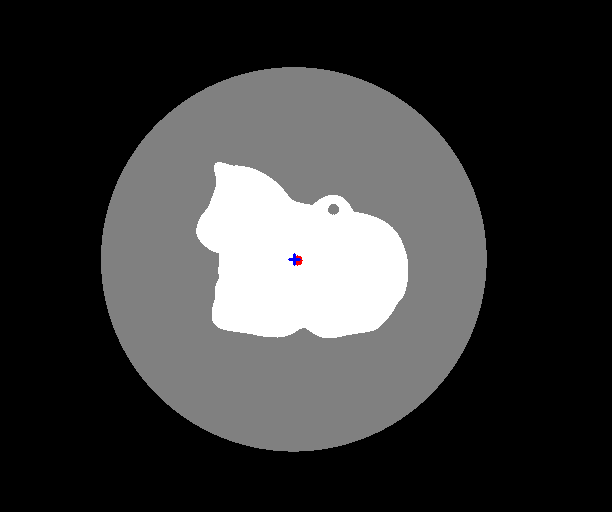} &
        \includegraphics[width=0.2\linewidth]{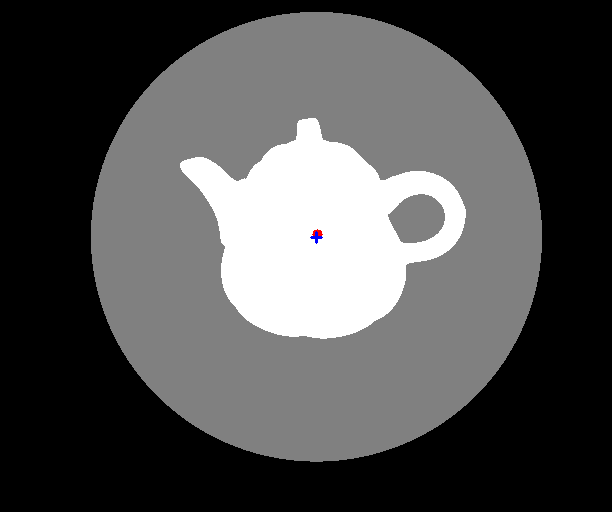} &
        \includegraphics[width=0.2\linewidth]{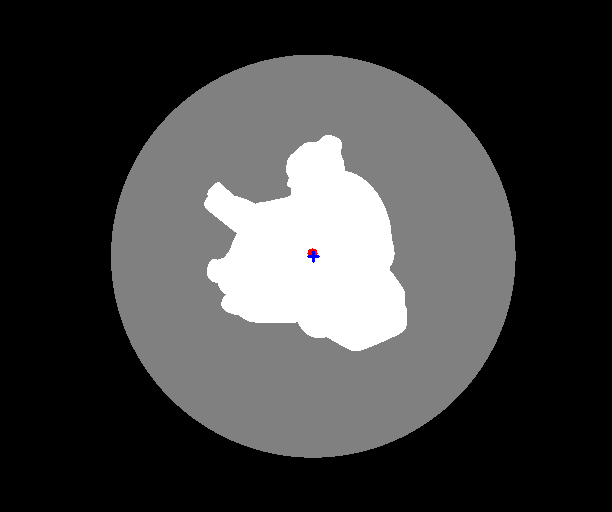}
        \\
        \includegraphics[width=0.2\linewidth]{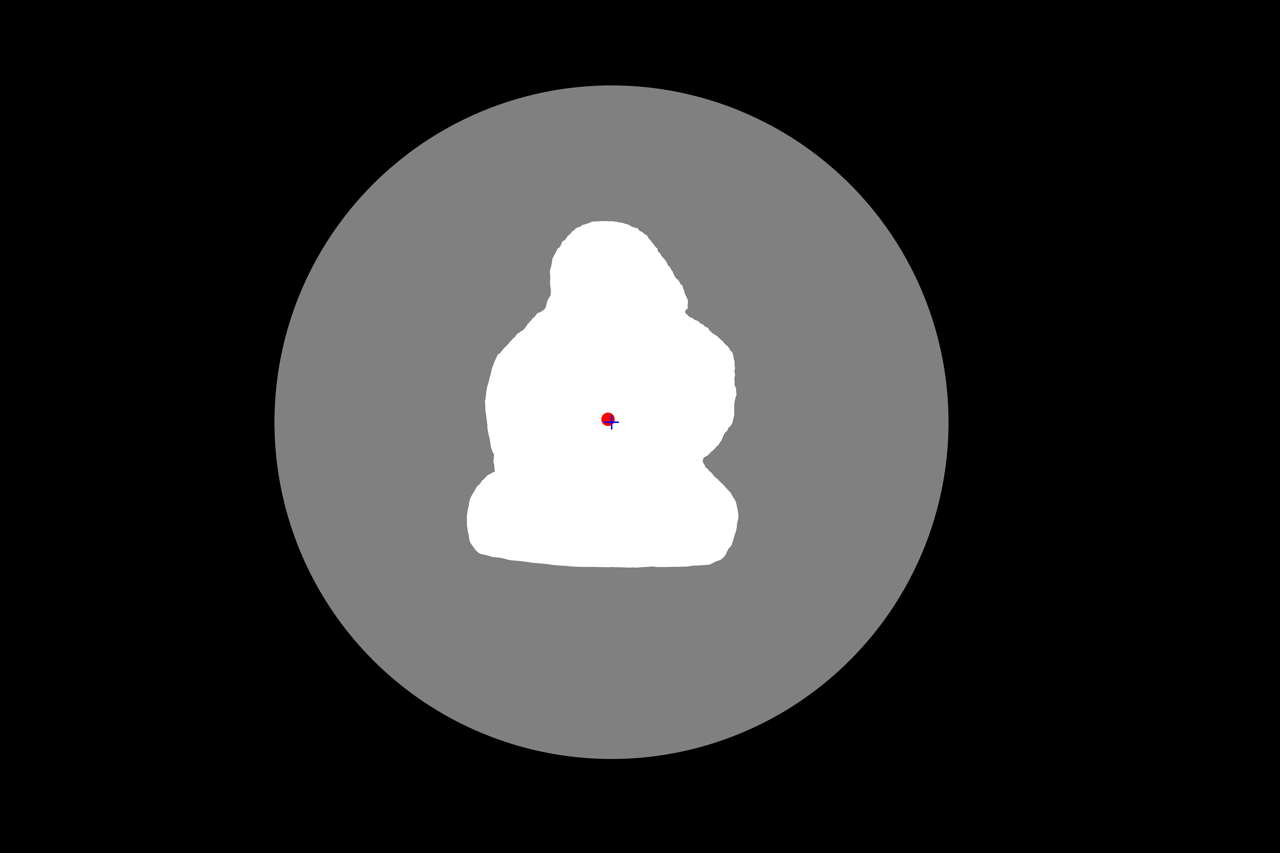} &
        \includegraphics[width=0.2\linewidth]{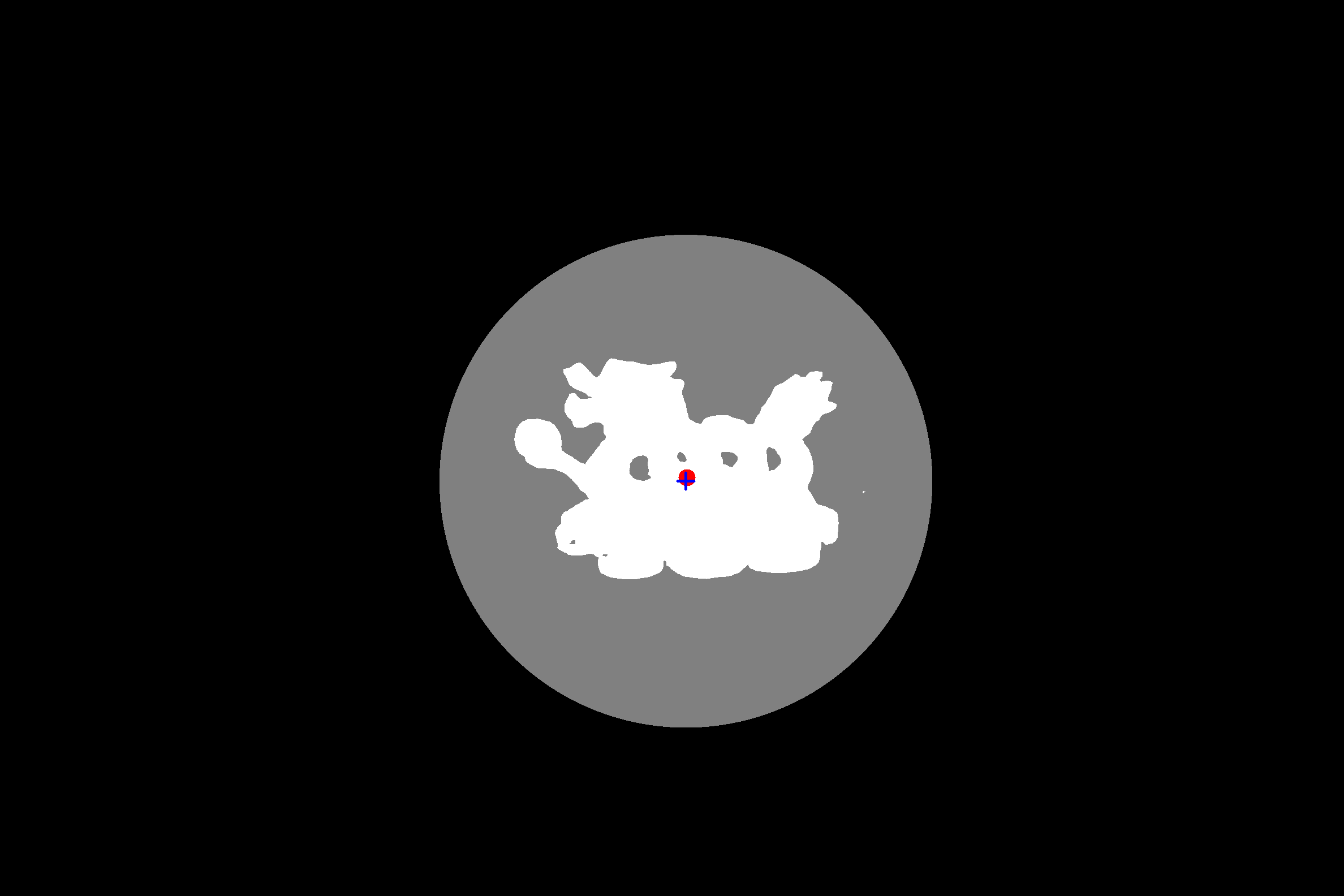} &
        \includegraphics[width=0.2\linewidth]{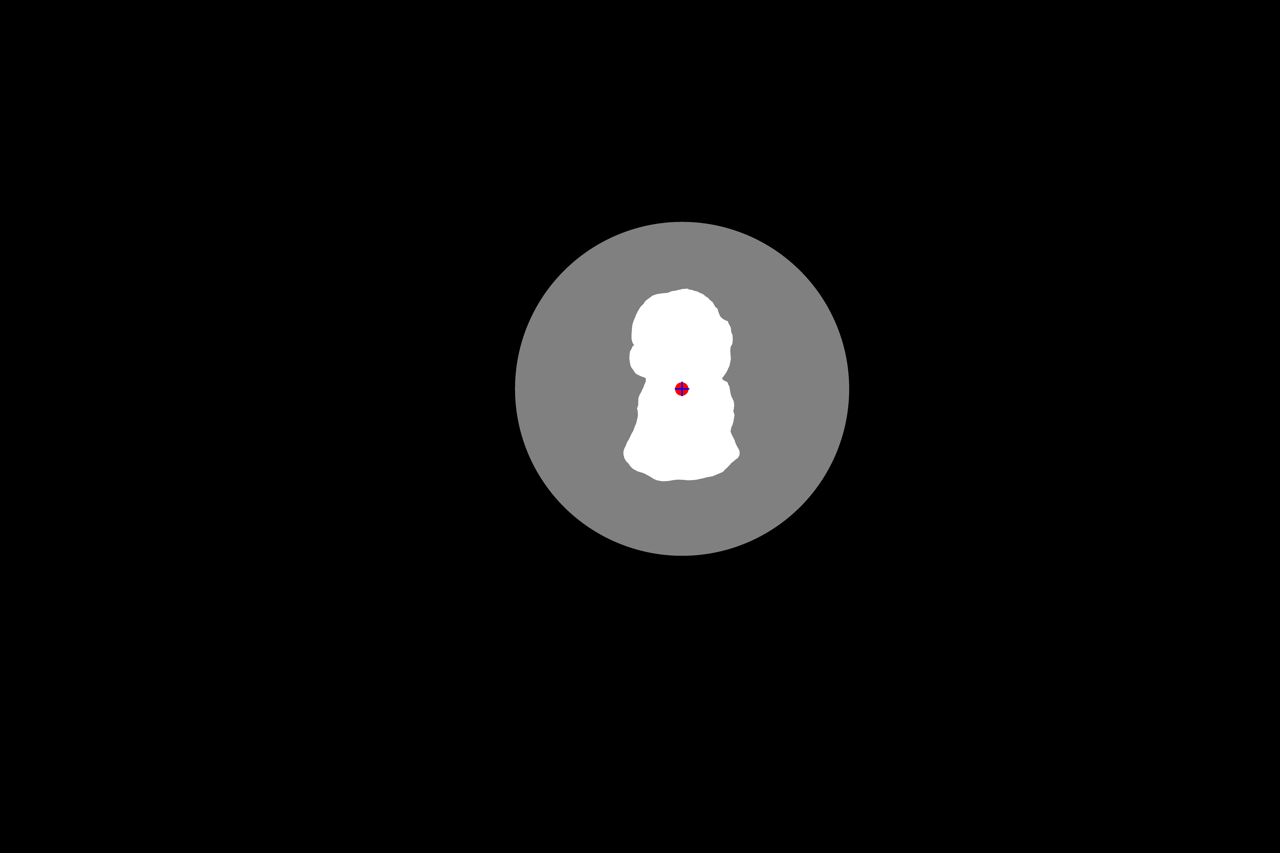} &
        \includegraphics[width=0.2\linewidth]{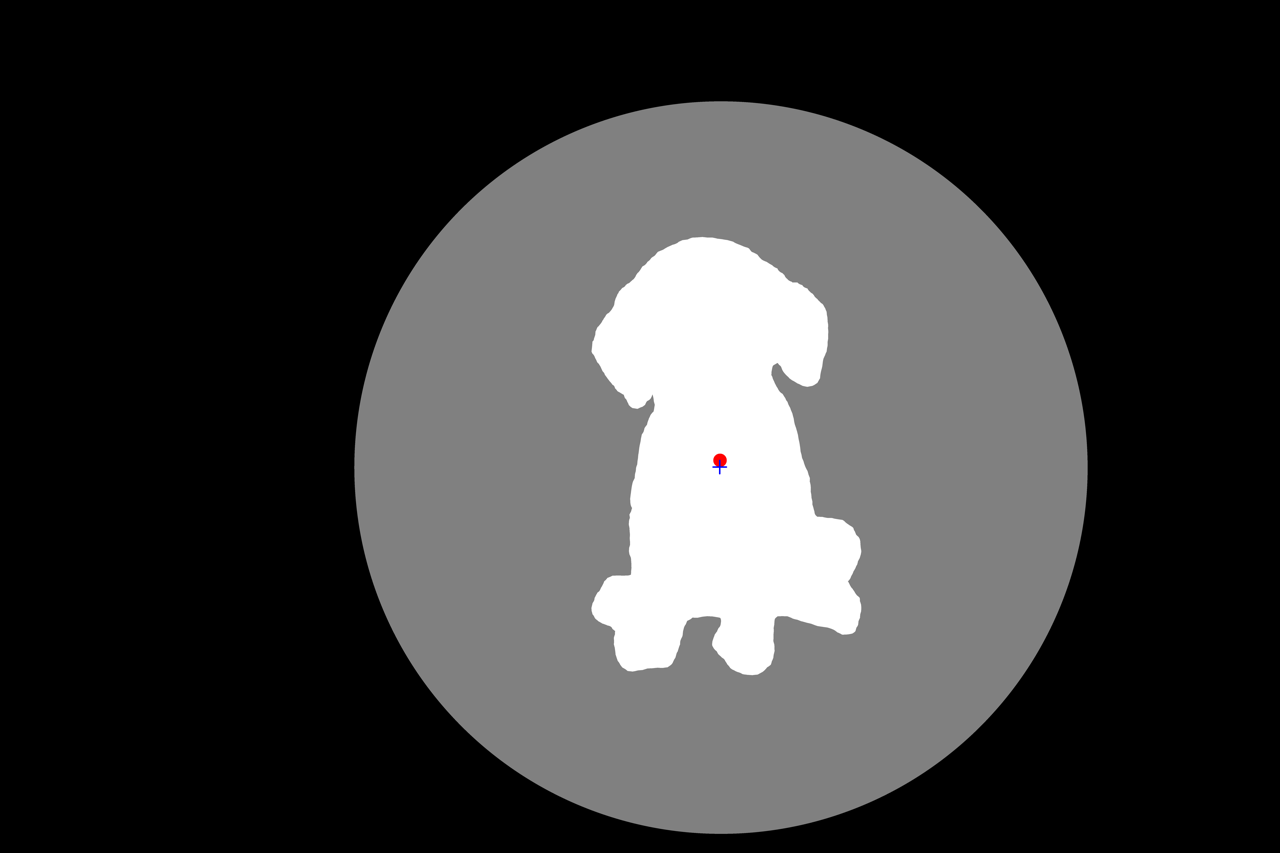} &
        \includegraphics[width=0.2\linewidth]{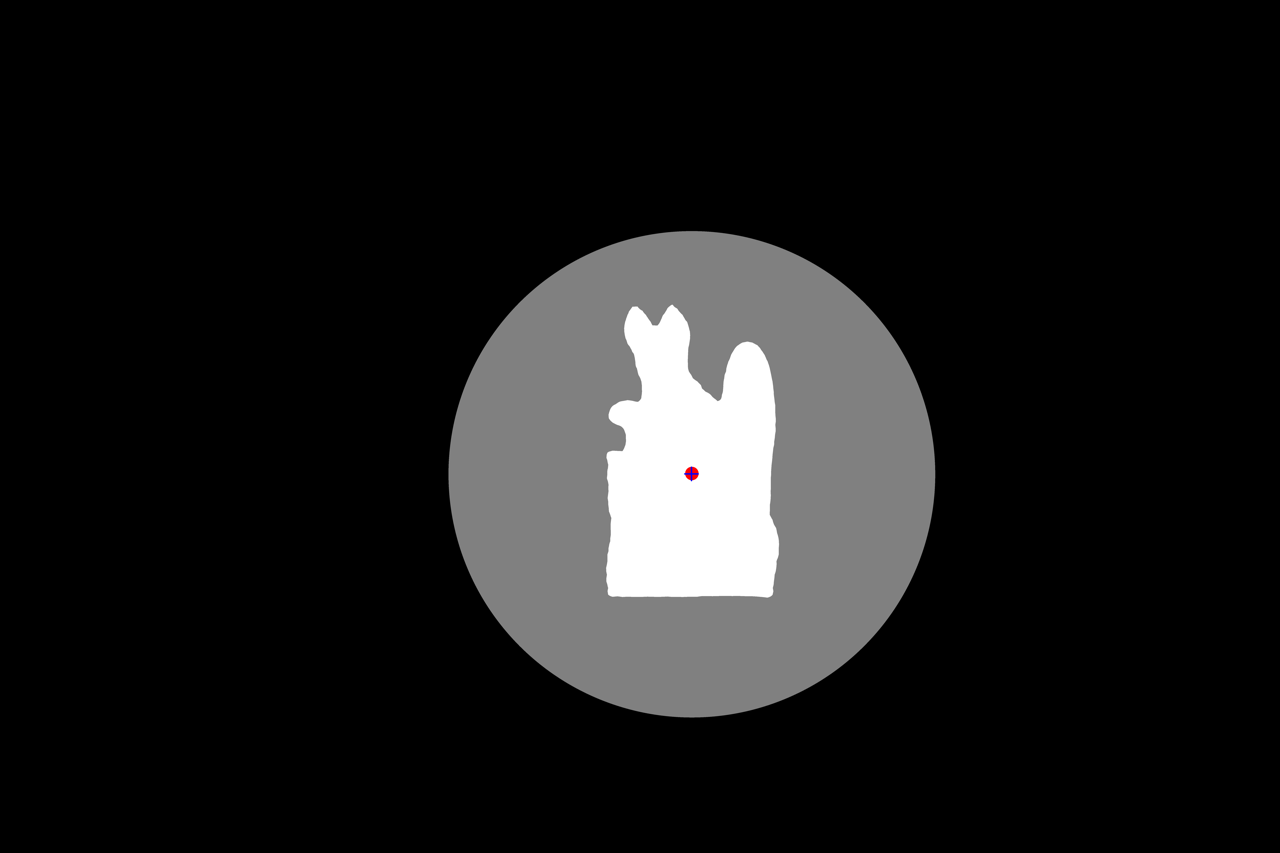}
    \end{tabular}
    \caption{
\textbf{Visualization of scene normalization results.}
(Top row) \diligentmv scenes. (Bottom row) Our self-captured scenes. White pixels indicate foreground regions, gray pixels denote the projected unit sphere, red circles mark the center of mass of the foreground regions, and blue crosses indicate the centers of the projected unit sphere.}
    \label{fig.scene_normalization}
\end{figure*}

\paragraph{O2W scale estimation.}
Once the O2W translation $\V{d}^*$ is determined, we then estimate the O2W scale $s$ such that the projected unit sphere fully encloses the foreground regions in all views.
This is enforced by ensuring that the sum of the projected sphere areas of all views is greater than a factor $k$ times the sum of all foreground areas:
\begin{equation}
    s^* = \argmin_s \left(\sum_i A_i(s)\right) \geq k\sum_i \hat{A}_i.
    \label{eq.O2W_scale_objective}
\end{equation}
$\hat{A}_i$ is the foreground area, and $A_i(s)$ is the projected sphere area in the $i$-th view.
To analytically compute $A_i(s)$, we make two approximations: (1) The projection of a 3D sphere onto the image plane is approximately circular, and (2) the radius $r_i$ of this projected circle can be approximated using the principle of similar triangles:
\begin{equation}
    \frac{r_i}{s} \approx \frac{f_i}{z_i},
\end{equation}
where $f_i$ is the focal length of $i$-th viewpoint, $z_i$ is the depth of object-space origin in $i$-th camera coordinates, \ie, $z_i = (\V{R}_i \V{d}^* + \V{t}_i)_z$.
With the approximations, the projected area of the unit sphere in the $i$-th view is given by
\begin{equation}
    A_i(s) = \pi r_i^2 \approx \frac{\pi s^2 f_i^2}{z_i^2}.
    \label{eq.projected_sphere_area}
\end{equation}
Substituting \cref{eq.projected_sphere_area} into \cref{eq.O2W_scale_objective} yields a closed-form solution for the O2W scale:
\begin{equation}
    s = \sqrt{\frac{k\sum_i \hat{A}_i}{\pi \sum_i\left(\frac{f_i^2}{z_i^2}\right)}}.
\end{equation}
Empirically, we find that a scaling factor of $k = 5$ works well for all \diligentmv scenes and our captured objects, as shown in \cref{fig.scene_normalization}.

\end{document}